\documentclass[a4paper,11pt,oneside]{article}
\usepackage{amsmath}
\usepackage{amssymb}
\usepackage{amsthm}
\usepackage{mathrsfs}
\usepackage{isomath}
\usepackage{dsfont}
\usepackage{thmtools}
\usepackage{xcolor}
\usepackage{natbib}
\usepackage{subcaption}
\usepackage{booktabs,multirow} 
\usepackage{algorithm}
\usepackage{algpseudocode}
\usepackage{graphicx}
\usepackage{nameref}

\newtheoremstyle{example}
  {\topsep}
  {\topsep}
  {\small}
  {0pt}
  {\bfseries}
  {.}
  { }
  {\thmname{#1}\thmnumber{ #2}\thmnote{ (#3)}}

\theoremstyle{remark} 
\theoremstyle{definition} \newtheorem{definition}{Definition}[section]
\theoremstyle{example} 
\theoremstyle{plain} 
\theoremstyle{plain} 
\theoremstyle{plain} 
\theoremstyle{plain} 
\theoremstyle{plain} 
\theoremstyle{plain} 

\newcommand \horizon {T} 
\newcommand \lookahead {H} 
\newcommand \mdp {\mu} 
\newcommand \pol {\pi} 
\newcommand \mdpk {\mu^{(k)}} 
\newcommand \MDPs {\mathcal{M}} 
\newcommand \Qb {\mathscr{Q}} 
\newcommand \VbC[3] {\mathscr{V}_{#1,#2}^{#3}} 
\newcommand \VbS[2] {\mathscr{V}_{#1,#2}^{*}} 
\newcommand \VC[3] {V_{#1,#2}^{#3}} 
\newcommand \Vktn {V^{(k)}_{i+1}} 
\newcommand \Vjkt {V^{(j,k)}_{i}} 
\newcommand \Vtn {V_{i+1}} 
\newcommand \Vt {V_{i}} 
\newcommand \CS {\mathcal{S}} 
\newcommand \CA {\mathcal{A}} 
\newcommand \CV {\mathcal{V}} 
\newcommand \CR {\mathcal{R}} 
\newcommand \rew {\rho} 

\newcommand \bel {\beta} 
\newcommand \vbel {\psi} 
\newcommand \vbelt {\psi_i} 
\newcommand \vbeltn {\psi_{i+1}} 

\newcommand \eye {\matrixsym{I}}
\newcommand \bm {\vectorsym{m}}
\newcommand \bS {\vectorsym{\Sigma}}

\newcommand \defn {\mathrel{\triangleq}}
\newcommand \Bellman {\mathscr{B}}

\newcommand \util {U}
\newcommand \val {\vectorsym{v}}
\newcommand \valk {\vectorsym{v}^{(k)}} 

\newcommand \discount {\gamma}

\newcommand \nofVals {N_V}
\newcommand \nofMDPs {N_\mdp}
\newcommand \nofStates {N_{\CS}}
\newcommand \nofActions {N_{\CA}}

\DeclareMathOperator{\argmax}{arg\,max}
\DeclareMathOperator{\argmin}{arg\,min}

\DeclareMathOperator{\E}{\mathbb{E}}
\newcommand \dd {\, \mathrm{d}}
\let\Pr\relax
\DeclareMathOperator \Pr {\mathbb{P}}

\newcommand \cset[2] {\left\{#1 ~\middle|~ #2\right\}}

\newcommand \ind[1] {\mathds{1}\left\{#1\right\}}

\DeclareMathAlphabet{\mathpzc}{OT1}{pzc}{m}{it}

\newcommand \Normal {{\mathpzc{N}}}

\def\clap#1{\hbox to 0pt{\hss#1\hss}}

\usepackage[scaled]{helvet}
\usepackage{microtype}
\usepackage{graphicx}
\usepackage{subcaption}
\usepackage{hyperref}
\usepackage{cleveref}
\usepackage{booktabs,multirow} 
\usepackage{algorithm}
\def \LongVersion {1}

\begin{document} 

\renewcommand*{\thefootnote}{\fnsymbol{footnote}}

\title{Inferential Induction: A Novel Framework for Bayesian Reinforcement Learning}
\author{Emilio Jorge\footnotemark[1]~\footnotemark[2] \and Hannes Eriksson\footnotemark[1]~\footnotemark[2] \and Christos Dimitrakakis\footnotemark[1]~\footnotemark[2]~\footnotemark[3] \and Debabrota Basu\footnotemark[2] \and Divya Grover\footnotemark[2]}
\maketitle
\footnotetext[1]{Equal contribution}
\footnotetext[2]{Chalmers University of Technology}
\footnotetext[3]{University of Oslo}
\renewcommand*{\thefootnote}{\arabic{footnote}}

	
	

	\begin{abstract}
  Bayesian reinforcement learning (BRL) offers a decision-theoretic solution for reinforcement learning. While ``model-based'' BRL algorithms have focused either on maintaining a posterior distribution on models or value functions and combining this with approximate dynamic programming or tree search, previous Bayesian ``model-free'' value function distribution approaches implicitly make strong assumptions or approximations. We describe a novel Bayesian framework, Inferential Induction, for correctly inferring value function distributions from data, which leads to the development of a new class of BRL algorithms. We design an algorithm, Bayesian Backwards Induction, with this framework. We experimentally demonstrate that the proposed algorithm is competitive with respect to the state of the art.
	\end{abstract} 
	\section{Introduction}

Many Reinforcement Learning (RL) algorithms are grounded on the application of dynamic programming to a Markov Decision Process (MDP)~\citep{sutton}. When the underlying MDP $\mdp$ is known, efficient algorithms for finding an optimal policy exist that exploit the Markov property by calculating value functions. Such algorithms can be applied to RL, where the learning agent simultaneously acts in and learns about the MDP, through e.g. stochastic approximations, without explicitly reasoning about the underlying MDP. Hence, these algorithms are called model-free.


In Bayesian Reinforcement Learning (BRL)~\citep{brl}, we explicitly represent our knowledge about the underlying MDP $\mdp$ through some prior distribution $\bel$ over a set $\MDPs$ of possible MDPs. While model-based BRL is well-understood, many works on BRL aim to become model-free by directly calculating distributions on value functions. Unfortunately, these methods typically make strong implicit assumptions or approximations about the underlying MDP.

This is the first paper to directly perform Bayesian inference over value functions without any implicit assumption or approximation. We achieve this through a novel BRL framework, called \textit{Inferential Induction}, extending backwards induction. This allows us to perform joint Bayesian inference on MDPs and value function as well as to optimise the agent's policy. We instantiate and experimentally analyse only one of the many possible algorithms, \textit{Bayesian Backwards Induction (BBI)}, in this family and show it is competitive with the current state of the art.

In the rest of this section, we provide background in terms of setting and related work. In \Cref{sec:inferential-induction}, we explain our Inferential Induction framework and three different inference methods that emerge from it, before instantiating one of them into a concrete procedure. Based on this, 
\ifdefined \Neurips
\Cref{sec:algorithms} describes the BBI algorithm. In \Cref{sec:experiments}, we experimentally compare BBI with state-of-the art BRL algorithms.
\ifdefined \LongVersion
This version of the paper has a significantly longer and more detailed main text, as well as an appendix with additional experimental results in \Cref{sec:additional-results} and some implementation details in \Cref{sec:impl-deta}.
\else
A longer version of the paper with further mathematical and experimental details is provided in the supplementary material.
\fi
\else 
\Cref{sec:algorithms} describes the BBI algorithm. In \Cref{sec:experiments}, we experimentally compare BBI with state-of-the art BRL algorithms.
\fi

\subsection{Setting and Notation}
\label{sec:setting}
In this paper, we generally use $\Pr$ and $\E$ to refer to probability (measures) and expectations while allowing some abuse of notation for compactness.

\textbf{Reinforcement Learning (RL)} is a sequential learning problem faced by agents acting in an unknown environment $\mdp$, typically modelled as a Markov decision process~\citep[c.f.][]{Puterman:MDP:1994}.
 \begin{definition}[Markov Decision Process (MDP)]
  An MDP $\mdp$ with state space $\CS$ and action space $\CA$ is equipped with a reward distribution $\Pr_\mdp(r \mid s)$ with corresponding expectation $\rew_\mdp(s)$ and a transition kernel $\Pr_\mdp(s' | s, a)$ for states $s, s' \in \CS$ and actions $a \in \CA$.
\end{definition}
At time $t$, the agent observes\footnote{In the partially-observable setting, the agent instead observes another variable dependent on the state.}  
the environment state $s_t$, and then selects an action $a_t$. Then, it receives and observes a reward $r_t$ and a next state $s_{t+1}$.  The agent is interested in the utility
$U_t \defn \sum_{k=t}^\horizon\discount^{k-t} r_k$, i.e. the sum of future rewards $r_t$. Here, $\discount \in (0,1]$ is the discount factor and $\horizon \in [1, \infty]$ is the problem horizon. Typically, the
agent wishes to maximise the \emph{expected utility}, but other objectives are possible.

The agent acts in the environment using a policy $\pol = (\pol_1, \ldots, \pol_t, \ldots)$ that takes an action $a_t$ at time $t$ with probability $\pol_t(a_t \mid s_t, r_{t-1}, a_{t-1}, s_{t-1}, \ldots, r_1, a_1, s_1)$. Dependence on the complete observation history is necessary, if the agent is learning from experience.
However, when $\mdp$ is known, the policy $\pol^*_\mdp$ maximising expected utility over finite horizon is Markovian\footnote{For infinite horizon problems this policy is still Markovian, but can be non-stationary.} of the form $\pol_t(a_t \mid s_t)$ and is computable using dynamic programming. A useful algorithmic tool for achieving this is the value function, i.e. the expected utility of a policy $\pol$ from different starting states and action:
\begin{definition}[Value Function]
  The state value function of policy $\pol$ in MDP $\mdp$ is $\VC{\mdp}{t}{\pol}(s) \defn \E^\pol_\mdp(U_t \mid s_t = s)$  and the corresponding state-action (or Q-)value function is
      $Q^\pol_{\mdp,t}(s, a) \defn \E^\pol_\mdp(U_t \mid s_t = s, a_t = a)$. $\Pr^\pol_\mdp$ and $\E^\pol_\mdp$ denote probabilities and expectations under the process induced by $\pol$ and $\mdp$.
\end{definition}
Finally, the \emph{Bellman operator}
$\Bellman_\mdp^\pol V(s) \defn \rew_\mdp(s) + \discount \sum_{s' \in \CS}  \Pr_\mdp^\pol(s' \mid s) V(s')$
allows us to compute the value function recursively through $\VC{\mdp}{t}{\pol} = \Bellman_\mdp^\pol \VC{\mdp}{t+1}{\pol}$.\footnote{In the discounted setting, the value function converges to $V^\pol_\mdp \equiv V^\pol_{\mdp,1}$ as $\horizon \to \infty$.}

\paragraph{Bayesian RL (BRL).} In BRL, our subjective belief is represented as a probability measure $\bel$ over possible MDPs.
We refer to the initial belief $\bel$ as the \textit{prior distribution}. By interacting with the environment until time $t$, the agent obtains \textit{data} $D = (s_1, a_1, r_1, \ldots, s_t)$. This data is used to calculate a \textit{posterior distribution} $\bel(\mdp \mid D)$ that represents agent's current knowledge about the MDP.\footnote{This is expressible in closed form. When the MDP is discrete, a Dirichlet-product prior can be used, or when the MDP is continuous and the dynamics are assumed to be linear, a Gaussian-Wishart prior can be used~\citep{Degroot:OptimalStatisticalDecisions}. Gaussian process inference can also be expressed in a closed-form but inference becomes approximate because the computational complexity scales quadratically with time.}
For a given belief and an adaptive policy\ifdefined \LongVersion
\footnote{Typically the adaptive policy's actions depends on the complete history, but we can equivalently write it as depending on the current belief and state instead. It is also possible to consider the Bayesian value function of policies whose beliefs disagrees with the actual MDP distribution, but this is beyond the scope of this paper.}
\fi
 $\pol_\bel(s)$, we define the \emph{Bayesian value function} to be:
\begin{align}
  \VbC{\bel}{t}{\pol_\bel}(s)  &\defn    \int_\MDPs \VC{\mdp}{t}{\pol_\bel}(s) \dd \bel(\mdp).
  \label{eq:bayesian-value-function}
\end{align}
The Bayesian value function is the expected value function under the distribution $\bel$.
The \emph{Bayes-optimal policy} achieves the Bayes-optimal value function $\VbS{\bel}{t}(s) = \sup_\pol \VbC{\bel}{t}{\pol}$.  
Calculating $\VbC{\bel}{t}{\pol}$ involves integrating $\VC{\mdp}{t}{\pol}$ for all $\mdp$, while  $\VbC{\bel}{t}{*}$ typically requires exponential time. 
Information about the value function distribution can be a useful tool for constructing near-optimal policies, as well a way to compute risk-sensitive policies.

\ifdefined \LongVersion
\paragraph{Distributions over Value Functions.} Let us consider the value function $V$, with $V = (V_1, \ldots, V_\horizon)$ for finite-horizon problems, a prior belief $\bel$ over MDPs, and a previously collected data $D = (s_1, a_1, r_1, \ldots, s_{t-1}, a_{t-1}, r_{t-1}, s_t, r_t)$ using some policy $\pol$. Now, the posterior value function distribution is expressed in terms of the MDP posterior:
\begin{equation}
\label{eq:value-function-distribution}
\Pr_\bel(V \mid D)
= \int_{\MDPs} \Pr_\mdp(V) \dd \bel(\mdp \mid D).
\end{equation}
\eqref{eq:value-function-distribution} induces an empirical measure $\hat{P}_{MC}^{E}$ that corresponds to the standard Monte-Carlo estimate:
\begin{equation}
  \hat{P}_{MC}^{E}(B) \defn \nofMDPs^{-1} \sum_{k=1}^K \ind{\valk \in B},
  \label{eq:empirical-monte-carlo}
\end{equation}
where $\ind{}$ is the indicator function.
The practical implementation is in Algorithm~\ref{alg:mc-vf}.\footnote{Algorithm~\ref{alg:mc-vf} has $O(\nofMDPs \nofStates^2 \nofActions \horizon)$ complexity for policy evaluation, while policy optimisation can be performed through approximate dynamic programming~\citep{dimitrakakis:mmbi:ewrl:2011} or Bayesian gradient ascent~\citep{ghavamzadeh:bpga}.}
\begin{algorithm}
  \caption{A Monte-Carlo Estimation of Value Function Distributions}
  \label{alg:mc-vf}
  \begin{algorithmic}[1]
    \State Select a policy $\pol$.
    \For{$k=1,\ldots, \nofMDPs$}
    \State Sample an MDP $\mdpk \sim \bel$.
    \State Calculate
    $\valk = \VC{\mdpk}{}{\pol} \sim \bel$.
    \EndFor
    \State \textbf{return} $\hat{P}_{MC}^{E}(\{\valk\})$
  \end{algorithmic}
\end{algorithm}
\fi
\subsection{Related Work and Our Contribution}
\label{sec:related-work}
\textbf{Model-free Bayesian Value Functions.}
Bayesian value function distributions have been considered extensively in model-free Bayesian Reinforcement Learning (BRL).
One of the first methods was Bayesian Q-learning~\citep{dearden1998bayesian}, which used a normal-gamma prior on the utility distribution. However, as i.i.d. utility samples cannot be obtained by bootstrapping from value function estimates, this idea had inherent flaws. \citet{engel2003bayes} developed a more sophisticated approach, the Gaussian Process Temporal Difference (GPTD) algorithm,
which has a Gaussian process (GP) prior $\bel(V)$ on value functions. It then combines this with the likelihood function $\Pr(D \mid V) \propto \prod_{i=1}^t \exp\{-|V(s_i)- r_i - \discount V(s_{i+1})|^2\}$.
However, this makes the implicit assumption that the deterministic empirical MDP model is true. 
\citet{Engel05}~tried to relax this assumption by allowing for correlation between sequentially visited states. 
 \citet{Deisenroth2009} developed a dynamic programming algorithm with a GP prior on value functions and an explicit GP model of the MDP. 
Finally, \citet{tang2018exploration} introduced VDQN, generalising such methods to Bayesian neural networks. 
The assumptions that these model-free Bayesian  methods implicitly make about the MDP are hard to interpret, and we find the use of an MDP model independently of the value function distribution unsatisfactory. We argue that explicitly reasoning about the joint value function and MDP distribution is necessary to obtain a coherent Bayesian procedure. Unlike the above methods, we calculate a value function posterior $\Pr(V | D)$  while simultaneously taking into account uncertainty about the MDP.

\textbf{Model-based Bayesian Value Functions.} 
If a posterior over MDPs is available, we can calculate a distribution over value functions in two steps: 
a) sample from the MDP posterior and b) calculate the value function of each MDP. 
 \citet{dearden1999model} suggested an early version of this approach that obtained approximate upper bounds on the Bayesian value function and sketched a Bellman-style update for performing it online. Posterior sampling approach was later used to obtain value function distributions in the discrete case by~\citet{dimitrakakis:mmbi:ewrl:2011} and in the continuous case by~\citet{Osband:icml:2016}. 
We instead focus on whether it is possible to compute value function distributions exactly or approximately through a \textit{backwards induction} procedure. 
In particular, how can we obtain $\Pr(V_i | D)$ from $\Pr(V_{i+1} | D)$?

\textbf{Utility Distributions.} A similar problem is calculating \emph{utility} (rather
than value) distributions through Bellman updates. 
Essentially, this is the problem of estimating $\Pr_\mdp(U_i \mid U_{i+1})$ for a given MDP $\mdp$. In this context,
\citet{morimura2010nonparametric} constructed risk-sensitive policies. More recently~\citet{bellemare2017distributional} showed that modelling the full utility distribution may also be useful
for exploration.
However, the utility distribution is due to the stochasticity of the transition
kernel $\Pr_\mdp(s_{t+1} \mid s_t, a_t)$ rather than uncertainty about
the MDP, and hence a different quantity from the value
function distribution, which this paper tries to estimate.

\textbf{Bayes-optimal approximations.} It is also possible to define the value function with respect to the information state $(s_t, \bel_t)$. This generates a Bayes-adaptive Markov decision process (BAMDP~\cite{duff2002olc}). However, BAMDPs are exponentially-sized in the horizon due to the increasing number of possible information states as we look further into the future. A classic approximate algorithm in this setting is Bayesian sparse sampling~\citep[BSS]{wang2005bayesian}.
\ifdefined \LongVersion
BSS in particular generates a sparse BAMDP by sampling a finite number of belief states at each step $i$ in the tree, up to some fixed horizon $\horizon$. In addition, it can also sparsely sample actions by selecting a random action $a$ through posterior sampling at each step $i$. 
\fi
In comparison, our value function distributions at future steps can be thought of as \emph{marginalising} over possible future information states. This makes our space complexity much smaller.


\paragraph{Our Contribution.}
We introduce \textit{Inferential Induction}, a new Bayesian
Reinforcement Learning (BRL) framework, which leads to a Bayesian form
of backwards induction. Our framework allows Bayesian inference over value functions without any implicit assumption or approximation unlike its predecessors.
The main idea is to calculate the
conditional value function distribution at step $i$ from the value
function distribution at step $i+1$ analogous to backwards induction
for the expectation (Eq.~\eqref{eq:inferential-induction-compact}).  Following this, we propose three possible marginalisation techniques (Methods 1, 2 and 3) and design a Monte-Carlo approximation with Method 1. We can combine this procedure with a policy optimisation mechanism.   
We use a Bayesian adaptation of dynamic programming for this and propose the \textit{Bayesian backwards induction (BBI)} algorithm.
Our experimental evaluation shows that BBI is competitive to the current state of the art. 
Inferential Induction framework provides the opportunity to further design more efficient algorithms of this family.


	\section{Inferential Induction}
\label{sec:inferential-induction}
The fundamental problem is calculating the value function distribution $\Pr_\bel^\pol(\Vt \mid D)$ for a policy\footnote{Here we drop the subscript $\bel$ from the policy for simplicity.} $\pol$ under the belief $\bel$. 
The main idea is to inductively calculate $\Pr^\pol_\bel(\Vtn \mid D)$ from $\Pr^\pol_\bel(\Vt \mid D)$ for $i \geq t$ as follows:
\begin{equation}
  \label{eq:inferential-induction-compact}
  \Pr^\pol_\bel(\Vt \mid D)
  = \int_{\CV} \Pr^\pol_\bel(\Vt \mid \Vtn, D) \dd \Pr^\pol_\bel(\Vtn \mid D).
\end{equation}
Let $\vbeltn$  be a (possibly approximate) representation of $\Pr_\bel^\pol(\Vtn  \mid D)$. If we can calculate the above integral, then we can also obtain $\vbelt \approx \Pr_\bel^\pol(\Vt \mid D)$ recursively, from time $T$ up to the current time step $t$.  
Then the problem reduces to defining the term $\Pr^\pol_\bel(\Vt \mid \Vtn, D)$ appropriately. 
We describe three methods for doing so, and derive and experiment on an algorithm for one specific case, in which Bayesian inference can also be performed through conventional priors. As all the methods that we describe involve some marginalisation over MDPs as an intermediate step, the main practical question is what form of sampling or other approximations suit each of the methods.

\textbf{Method 1: Integrating over $\Pr^\pol_\bel(\mdp \mid \Vtn, D)$.}\label{method1} A simple idea for dealing with the term linking the two value functions is to directly marginalise over the MDP as follows:
\begin{equation}\label{eq:method1}
  \Pr^\pol_\bel(\Vt \mid \Vtn, D) 
= \int_{{\MDPs}} \Pr^\pol_\mdp(\Vt \mid \Vtn) \dd \Pr^\pol_\bel(\mdp \mid \Vtn, D).
\end{equation}
This equality holds because given $\mdp$, $\Vt$ is uniquely determined by the policy $\pol$ and $\Vtn$ through the Bellman operator. However, it is crucial to note that $\Pr^\pol_\bel(\mdp \mid \Vtn, D) \neq \Pr_\bel(\mdp \mid D)$, as knowing the value function gives information about the MDP.\footnote{Assuming otherwise results in a mean-field approximation.
\ifdefined \LongVersion
See Sec.~\ref{sec:mean-field}.
\else
See Sec. 2.2. in the extended version.
\fi
}

\textbf{Method 2: Integrating over $\Pr^\pol_\bel(\mdp \mid \Vt, \Vtn)$.}
From Bayes' theorem, we can write the conditional probability of $\Vt$ given $\Vtn, D$ in terms of the data likelihood of $\Vt, \Vtn$ and the conditional distribution $\Vt | \Vtn$, as follows\footnote{Here, $B\subset \CV$ are sets of value functions in an appropriate $\sigma$-algebra.}:
\ifdefined \Neurips
\begin{align}
  \Pr^\pol_\bel(\Vt \in B \mid \Vtn, D)
 = \frac{\int_B \Pr^\pol_\bel(D \mid \Vt, \Vtn) \dd \Pr^\pol_\bel(\Vt \mid \Vtn)}
{\int_{\CV} \Pr^\pol_\bel(D \mid \Vt, \Vtn) \dd \Pr^\pol_\bel(\Vt \mid \Vtn)}.
\end{align}
\else
\begin{align*}
  \Pr^\pol_\bel(\Vt \in B \mid \Vtn, D)
= \frac{\int_B \Pr^\pol_\bel(D \mid \Vt, \Vtn) \dd \Pr^\pol_\bel(\Vt \mid \Vtn)}
{\int_{\CV} \Pr^\pol_\bel(D \mid \Vt, \Vtn) \dd \Pr^\pol_\bel(\Vt \mid \Vtn)}.
\end{align*}
\fi
The \emph{likelihood term} is crucial in this formulation. One way to write it is as follows:
\ifdefined \Neurips
\begin{align}
\label{eq:V-likelihood}
\Pr_\bel^\pol (D \mid \Vt, \Vtn)
& = \int_\MDPs \Pr_\bel^\pol(D \mid \mdp) \dd \Pr^\pol_\bel(\mdp \mid \Vt, \Vtn).
\end{align}
\else
\begin{align*}
  \Pr_\bel^\pol (D \mid \Vt, \Vtn)
  & = \int_\MDPs \Pr_\bel^\pol(D \mid \mdp) \dd \Pr^\pol_\bel(\mdp \mid \Vt, \Vtn).
\end{align*}
\fi
This requires us to specify some appropriate distribution  $\Pr^\pol_\bel(\mdp \mid \Vt, \Vtn)$ that we can sample from, meaning that standard priors over MDPs cannot be used. On the other hand, it allows us to implicitly specify MDP distributions given a value function, which may be an advantage in some settings. 

\textbf{Method 3: Integrating over $\bel(\mdp \mid D)$.} Using the same idea as Method 2, but using Bayes's theorem once more, we obtain:
\ifdefined \Neurips
\begin{align}
\label{eq:V-likelihood-posteriora}
\Pr_\bel^\pol (D \mid \Vt, \Vtn)
& = \int_\MDPs
\frac{\Pr(\mdp \mid \Vt, \Vtn, \pol)\Pr_\bel^\pol (D) }
{\bel (\mdp) }
\dd \bel (\mdp \mid D).
\end{align}
\else
\begin{align*}
\Pr_\bel^\pol (D \mid \Vt, \Vtn)
& = \int_\MDPs
\frac{\Pr(\mdp \mid \Vt, \Vtn, \pol)\Pr_\bel^\pol (D) }
{\bel (\mdp) }
\dd \bel (\mdp \mid D).
\end{align*}
\fi 

While there are many natural priors from which sampling from $\bel (\mdp \mid D)$ is feasible, we still need to specify $\Pr_\bel(\mdp \mid \Vt, \Vtn, \pol)$. This method might be useful when the distribution that we specify is easier to evaluate than to sample from.
It is interesting to note that if we replace $\bel(\mdp \mid D)$ with a point distribution (e.g. the empirical MDP), the inference becomes similar in form to GPTD~\citep{engel2003bayes} and GPDP~\citep{Deisenroth2009}.
\ifdefined \LongVersion
In particular, this occurs when we set $\Pr(\mdp \mid \Vt, \Vtn, \pol) \propto \exp\{-\|\Vtn- \rew_\mdp - \discount P_\mdp^\pol \Vt\|^2\}$. However, this has the disadvantage of essentially ignoring our uncertainty about the MDP.
\fi

\ifdefined \Neurips
\subsection{A Monte-Carlo Approach to \nameref{method1}}
\else
\subsection{A Monte-Carlo Approach to Method 1}
\fi
\label{sec:cond-mdp-distr}
We will now detail such a Monte-Carlo approach for Method 1.
We first combine the induction step in~\eqref{eq:inferential-induction-compact} and marginalisation of Method 1 in~\eqref{eq:method1}. We also substitute an approximate representation $\vbeltn$ for the next-step belief $\Pr(\Vt \mid D)$, to obtain the following conditional probability measure on value functions:
\ifdefined \Neurips
\begin{align*}
 \vbelt (B) \triangleq \Pr_\bel^\pol(\Vt \in B | D)
  &= \int_\CV \int_\MDPs \ind{\Bellman^\pol_\mdp \Vtn \in B} \dd \Pr^\pol_\bel(\mdp | \Vtn, D) \dd \vbeltn(\Vtn)
\end{align*}
\else
\begin{align*}
	&\vbelt (B) \triangleq \Pr_\bel^\pol(\Vt \in B | D)\\
	&= \int_\CV \int_\MDPs \ind{\Bellman^\pol_\mdp \Vtn \in B} \dd \Pr^\pol_\bel(\mdp | \Vtn, D) \dd \vbeltn(\Vtn)
\end{align*}
\fi
\ifdefined \LongVersion
Following Monte Carlo approach, we can estimate the outer integral as the sample mean over the samples value functions $\Vtn$.
\begin{equation}
  \vbelt (B) \approx \frac{1}{\nofVals} \sum_{k=1}^{\nofVals} \int_{\MDPs} \ind{\Bellman^\pol_\mdp \Vktn \in B} \dd \Pr^\pol_\bel(\mdp | \Vktn, D).
  \label{eq:monte-carlo-linkage}
\end{equation}
Here, $\nofVals$ is the number $\Vtn$ samples.
\fi

Let us focus on calculating $\Pr^\pol_\bel(\mdp \mid \Vtn, D)$. Expanding
it, we obtain, for any subset of MDPs $A \subseteq \MDPs$, the following measure:
\begin{equation}
\begin{aligned}
\Pr^\pol_\bel(\mdp \in A \mid \Vtn,D)
= \frac{\int_A \Pr_\mdp^\pol(\Vtn) \dd \bel(\mdp \mid D)} {\int_\MDPs \Pr_\mdp^\pol(\Vtn) \dd \bel(\mdp \mid D)},
\end{aligned}
\label{eq:mdp-given-V-D}
\end{equation}
since $\Pr^\pol_\mdp(\Vtn \mid  D) = \Pr_\mdp^\pol(\Vtn)$, as $\mdp, \pol$ are sufficient for calculating $\Vtn$.

To compute $\Pr_\mdp^\pol(\Vtn)$,
we can marginalise over utility rollouts $\util$ and states:
\begin{equation*}
  \Pr^\pol_\mdp(\Vtn)
  = \int_\CS \dd q(s) \int_{-\infty}^\infty \Pr^\pol_\mdp(\Vtn \mid \util, s) \Pr^\pol_\mdp(\util |s) \dd \util.
  \label{eq:marginal-Vtn}
\end{equation*}
\ifdefined \LongVersion
The details of computing rollouts are in \Cref{sec:rollout}.
\fi
In order to understand the meaning of the term $\Pr^\pol_\mdp(\Vtn \mid \util, s)$, note that $\Vtn(s) = \E[\util \mid s_{i+1} = s]$. Thus, a rollout from state $s$ gives us partial information about the value function. Finally, the starting state distribution $q$ is used to measure the goodness-of-fit, similarly to e.g. fitted-Q iteration\footnote{As long as $q$ has full support over the state space, any choice should be fine. For discrete MDPs, we use a uniform distribution $q$ over states and sum over all of them, while we sample from $q$ in the continuous case.}.

As a design choice, we define the density of $\Vtn$ given a sample $u_m$ from state $s_m \sim q$ to be a Gaussian with variance $\sigma^2$
\ifdefined \LongVersion
: $$\frac{\dd}{\dd \lambda} \Pr(\Vtn \mid u_m, s_m) \defn \frac{1}{\sqrt{2\pi}} e^{-\frac{\left|\Vtn(s_m)-u_m \right|^2}{2\sigma^2}}.
$$
\else
.
\fi
\ifdefined \LongVersion
In practice, we can generate utility samples from the sampled MDP $\mdp$ and the policy $\pol$ from step $t+1$ onwards and re-use those samples for all starting times $i > t$.

Finally, we can write:
\begin{align*}
    \Pr^\pol_\mdp(\Vtn)
  & \approx  \frac{1}{n} \sum^n_{m=1} \Pr^\pol_\mdp(\Vtn \mid u_m, s_m),
  & u_m &\sim \Pr^\pol_\mdp(\util)
          \label{eq:marginal-Vtn-MC}
\end{align*}
This leads to the following approximation for \eqref{eq:mdp-given-V-D}:
\begin{equation*}
  \begin{aligned}
     \Pr^\pol_\bel(\mdp \in A | \Vtn,D) 
     \approx \frac{\int_A \sum_{m} e^{-\frac{\left|\Vtn(s_m)-u_m \right|^2}{2\sigma^2}} \dd \bel(\mdp | D)} {\int_\MDPs \sum^n_{m} e^{-\frac{\left|\Vtn(s_m)-u_m \right|^2}{2\sigma^2}} \dd \bel(\mdp | D)}.
  \end{aligned}
\end{equation*}
\fi
If we generate $\nofMDPs$ number of MDPs $\mdp^{(j)} \sim \bel(\mdp \mid D)$ and set:
\begin{equation}
  w_{jk} \defn
   \frac{\sum^n_{m=1} e^{-\frac{\left|\Vktn(s_m)-u^j_m \right|^2}{2\sigma^2}}}
  {\sum_{j'=1}^{\nofMDPs} \sum^n_{m=1} e^{-\frac{\left|\Vktn(s_m)-u^{j'}_m \right|^2}{2\sigma^2}}},
  \label{eq:is-val-mdp}
\end{equation}
we get $\E w_{jk} = \Pr_\bel^\pol(\mdp \in M \mid \Vtn,D)$.
This allows us to obtain value function samples for step $i$,
\begin{equation}
\Vjkt \defn  \Bellman^\pol_{\mdp^{(j)}} \Vktn,
\label{eq:mc-current-value}
\end{equation}
each weighted by $w_{jk}$, leading to the following Monte Carlo estimate of the value function distribution at step $i$
\begin{equation}
  \label{eq:mc-belief-update}
  \vbelt (B) 
  = \frac{1}{\nofVals \nofMDPs} \sum_{k=1}^{\nofVals} \sum_{j=1}^{\nofMDPs} \ind{\Vjkt \in B} w_{jk}.
\end{equation}
\ifdefined \LongVersion
\
\else
Here, $\nofVals$ is the number value function samples $\Vktn$.
\fi 
This ends the general description of the Monte-Carlo method. Detailed design of an algorithm depends on the representation that we use for
$\vbelt$ and whether the MDP is discrete or continuous. 

\ifdefined \Neurips
Section~\ref{sec:algorithms} gives algorithmic details, while specifications of hyperparameters and distributions are given in Section~\ref{sec:experiments}.
\else 
\textbf{Bayesian Backwards Induction (BBI).} For instantiation, we construct a policy optimisation component and two approximate representations to use with the inferential induction based policy evaluation. 
For policy optimisation, we use a dynamic programming algorithm that looks ahead $\lookahead$ steps, and at each step $i$ calculates a policy maximising the Bayesian expected utility in the next $i+1$ steps. 
For approximate representation of the distribution of $\Vtn$, we use a multivariate Gaussian and $A$ multivariate Gaussians for discrete and continuous state $A$-action MDPs respectively.
We refer to this algorithm as Bayesian Backwards Induction (BBI) (\Cref{sec:algorithms}). Specifications of results, hyperparameters and distributions are in \Cref{sec:experiments} and~\Cref{sec:impl-deta}.
\fi

\ifdefined \LongVersion
\subsection{A Parenthesis on Mean-field Approximation} 
\label{sec:mean-field}
If we ignore the value function information by assuming that $\Pr^\pol_\bel(\mdp \mid \Vtn, D) = \Pr_\bel(\mdp \mid D)$, we obtain
\[\Pr^\pol_\bel(\Vt | D)
 =
\int_{\CV}\int_{\MDPs} \Pr^\pol_\mdp(\Vt \mid \Vtn) \dd \bel(\mdp \mid D) \dd \Pr^\pol_\bel(\Vtn \mid D).\]
Unfortunately, this corresponds to a mean-field approximation. For example, deploying similar methodology as \Cref{sec:cond-mdp-distr} would lead us to
\[\vbelt (B) 
= \frac{1}{\nofVals \nofMDPs} \sum_{k=1}^{\nofVals} \sum_{j=1}^{\nofMDPs} \ind{\Vjkt \in B}.\]
This will eventually eliminate all the uncertainty about the correspondence between value function and underlying MDPs because it is equivalent to assuming the mean MDP obtained from the data $D$ is true. 
For that reason, we do not consider this approximation any further.
\fi


	\section{Algorithms}
\label{sec:algorithms}
Algorithm~\ref{alg:induction-pe} is a concise description of the Monte Carlo procedure that we develop. At each time step $t$, the algorithm is called with the prior and data $D$ collected so far, and it looks ahead up to some lookahead factor $\lookahead$~\footnote{When the horizon $\horizon$ is small, we can set $\lookahead = \horizon - t$.}.
We instantiate it below for discrete and continuous state spaces. 
\begin{algorithm}[h]
	\caption{Policy Evaluation with Method 1}\label{alg:induction-pe}
	\begin{algorithmic}[1]
		\State \textbf{Input:} Prior $\bel$, data $D$, lookahead $\lookahead$, discount $\discount$, policy $\pol$, $\nofMDPs, \nofVals$.
		\State Initialise $\vbel_\lookahead$.
		\State Sample $\hat{M} \defn \cset{\mdp^{(j)}}{j \in [\nofMDPs]}$ from $\bel(\mdp \mid D)$.
		\For{$i = \lookahead-1, \ldots, 1$}
		\State Sample $V^{(k)} \sim \vbeltn(\val)$ for $k \in [\nofVals]$.
		\State Generate $n$ utility samples $u_m$
		\State Calculate $w_{jk}$ from \eqref{eq:is-val-mdp} and $\Vjkt$ from \eqref{eq:mc-current-value}.
		\State Calculate $\vbelt$ from \eqref{eq:mc-belief-update}.
		\EndFor
		\State \textbf{return} $\cset{\vbelt}{i=1, \ldots, \lookahead}$
	\end{algorithmic}
\end{algorithm}

\paragraph{Discrete MDPs.} 
When the MDPs are discrete, the algorithm is straightforward. Then the belief $\bel(\mdp \mid D)$ admits a conjugate prior in the form of a Dirichlet-product for the transitions.
In that case, it is also possible to use a histogram representation for $\vbelt$, so that it can be calculated by simply adding weights to bins according to \eqref{eq:mc-belief-update}.

However, as a histogram representation is not convenient for a large number of states, we model using a Gaussian $\vbel_t$. 
In order to do this, we use the sample mean and covariance of the weighted value function samples $\Vjkt$:
\ifdefined \Neurips
\begin{equation}
  \label{eq:guassian-belief-update}
    \bm_i = \frac{1}{\nofVals \nofMDPs} \sum_{k=1}^{\nofVals} \sum_{j=1}^{\nofMDPs} \Vjkt w_{jk},
    \bS_i = \frac{1}{\nofVals \nofMDPs} \sum_{k=1}^{\nofVals} \sum_{j=1}^{\nofMDPs} (\Vjkt - \bm_i) (\Vjkt - \bm_i)^\top  w_{jk}.
\end{equation}
\else
\begin{equation}
\begin{aligned}
\label{eq:guassian-belief-update}
\bm_i& = \frac{1}{\nofVals \nofMDPs} \sum_{k=1}^{\nofVals} \sum_{j=1}^{\nofMDPs} \Vjkt w_{jk}\\
\bS_i &= \frac{1}{\nofVals \nofMDPs} \sum_{k=1}^{\nofVals} \sum_{j=1}^{\nofMDPs} (\Vjkt - \bm_i) (\Vjkt - \bm_i)^\top  w_{jk}.
\end{aligned}
\end{equation}
\fi
such that $\vbelt = \Normal(\bm_i, \bS_i)$ is a multivariate normal distribution. 

\paragraph{Continuous MDPs.}In the continuous state case, we obtain $\vbel$ through fitted Q-iteration~\citep[c.f.][]{ernst2005tree}. 
For each action $a$ in a finite set, we fit a weighted linear model $Q_i(s_i, a) = s_i^T \omega_a + \epsilon_a$, where $s_i, \omega_a \in \mathbb{R}^d$ and $\epsilon_a \sim\Normal(0,\sigma_a^2)$.  Finding the representation $\omega_a$ is equivalent to solving $A$ weighted linear regression problems over $\nofVals$ state and Q-value samples for each action $a$:

\begin{align*}
	\omega_a &\triangleq \argmin_{\omega} \sum_{k=1}^{\nofVals}\sum_{j=1}^{\nofMDPs}w_{jk}\Big(Q^{(j,k)}_i(a) - (s^{(j,k)}_i)^T \omega\Big)^2 + \lambda \|\omega\|^2\\
	&= \argmin_{\omega} \|W^{1/2}(Q_i(a) - S_i^T(\omega))\|^2 + \lambda \|\omega\|^2.
\end{align*} 

Here, $W$ is the diagonal weight matrix. $S_i$ and $Q_i(a)$ are the $\nofVals \nofMDPs \times d$ matrices for the states and Q-values corresponding to sampled states and Q-values. We add an $l_2$ regulariser $\lambda \|\omega\|^2$ for efficient regression. We obtain $\omega_a = (S_i^TWS_i + \lambda \eye)^{-1}S_i^TWQ_i(a)$.
This is equivalent to estimating a multivariate normal distribution of Q-values $\vbelt(a) = \Normal(\bm^a_i, \bS^a_i)$, where
\begin{equation}
\begin{aligned}
\label{eq:cont-belief-update}\hspace*{-.5em}
\bm^a_i &= \frac{1}{\nofVals \nofMDPs} \sum_{k=1}^{\nofVals} \sum_{j=1}^{\nofMDPs} (s_i^{(j,k)})^T\omega_a\\
\bS^a_i &= \frac{\sigma_a^2}{\nofVals \nofMDPs} \sum_{k=1}^{\nofVals} \sum_{j=1}^{\nofMDPs} (Q^{(j,k)}_i(a) - \bm_i) (Q^{(j,k)}_i(a) - \bm_i)^\top  w_{jk}.
\end{aligned}
\end{equation}
In practice, we often use a feature map $\phi: \mathbb{R}^d \rightarrow \mathbb{R}^f$ for states. 

\subsection{Bayesian Backwards Induction}
\label{sec:bayes-backw-induct}
We now construct  a policy optimisation component to use with the inferential induction based policy evaluation and the aforementioned two approximation techniques. 
We use a dynamic programming algorithm that looks ahead $\lookahead$ steps, and at each step $i$ calculates a policy maximising the Bayesian expected utility in the next $i+1$ steps. 
We describe the corresponding pseudocode in \Cref{alg:bayesian-backwards-induction}.

\textbf{\Cref{alg:bayesian-backwards-induction} Line~\ref{line:approx_rep}: Discrete MDPs.} Just as in standard backwards induction, at each step, we can calculate $\pol_i$ 
by keeping $\pol_{i+1}, \ldots, \pol_{\lookahead}$ fixed:
\ifdefined \Neurips
\begin{multline}
  \Qb_i(s,a)
  \defn  \E_\bel(\util \mid s_i = s, a_i = a, D)
  =  \int_\MDPs
    \rew_{\mdp}(s,a)
    +
    \sum_{s'}
    \Pr_\mdp^{(j)}(s'|s,a) V_{\mdp,i}^{\pol_{i+1}, \ldots, \pol_{\lookahead}}(s') 
    \\
  \approx \sum_{j,k} [
    \rew_{\mdp^{(j)}}(s,a)
+
\sum_{s'}
\Pr_\mdp^{(j)}(s'|s,a) \Vktn(s')]  \frac{w_{jk}}{\nofMDPs \nofVals}.
\label{eq:bi-max}
\end{multline}
\else
\begin{multline}
	\Qb_i(s,a)
	\defn  \E_\bel(\util \mid s_i = s, a_i = a, D) \\
	=  \int_\MDPs
	\rew_{\mdp}(s,a)
	+
	\sum_{s'}
	\Pr_\mdp^{(j)}(s'|s,a) V_{\mdp,i}^{\pol_{i+1}, \ldots, \pol_{\lookahead}}(s') 
	\\
	\approx \sum_{j,k} [
	\rew_{\mdp^{(j)}}(s,a)
	+
	\sum_{s'}
	\Pr_\mdp^{(j)}(s'|s,a) \Vktn(s')]  \frac{w_{jk}}{\nofMDPs \nofVals}.
	\label{eq:bi-max}
\end{multline}
\fi

\textbf{\Cref{alg:bayesian-backwards-induction} Line~\ref{line:approx_rep}: Continuous MDPs.} As we are using fitted Q-iteration, we can directly use the state-action value estimates. So we simply set $\Qb_i(s,a) = \hat{Q}_i(s,a)$. 

The $\Qb_i$ estimate is then used to select actions for every state.
We set $\pol_i(a | s) = 1$ for $a = \argmax \Qb_i(s,a)$ (Line~\algref{alg:bayesian-backwards-induction}{line:bbi:pol}) and calculate the value function distribution (Lines~\algref{alg:bayesian-backwards-induction}{alg:bbi:weights} and~\algref{alg:bayesian-backwards-induction}{alg:bbi:next-belief}) for the partial policy $(\pol_i, \pol_{i+1}, \ldots, \pol_\lookahead)$.

\begin{algorithm}
  \caption{Bayesian Backwards Induction (BBI) with Method 1}\label{alg:bayesian-backwards-induction}
  \begin{algorithmic}[1]
    \State \textbf{Input:} Prior $\bel$, data $D$, lookahead $\lookahead$, discount $\discount$, $\nofMDPs, \nofVals$.
    \State Initialise $\vbelt$.
    \State Sample $\hat{M} \defn \cset{\mdp^{(j)}}{j \in [\nofMDPs]}$ from $\bel(\mdp \mid D)$.
    \For {$i = \lookahead-1, \ldots, 1$}
    \State Sample $V^{(k)} \sim \vbeltn(\val)$ for $k \in [\nofVals]$.
    \State Generate $n$ utility samples $u_i$
    \State Calculate $w_{jk}$ from \eqref{eq:is-val-mdp}.
    \State Calculate $\Qb_i$ from \eqref{eq:bi-max} or fitted Q-iteration.\label{line:approx_rep}
    \State Set $\pol_i(a | s) = 1$ for $a \in \argmax \Qb_i(s,a)$. \label{line:bbi:pol}
    \State Calculate $w_{jk}$ from \eqref{eq:is-val-mdp} and $\Vjkt$ from \eqref{eq:mc-current-value}
    with policy $\pol_i$:
	$\Vjkt \defn  \Bellman^{\pol_i}_{\mdp^{(j)}} \Vktn$.
	\label{alg:bbi:weights}
	\State Calculate $\vbelt$ from \eqref{eq:guassian-belief-update} or \eqref{eq:cont-belief-update}.
	\label{alg:bbi:next-belief}
    \EndFor
    \State \textbf{return} $\pol = (\pol_1, \ldots, \pol_\lookahead)$.
\end{algorithmic}
\end{algorithm}

	\section{Experimental Analysis}
\label{sec:experiments}
For performance evaluation, we compare Bayesian Backwards Induction (BBI, \Cref{alg:bayesian-backwards-induction}) with exploration by distributional reinforcement learning ~\citep[VDQN,][]{tang2018exploration}. We also compare BBI with posterior sampling ~\citep[PSRL,][]{strens2000bayesian,thompson1933lou}, MMBI~\citep{dimitrakakis:mmbi:ewrl:2011}, BSS~\citep{wang2005bayesian} and BQL~\cite{dearden1998bayesian} for the discrete MDPs and with Gaussian process temporal difference~\citep[GPTD,][]{engel2003bayes} for the continuous MDPs. In \Cref{sec:setup}, we describe the experimental setup and the priors used for implementation. In \Cref{sec:environments}, we illustrate different environments used for empirical evaluation. In \Cref{sec:results}, we analyse the results obtained for different environments in terms of average reward obtained over time. 

\subsection{Experimental Setup}~\label{sec:setup}
\textbf{Parameters.} We run the algorithms for the infinite-horizon formulation of value function with discount factor $\discount = 0.99$. We evaluate their performance in terms of the evolution of average reward to $\horizon = 10^6$ and $10^5$ time-steps for discrete and continuous MDPs respectively . 
Each algorithm updates its policy at steps $t = 1, 3, 6, 10, \ldots$.
We set $\lookahead$ to $100$ and $20$ for discrete and continuous MDPs respectively. 
More implementation details can be found in the supplementary material.

\textbf{Prior.} For discrete MDPs, we use Dirichlet $Dir(\alpha)$ priors over each of the transition probabilities $\Pr(s' | s, a)$. The prior parameter $\alpha$ for each transition is set to $0.5$. We use separate NormalGamma $\mathcal{NG}(\mu, \kappa, \alpha, \beta)$ priors for each of the reward distributions $\Pr(r | s, a)$. We set the prior parameters to $[\mu_0, \kappa_0, \alpha_0, \beta_0] = [0, 1, 1, 1]$. While we use the same prior parameters for all algorithms, we have not attempted to do an exhaustive unbiased evaluation by tuning their hyperparameters on a small set of runs, hence, our results should be considered preliminary.

For continuous MDPs, we use factored Bayesian Multivariate Regression \citep{minka:regression} models as priors over transition kernels and reward functions for the continuous environments. This implies that the transition kernel $\Pr(s' | s, a)$ and reward kernel $\Pr(r | s, a)$  modelled as $\mathcal{N}(A^{\mathrm{Trans}}_a s, \Sigma)$ and $\mathcal{N}(A^{\mathrm{Reward}}_a s, \sigma^2)$.
$\Sigma$ is sampled from inverse Wishart distribution with corresponding $d \times d$ dimensional scale matrix, while $\sigma$ is sampled from inverse Gamma with prior parameters $(\frac{1}{2}, \frac{1}{2})$. For transitions, we set the prior parameters to $\matrixsym{\Psi}_0 = 0.001 \eye$ and degrees of freedom $\nu_0 = rank(\matrixsym{\Psi}_0)$. 

For the InvertedPendulum, we use Bayesian multivariate regressor priors on $\Pr(s' \, | \, \phi(s), a)$ and $\Pr(r \, | \, \phi(s), a)$, where the feature map $\phi : \mathbb{R}^d \rightarrow \mathbb{R}^f$ is given by the mentioned basis functions. GPTD uses the same feature map as BBI, while VDQN only sees the actual underlying state $s$. The choice of state distribution $q(s)$ is of utmost importance in continuous environments. In this environment, we experimented with a few options, trading of sampling states from our history, sampling from the starting configuration of the environment and sampling from the full support of the state space.

\subsection{Description of Environments}~\label{sec:environments}
We evaluate the algorithms on four discrete and one continuous environments.

\textbf{NChain.} This is a discrete stochastic MDP with 5 states, 2 actions~\citep{strens2000bayesian}.
Taking the first action returns a reward $2$ for all states and transitioning to the first state. Taking the second action returns $0$ reward in the first four states (and the state increases by one) but returns $10$ for the fifth state and the state remains unchained.  There is a probability of slipping of $0.2$ with which its action has the opposite effect. This environment requires both exploration and planning to be solved effectively and thus acts as an evaluator of posterior estimation, efficient performance and effective exploration.

\textbf{DoubleLoop.} This is a slightly more complex discrete deterministic MDP with with two loops of states~\citep{strens2000bayesian}.
Taking the first action yields traversal of the right loop and a reward $1$ for every $5$ state traversal. Taking the second action yields traversal of the left loop and a reward $2$ for every $5$ state traversal. This environment acts as an evaluator of efficient performance and effective exploration.

\textbf{LavaLake.} This is a stochastic grid world~\citep{leike2017ai} where every state gives a reward of -1, unless you reach the goal, in which case you get 50, or fall into lava, where you get -50. 
We tested on the $5 \times 7$ and a $10 \times 10$ versions of the environment.
The agent moves in the direction of the action (up,down,left,right) with probability 0.8 and with probability 0.2 in a direction perpendicular to the action.

\textbf{Maze.} This is a grid world with four actions (ref. Fig. 3 in~\citep{strens2000bayesian}). The agent must obtain 3 flags and reach a goal.
There are 3 flags throughout the maze and upon reaching the goal state the agent obtains a reward of 1 for each flag it has collected and the environment is reset. Similar to LavaLake, the agent moves with probability 0.9 in the desired direction and 0.1 in one of the perpendicular directions.
The maze has 33 reachable locations and 8 combination of obtained flags for a total of 264 states.

\textbf{LinearModel.} This is a continuous MDP environment consisting of $4$ state dimensions and $11$ actions. The transitions and rewards are generated from a linear model of the form $s_{t+1} = A_a^{\CS}s_t, r_t = (A_a^{\CR})^Ts_t$ where $s, A_a^{\CR} \in \mathbb{R}^4$ and $A_a^{\CS}\in \mathbb{R}^{4\times 4}$ for all $a$'s. 

\textbf{InvertedPendulum.}
To extend our results for the continuous domain we evaluated our algorithm in a classical environment described in~\citep{lagoudakis2003least}. The goal of the environment is to stabilize a pendulum and to keep it from falling. If the pendulum angle $\theta$ falls outside $[\frac{-\pi}{2}, \frac{\pi}{2}]$ then the episode is terminated and the pendulum returned to its starting configuration. The state dimensionality is a tuple of the pendulum angle as well as its angular velocity, $\dot{\theta}$, $s = (\theta, \dot{\theta})$. The environment is considered to be completed when the pendulum has been kept within the accepted range for $3000$ steps. For further details, we refer to~\citep{lagoudakis2003least}.

We use the features recommended by~\cite{lagoudakis2003least}, which are $10$ basis functions that correspond to a constant term as well as $3 \times 3$ RBF kernels with $\sigma^2 = 1.0$ and
\[ \mu_{\theta, \dot{\theta}} = \left( \begin{array}{ccc}
(\frac{-\pi}{4}, -1) & (\frac{-\pi}{4}, 0)  & (\frac{-\pi}{4}, 1)  \\
(0, -1) & (0, 0)  & (0, 1)  \\
(\frac{\pi}{4}, -1) & (\frac{\pi}{4}, 0)  & (\frac{\pi}{4}, 1) \end{array} \right).\] 

We also add a regularizing term with $\lambda \eye$, $\lambda=0.01$ for stabilising the fitted Q-iteration.

\subsection{Experimental Results}
\label{sec:results}

The following experiments are intended to show that the general methodological idea is indeed sound, and can potentially lead to high performance algorithms.  


\begin{figure*}[t!]
	\begin{subfigure}[t]{0.5\textwidth}
		
		\includegraphics[width=\linewidth]{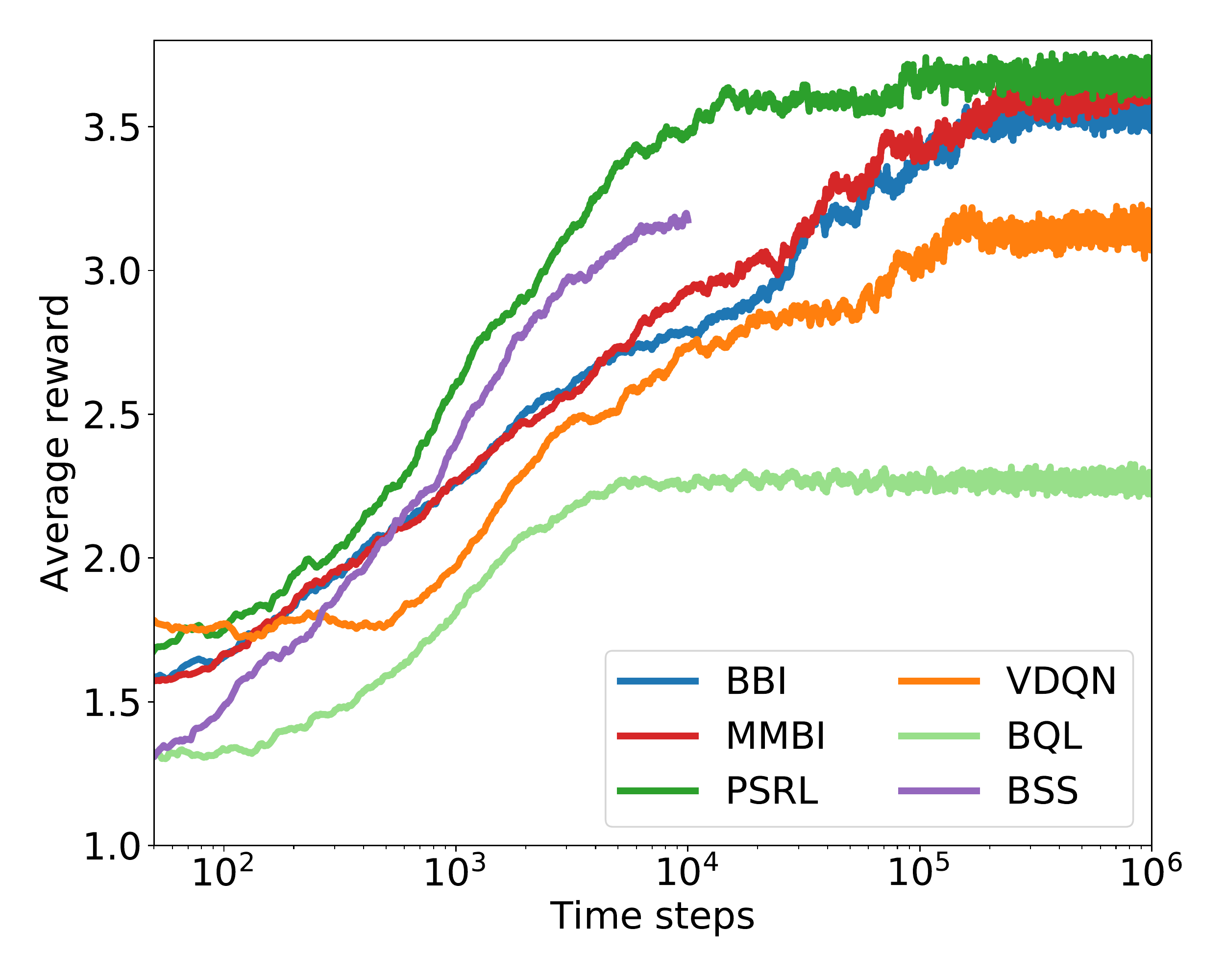}
		\caption{NChain}
		\label{fig:full_chain}
	\end{subfigure}
	\begin{subfigure}[t]{0.5\textwidth}
		\includegraphics[width=\linewidth]{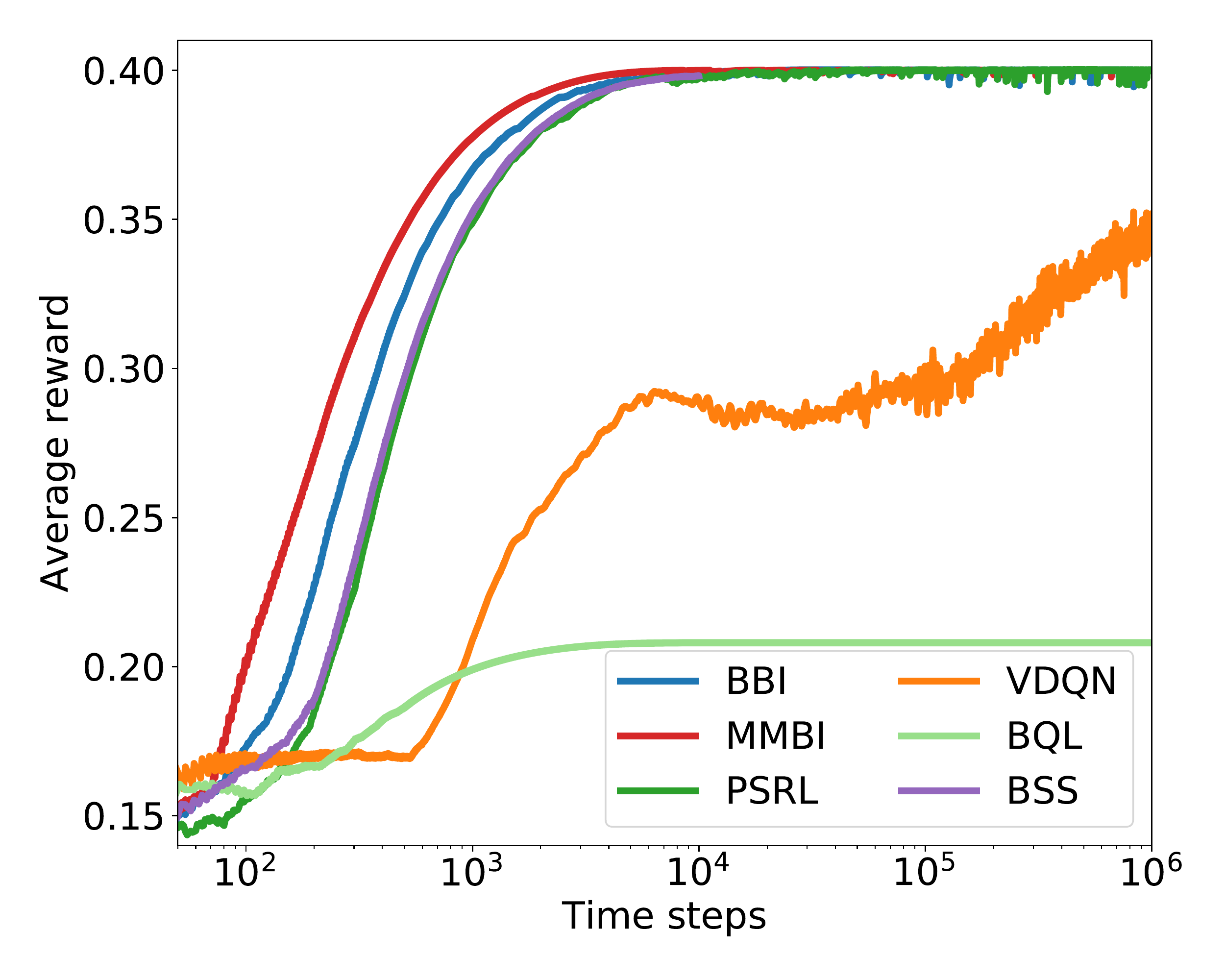}
		\caption{DoubleLoop}
		\label{fig:full_loop}
	\end{subfigure}
	\caption{Evolution of average reward for NChain and DoubleLoop environments, averaged over 50 runs of length $10^6$ for each algorithm. For computational reasons BSS is only run for $10^4$ steps. The runs are exponentially smoothened with a half-life $1000$ before averaging.}\label{fig:full_chainloop}
	
\end{figure*}
\begin{figure*}[h]
	\begin{subfigure}[t]{0.5\textwidth}
		\includegraphics[width=\linewidth]{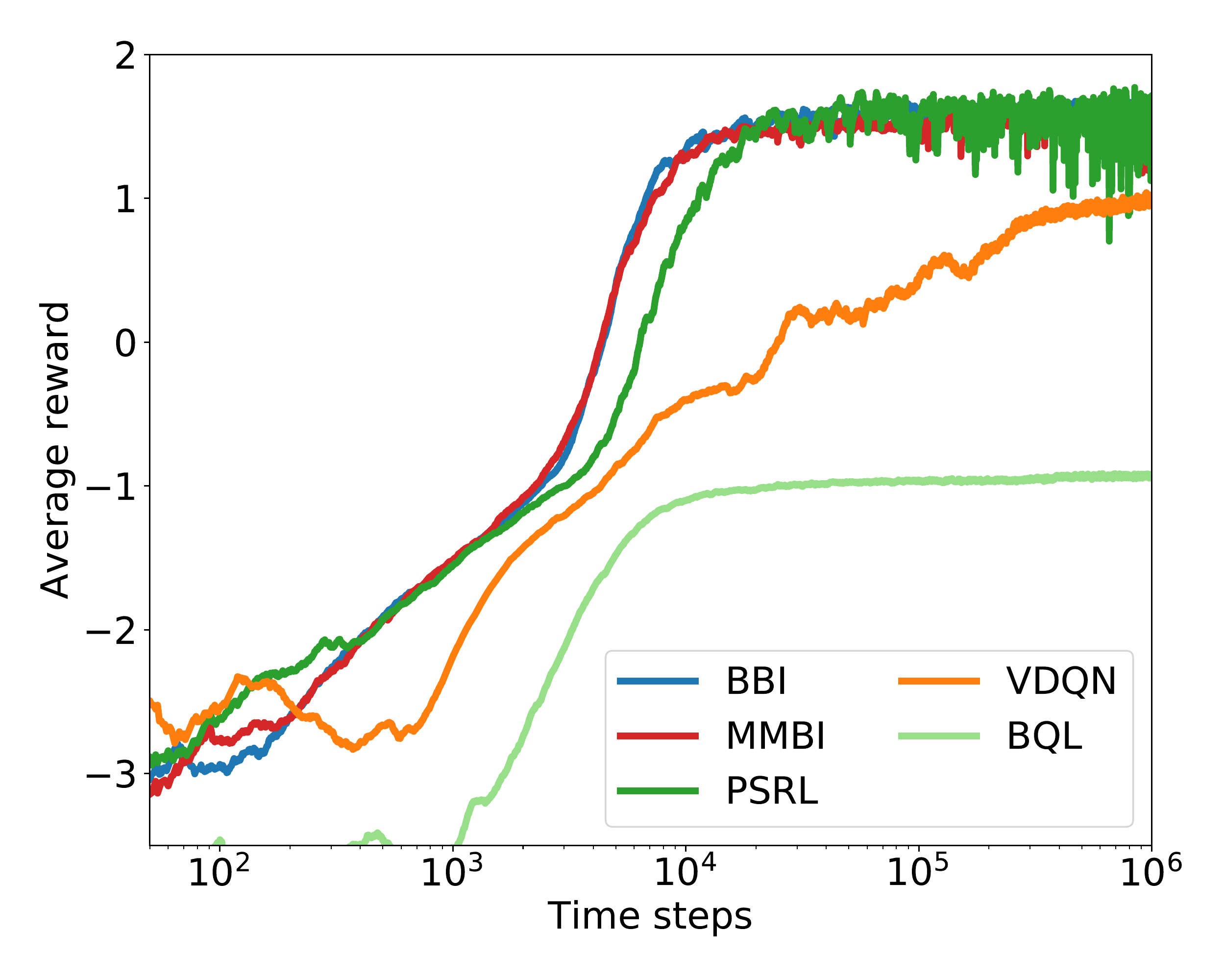}
		\caption{LavaLake $5\times 7$}
		\label{fig:full_lava_small}
	\end{subfigure}%
	\begin{subfigure}[t]{0.5\textwidth}
		\includegraphics[width=\linewidth]{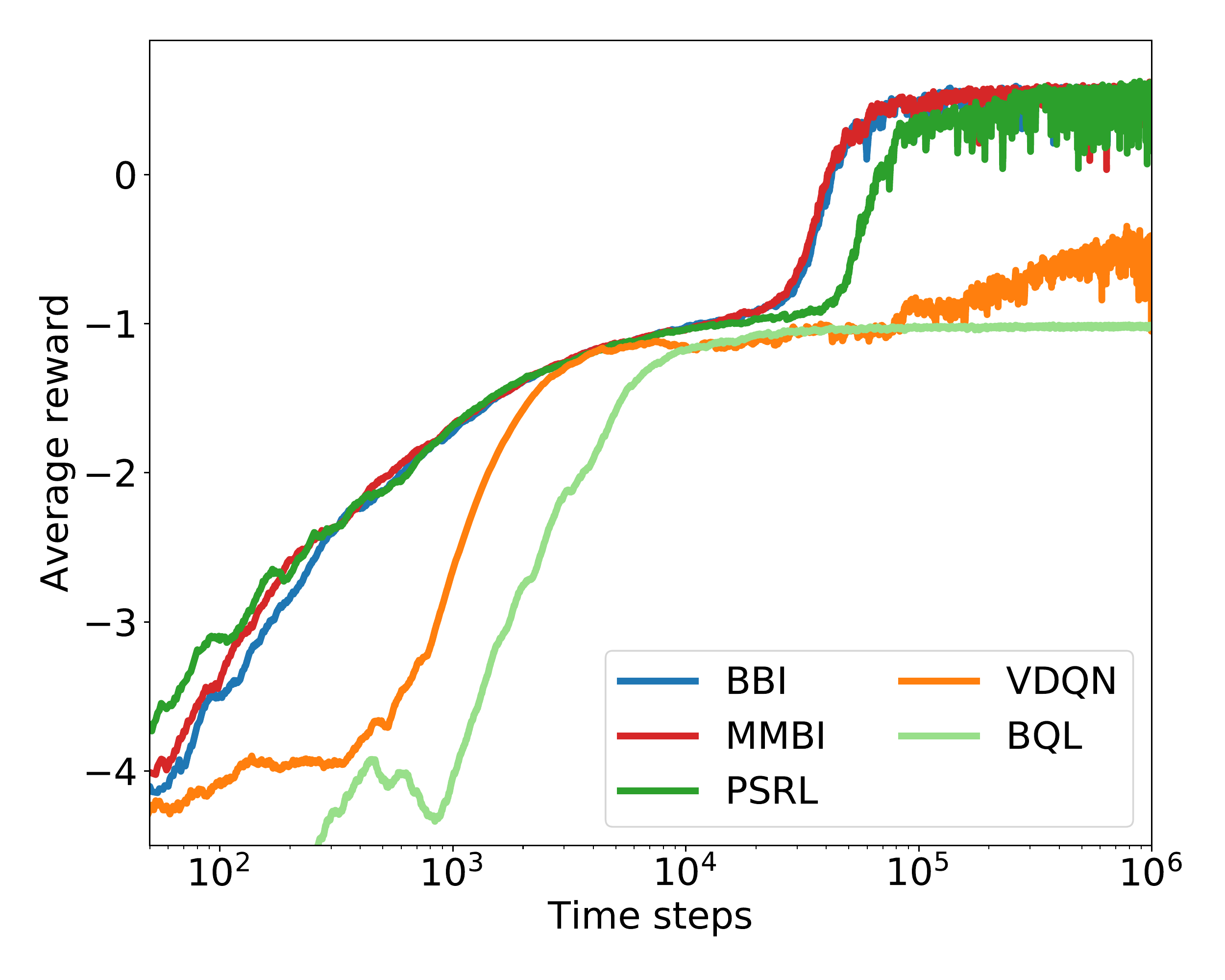}
		\caption{LavaLake $10\times 10$}
		\label{fig:full_lava_big}
	\end{subfigure}%
	
	\caption{Evolution of average reward for $5\times 7$ and $10\times10$ LavaLake environments. The results are averaged over 20 and 30 runs respectively with a length of $10^6$ for each algorithm.  The runs are exponentially smoothened with a half-life $1000$ before averaging.}
	\label{fig:full_lavaplots}
\end{figure*}

\begin{figure}[h]
	\centering
	\includegraphics[width=0.75\linewidth]{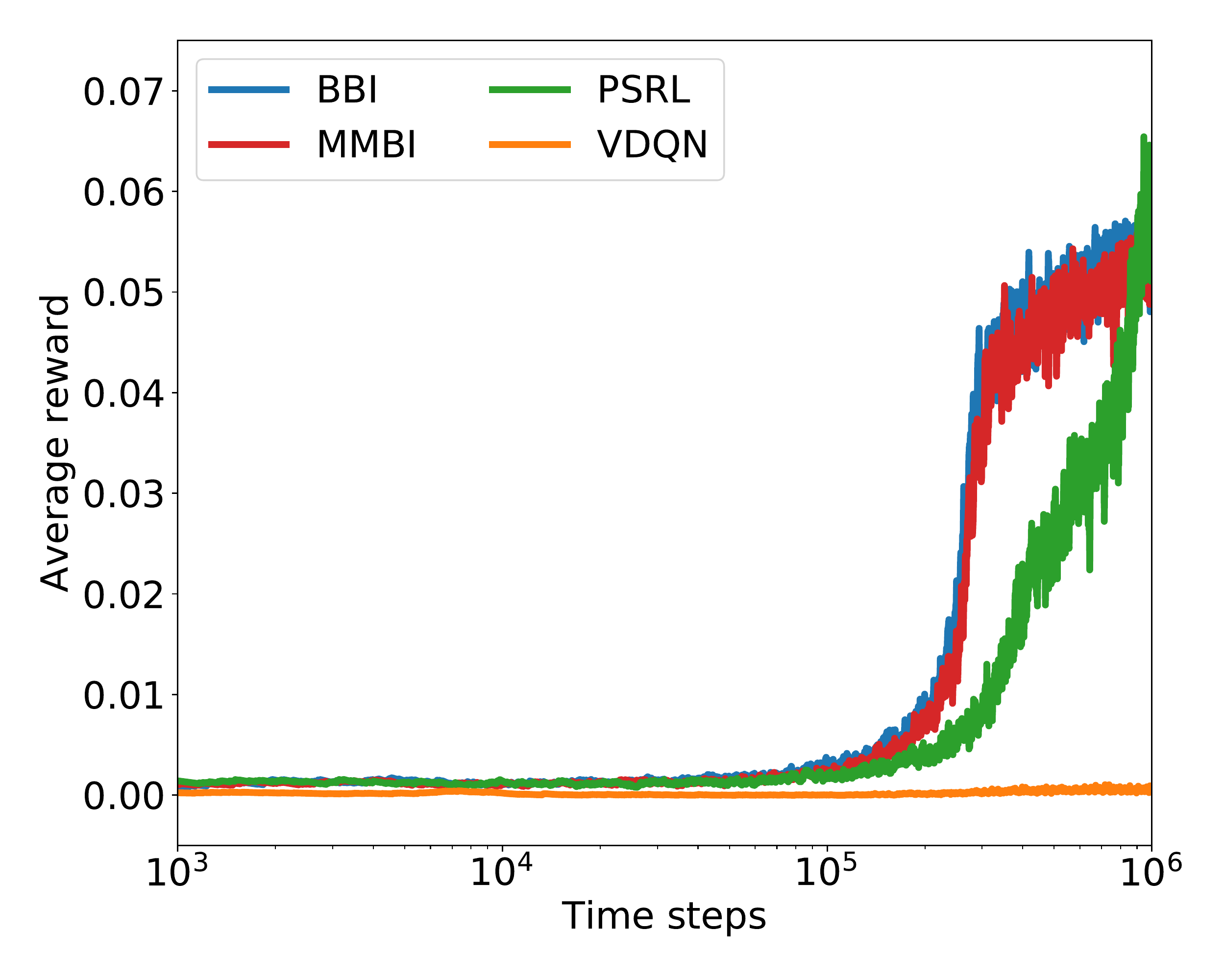}
	\caption{Evolution of average reward for the Maze environment. The results are averaged over 30 runs with a length of $10^6$ for each algorithm. The runs are exponentially smoothened with a half-life $1000$ before averaging.}
	\label{fig:full_maze}
\end{figure}

\begin{figure*}[t!]
	\begin{subfigure}[t]{0.48\textwidth}
		\includegraphics[width=\linewidth]{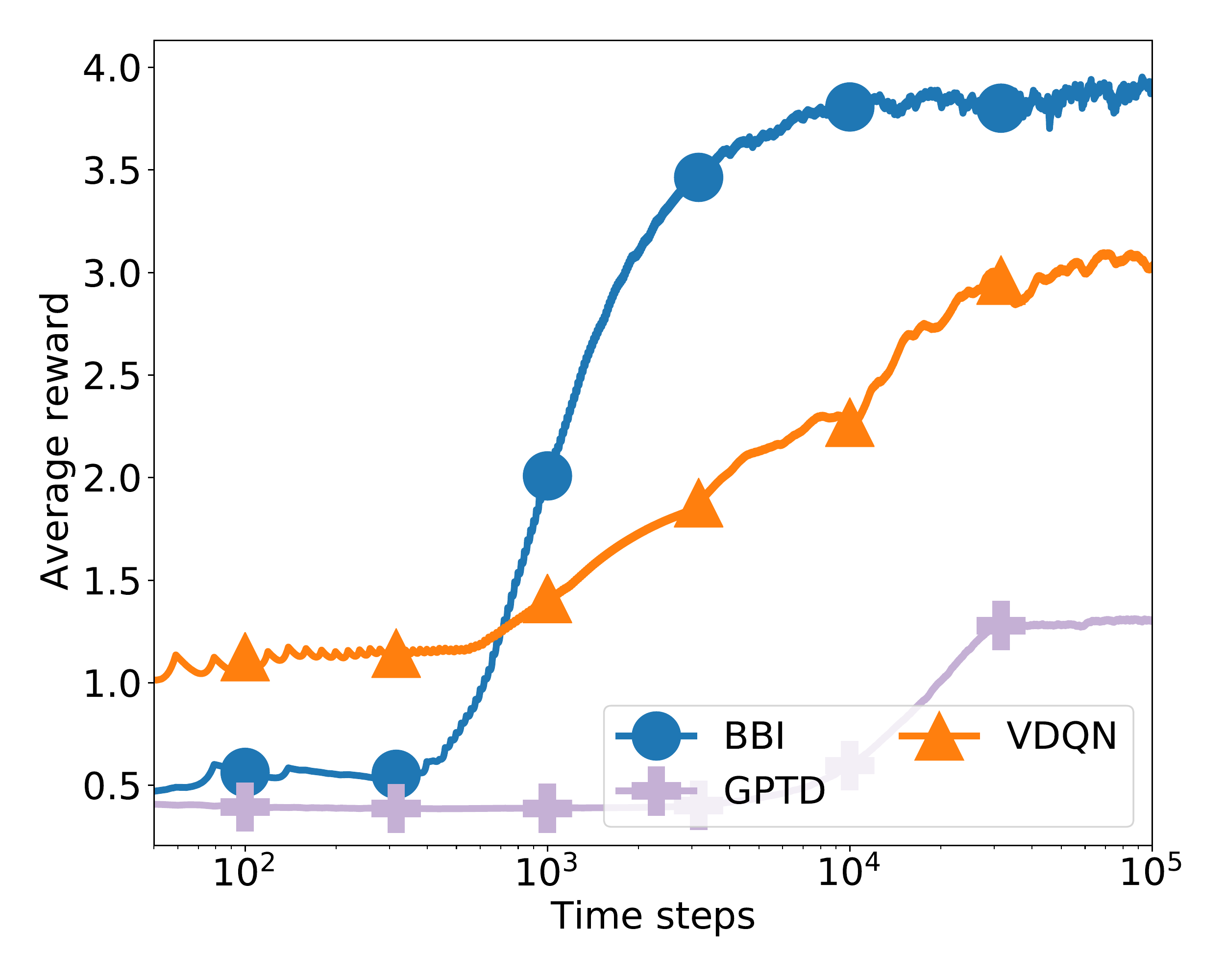}
		\caption{LinearModel}
		\label{fig:linear_model}
	\end{subfigure}%
	\begin{subfigure}[t]{0.48\textwidth}
		\includegraphics[width=\linewidth]{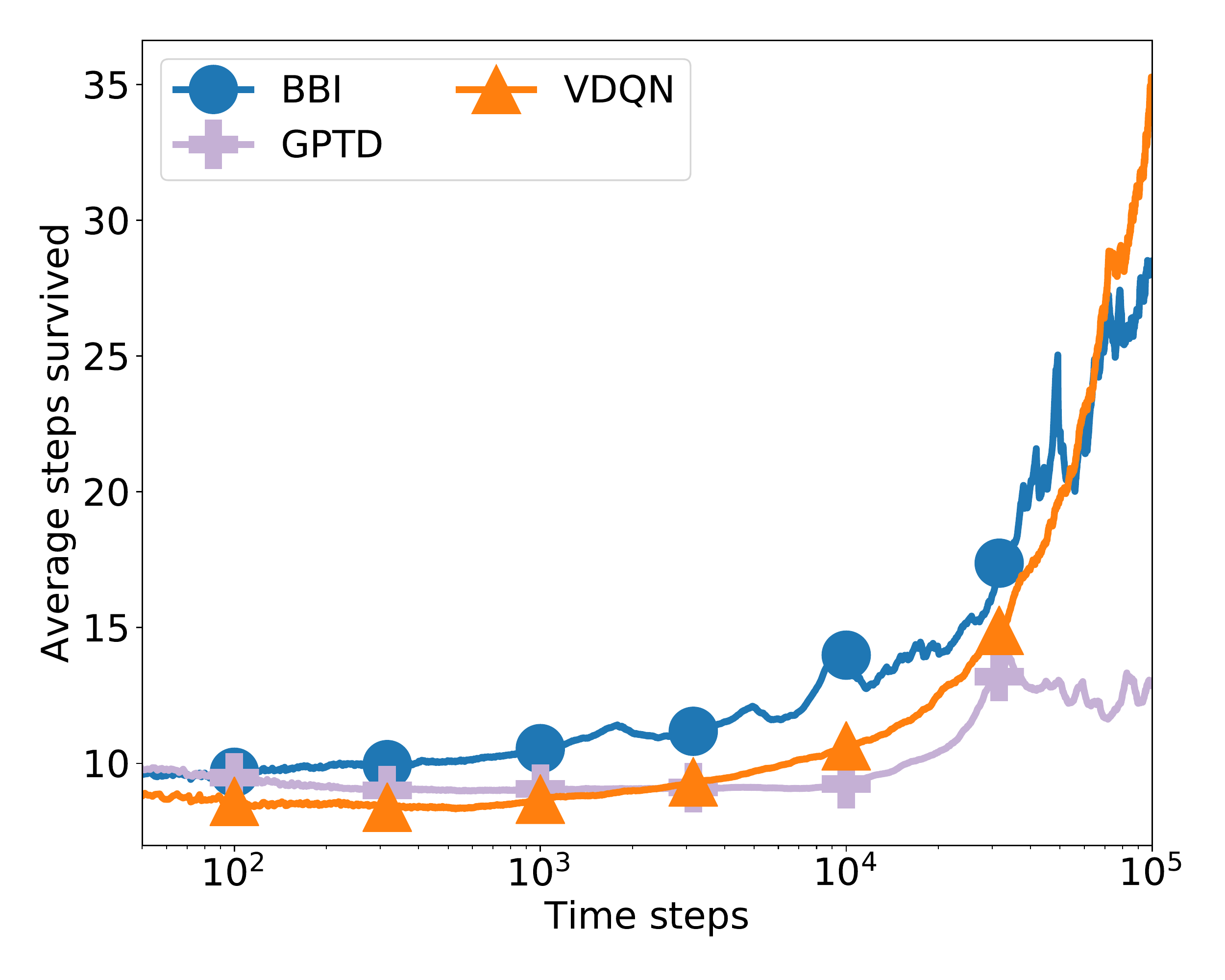}
		\caption{InvertedPendulum}
		\label{fig:inverted_pendulum}
	\end{subfigure}%
	\caption{Evolution of average steps survived during an episode for the LinearModel and InvertedPendulum environment, averaged over 100 and 30 runs respectively with runs of length $10^5$ for each algorithm. The runs are exponentially smoothened with a half-life $1000$ and $2500$ respectively before averaging.}
	\label{fig:continuous}
\end{figure*}

\Cref{fig:full_chain,fig:full_loop,fig:full_lava_big,fig:full_lava_small,fig:full_maze} illustrate the evolution of average reward for  BBI, PSRL,VDQN, MMBI, BQL and BSS on the discrete MDPs. BBI performs similarly to to MMBI and PSRL. This is to be expected, as the optimisation algorithm used in MMBI is close in spirit to BBI, with only the inference being different. In particular, this algorithm takes $k$ MDP samples from the posterior, and then performs backward induction in all the MDP simultaneously to obtain a Markov policy. In turn, PSRL can be seen as a special case of MMBI with just one sample. This indicates that the BBI inference procedure is sound. The near-optimal Bayesian approximation performs slightly worse in this setting, perhaps because it was not feasible to increase the planning horizon sufficiently.\footnote{For computational reasons we used a planning horizon of two with four next state samples and two reward samples in each branching step. We hope to be able to run further experiments with BSS at a later point.}
Finally, the less principled approximations, like VDQN and BQL do not manage to have a satisfactory performance in these environments.
 In \Cref{fig:linear_model,fig:inverted_pendulum}, we also compare with GPTD, a classical method for Bayesian value function estimation instead of PSRL. In \Cref{fig:linear_model} it is evident that GPTD cannot leverage its sophisticated Gaussian Process model to learn as well as BBI. The same is true for  VDQN, except for when the amount of data is very small.  \Cref{fig:inverted_pendulum} shows a comparison on the InvertedPendulum environment. Here our algorithm is competitive, and in particular performs much better than GPTD, while it performs similarly to VDQN, which is slightly worse initially and slightly better later in terms of average steps survived.  This performance could partially be explained by the use of a linear value function $Q(\phi(s), a)$, in contrast to VDQN which uses a neural network. We thus feel that further investment in our methodology is justified by our results.
 


	\section{Discussion and Future Work}
	\label{sec:conclusion}
	We offered a new perspective on Bayesian value function estimation. The central idea is to calculate \textit{the conditional value function distribution} $\Pr^\pol_\bel(V_i \mid V_{i+1}, D)$ using the data and to apply it inductively for computing \textit{the marginal value function distribution} $\Pr^\pol_\bel(V_i \mid D)$.
	Following this, we propose three possible marginalisation techniques (Methods 1, 2 and 3) and design a Monte-Carlo approximation for Method 1. We also combined this procedure with a suitable policy optimisation mechanism and showed that it can be competitive with the state of the art.
	
	Inferential Induction differs from existing Bayesian value function methods, which essentially cast the problem into regression. For example, GPTD~\citep{engel2003bayes} can be written as Bayesian inference with a GP prior
	over value functions and a data likelihood that uses a deterministic empirical model of the MDP. While this can be relaxed by using temporal correlations as in~\citep{Engel05}, the fundamental problem remains.
	Even though such methods have practical value, we show that Bayesian estimation of value functions requires us to explicitly think about the MDP distribution as well. 
	
	We use specific approximations for discrete and continuous MDPs to propose the Bayesian Backwards Induction (BBI) algorithm. Though we only developed one algorithm, BBI, from this family, our experimental results appear promising. We see that BBI is competitive with state-of-the art methods like PSRL, and it significantly outperforms the algorithms relying on approximate inference, such as VDQN. Thus, our proposed framework of inferential induction offers a new perspective, which can provide a basis for developing new Bayesian reinforcement learning algorithms.

	\section*{Acknowledgements} 
	Thank you to Nikolaos Tziortziotis for his useful discussions. This work was partially supported by the Wallenberg AI, Autonomous Systems and Software Program (WASP) funded by the Knut and Alice Wallenberg Foundation. The experiments were partly performed on resources at Chalmers Centre for Computational Science and Engineering (C3SE) provided by the Swedish National Infrastructure for Computing (SNIC).

	\bibliography{references} 
	\bibliographystyle{apalike}
	
	\newpage
	\appendix

	\section{Implementation Details}
\label{sec:impl-deta}

In this section we discuss some additional implementation details, in particular how exactly we performed the rollouts and the selection of some algorithm hyperparameters, as well as some sensitivity analysis.
\subsection{Computational Details of Rollouts}\label{sec:rollout}
To speed up the computation of rollouts, we have used three possible methods that essentially bootstrap previous rollouts or use value function samples:
\begin{equation}
u_t^{\mdp, \pi_t}(s) = r(s,a) + \discount u_{t+1}^{\mdp, \pi_{t+1}}(s')
\end{equation}
\begin{equation}
u_t^{\mdp, \pi_t}(s) = \sum_{s'} r(s,a) + \gamma P(s'|s,a) u_{t+1}^{\mdp, \pi_{t+1}}(s')
\end{equation}
\begin{equation}
u_t^{\mdp, \pi_t}(s) = \sum_{s'} r(s,a) + \gamma P(s'|s,a) V_{t+1}(s') \label{ours}
\end{equation}
 where $V_{t+1} \sim \psi_{t+1}$.
In experiments, we have found no significant difference between them. All results in the paper use the formulation in~\eqref{ours}.

\subsection{Hyperparameters}
\label{sec:hyperparameters}
For the experiments, we use the following hyperparameters.

We use 10 MDP samples, a planning horizon T of 100, $\gamma=0.99$ and we set the variance of the Gaussian to be $\sigma^2 = V_\text{span}^2 10^{-4}$, where $V_\text{span}$ is the span of possible values for each environment (obtained assuming maximum and minimum reward).
We use Eq. \ref{ours} for rollout computation with 10 samples from $V_{t+1}$ and 50 samples from $V_{t}$ (20 for LavaLake $10\times10$ and Maze).
If the weights obtained in~\eqref{eq:is-val-mdp} are numerically unstable we attempt to resample the value functions and then double $\sigma$ until it works (but is reset to original value when new data is obtained). This is usually only a problem when very little data has been obtained. 

\begin{figure*}[t!]
\begin{subfigure}[t]{0.5\textwidth}
	\includegraphics[width=\linewidth]{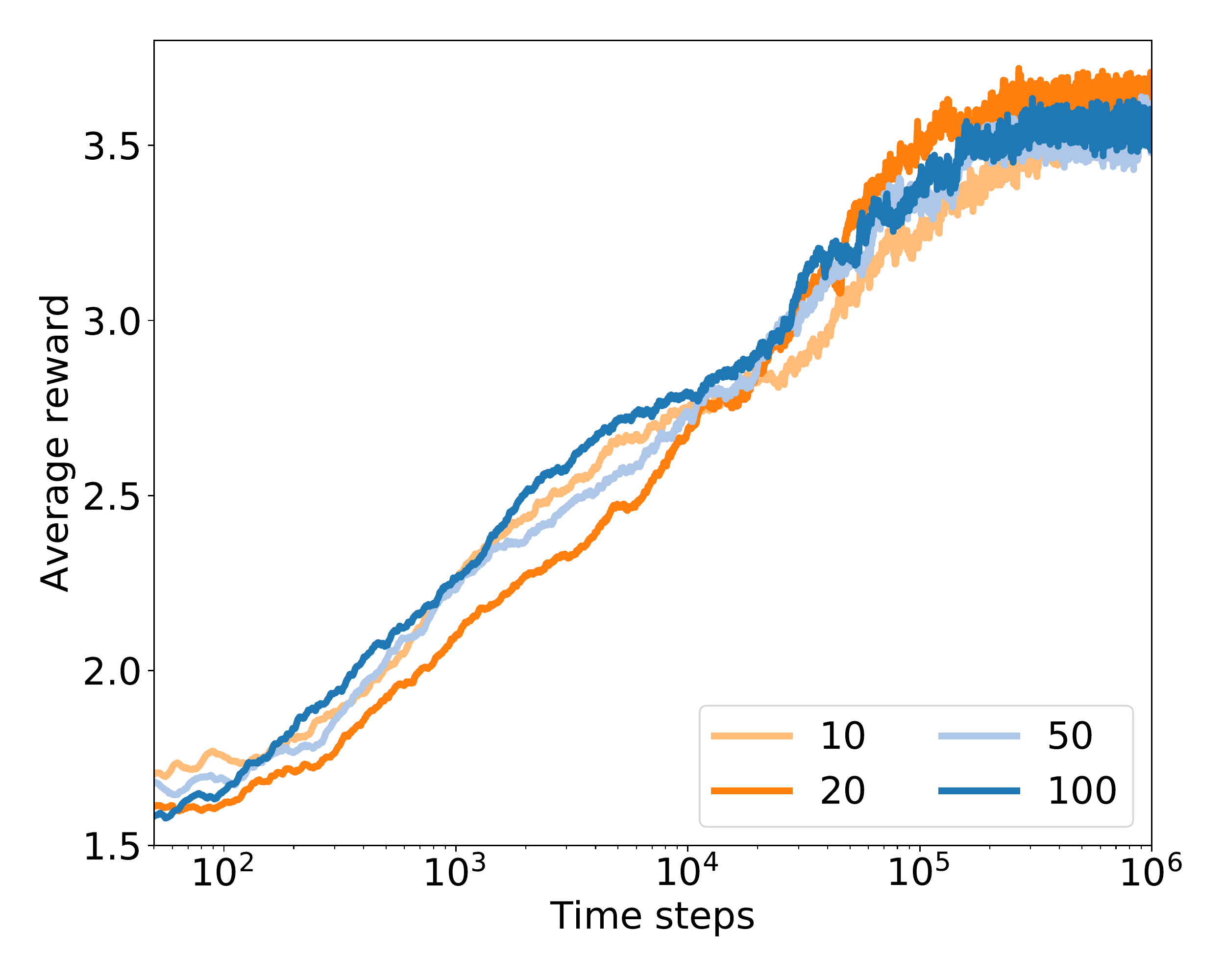}
	\caption{NChain}
	\label{fig:chainhorizon}
\end{subfigure}%
\begin{subfigure}[t]{0.5\textwidth}
	\includegraphics[width=\linewidth]{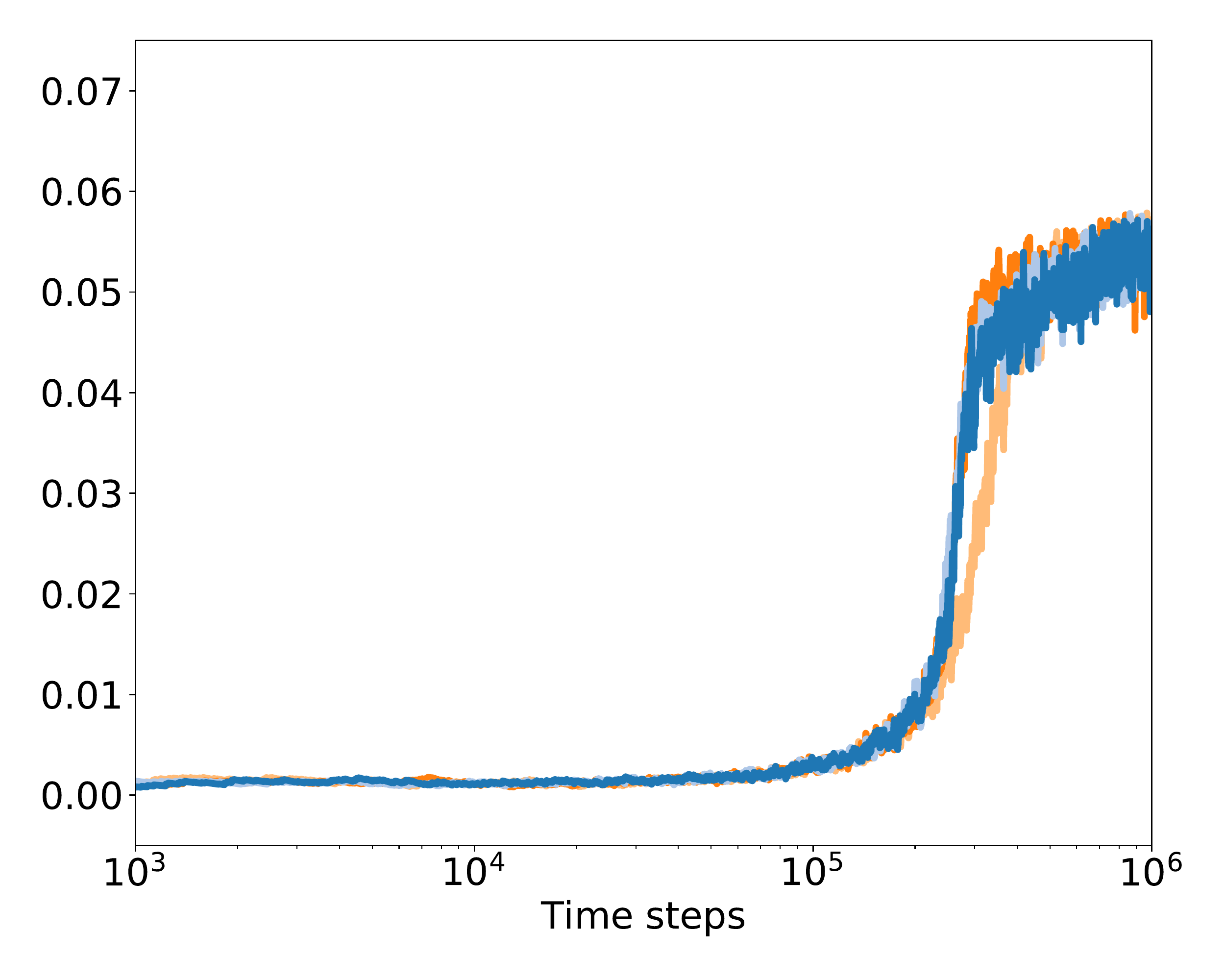}
	\caption{Maze}
	\label{fig:mazehorizon}
\end{subfigure}%
\caption{Illustration of the impact of varying the horizon T in BBI. The results are averaged over 50 and 30 runs respectively with a length of $10^6$ for each algorithm.  The runs are exponentially smoothened with a half-life $1000$ before averaging.}
\label{fig:horizon}
\end{figure*}

In order to check the sensitivity on the choice of horizon $T$, we perform a sensitivity analysis with $T= 10, 20, 50, 100$. In Figure~\ref{fig:horizon}, we can see that varying the horizon has a very small impact for NChain and Maze environments. 
	\section{Additional Results}
\label{sec:additional-results}
Here we present some experiments that examine the performance of inferential induction in terms of value function estimation, inference and utility obtained.

\subsection{Bayesian Value Function Estimation} 
\begin{figure}[t!]
		\centering
		\includegraphics[width=0.95\linewidth]{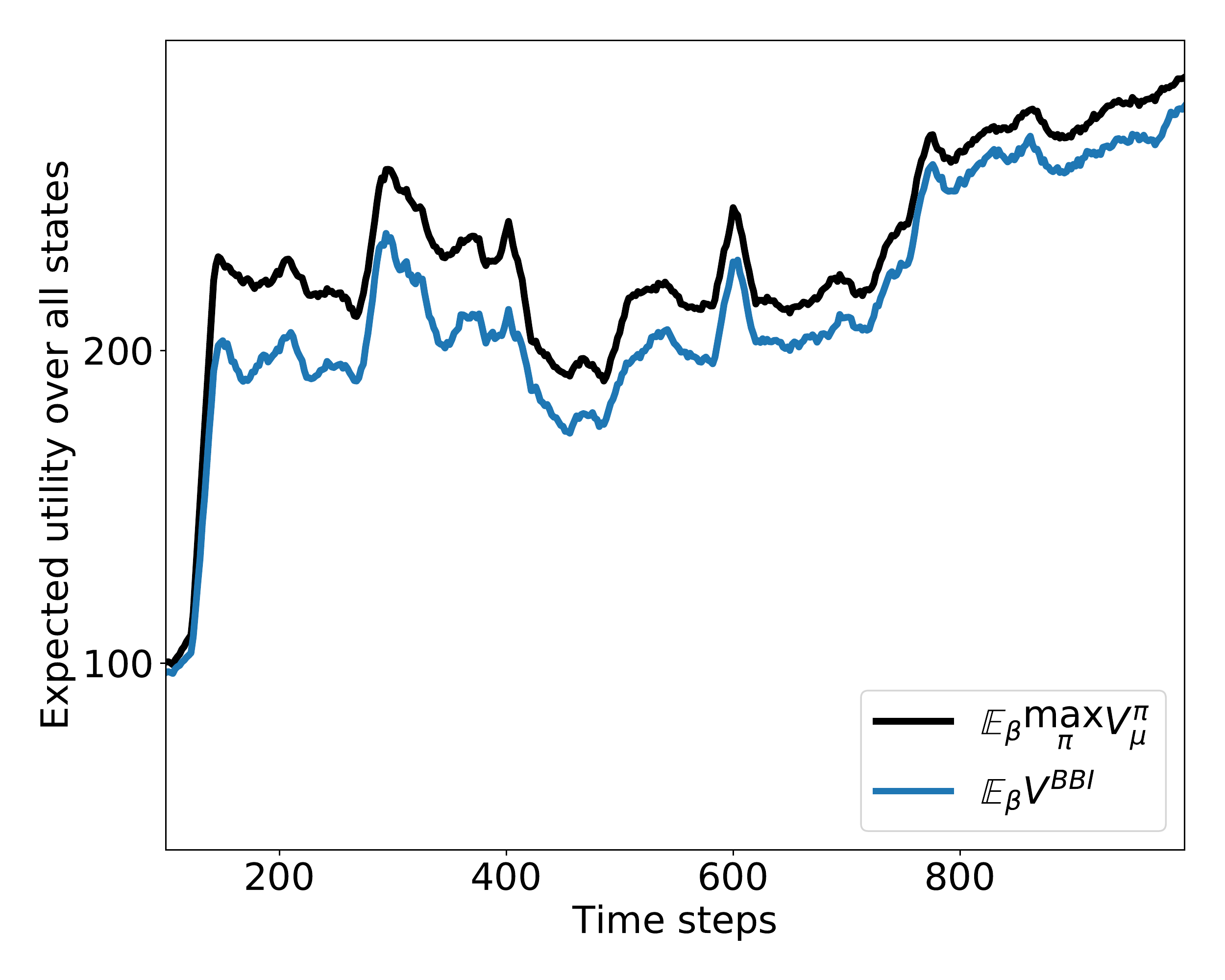}
		\caption{Comparisons of the achieved value functions of BBI with the upper bound on Bayes-optimal value functions. Upper bound and BBI are calculated from $100$ MDPs and plotted for $10^5$ time steps.}
		\label{fig:bayes_bounds}
\end{figure}

In this experiment, we evaluate the Bayesian (i.e. mean) value function of the proposed algorithm (BBI) with respect to the upper bound on the Bayes-optimal value function. 
The upper bound is calculated from $\int_\MDPs \max_\pol V_\mdp^\pol d\bel(\mdp \mid D)$. We estimate this bound through $100$ MDP samples for NChain. We plot the time evolution of our value function and the simulated Bayes bound in Figure~\ref{fig:bayes_bounds} for $10^5$ steps. We observe that this is becomes closer to the upper bound as we obtain more data.


\subsection{Value Function Distribution Estimation}
\begin{figure*}[ht!]
	\hspace{-1em}
	\begin{subfigure}[t]{0.33\linewidth}
		\includegraphics[width=\linewidth]{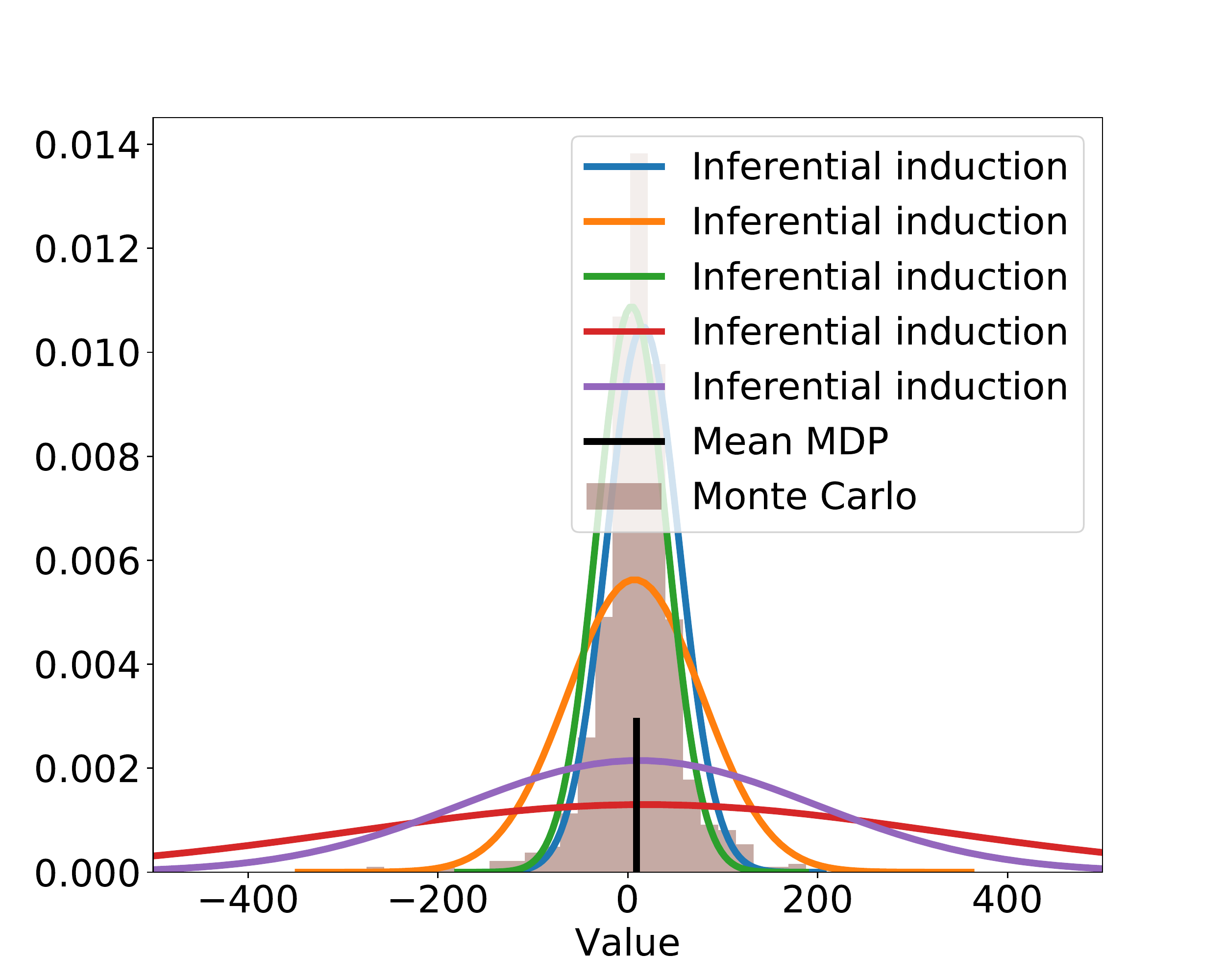}\caption{$T = 10$}
	\end{subfigure}
	\begin{subfigure}[t]{0.33\linewidth}
		\includegraphics[width=\linewidth]{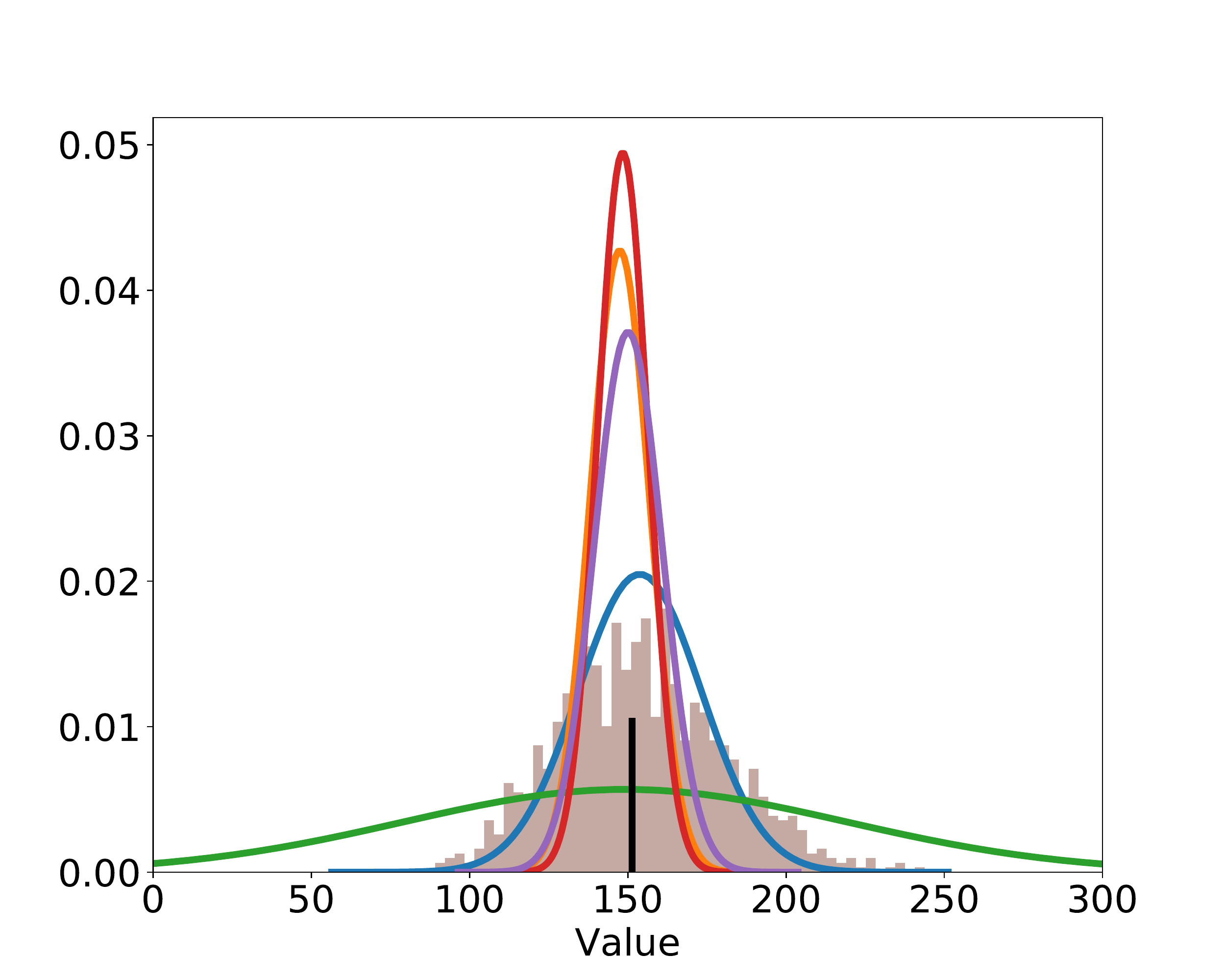}\caption{$T=100$}
	\end{subfigure}
	\begin{subfigure}[t]{0.33\linewidth}
		\includegraphics[width=\linewidth]{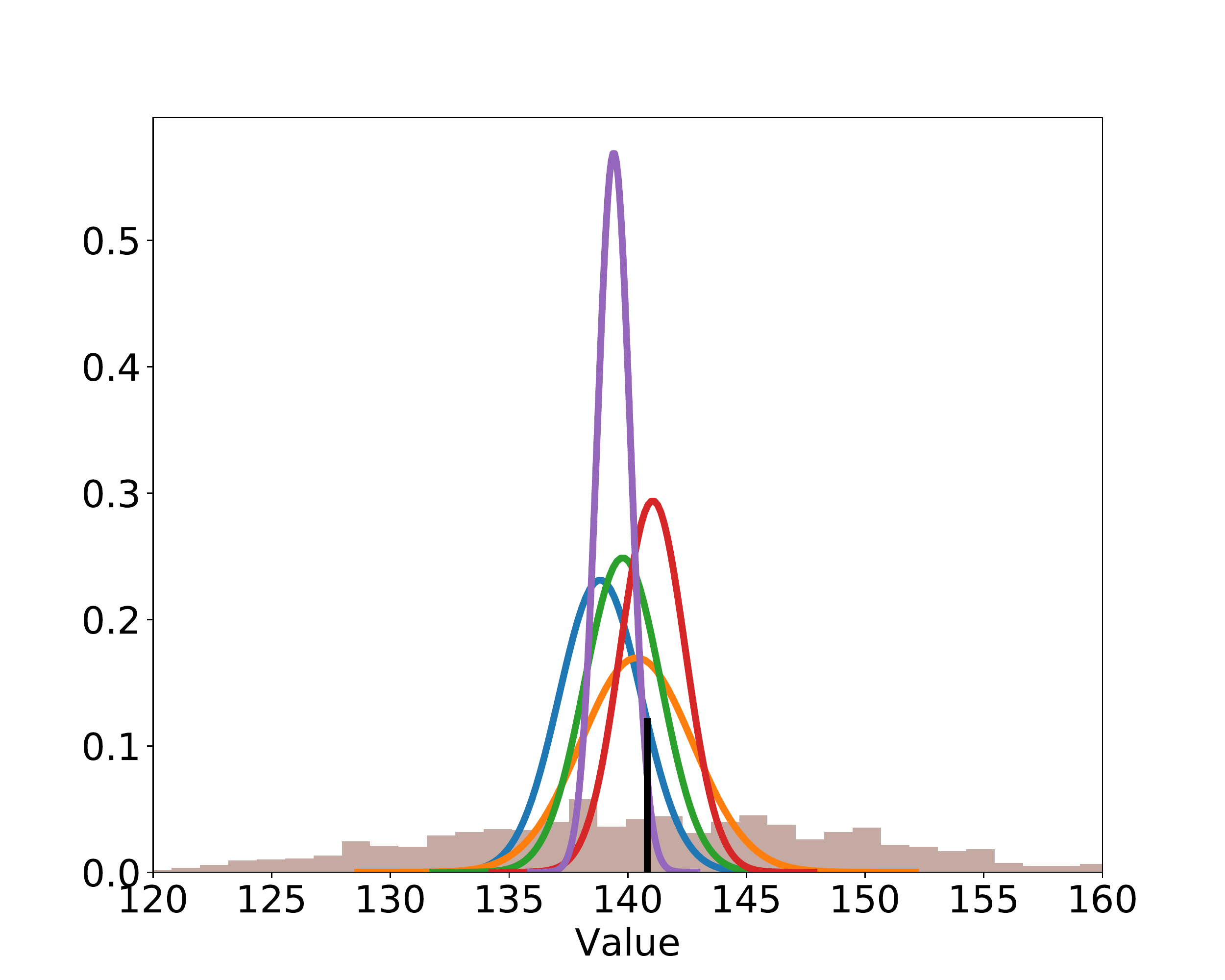}\caption{$T=1000$}
	\end{subfigure}
	\caption{Comparison of value function posteriors obtained by inferential induction and Monte Carlo evaluations at different time steps for a fixed policy. We plot for five runs of inferential induction at each time step. The value of the mean MDP is shown by a vertical line.}
	\label{fig:posterior_plots}
\end{figure*}

Here we evaluate whether inferential induction based policy evaluation (Alg.~\ref{alg:induction-pe}) results in a good approximation of the actual value function posterior.
 In order to evaluate the effectiveness of estimating the value function distribution using inferential induction (Alg.~\ref{alg:induction-pe}), we compare it with the \textit{Monte Carlo} distribution and the mean MDP. We compare this for posteriors after 10, 100 and 1000 time steps, obtained with a fixed policy in NChain that visits all the states, in Figure~\ref{fig:posterior_plots} for 5 runs of Alg.~\ref{alg:induction-pe}. The fixed policy selects the first action with probability 0.8 and the second action with probability 0.2. The Monte Carlo estimate is done through $1000$ samples of the value function vector ($\discount = 0.99$). This shows that the estimate of Alg.~\ref{alg:induction-pe} reasonably captures the uncertainty in the true distribution. For this data, we also compute the Wasserstein distance~\citep{fournier2015rate} between the true and the estimated distributions at the different time steps as can be found in Table~\ref{tab:wasserstein}. There we can see that the distance to the true distribution decreases over time.

\begin{table}[ht!]
	\centering
	\caption{Wasserstein distance to the true distribution of the value function, for Alg.~\ref{alg:induction-pe} and the mean MDP model, for NChain. For Inferential Induction, the distances are averaged over 5 runs. The distances correspond to the plots in Figure~\ref{fig:posterior_plots}.}
	\begin{tabular}{l l l}
		\toprule
		Time steps& Inf. Induction & Mean MDP \\
		\midrule
		10  & 22.80 & 30.69\\
		100 & 16.41& 17.90\\
		1000 & 4.18 & 4.27\\
		\bottomrule
	\end{tabular}
	
	\label{tab:wasserstein}
\end{table}

\subsection{Variance in Performance}
\begin{figure*}[t!]
	\begin{subfigure}[t]{0.32\textwidth}
		\includegraphics[width=\linewidth]{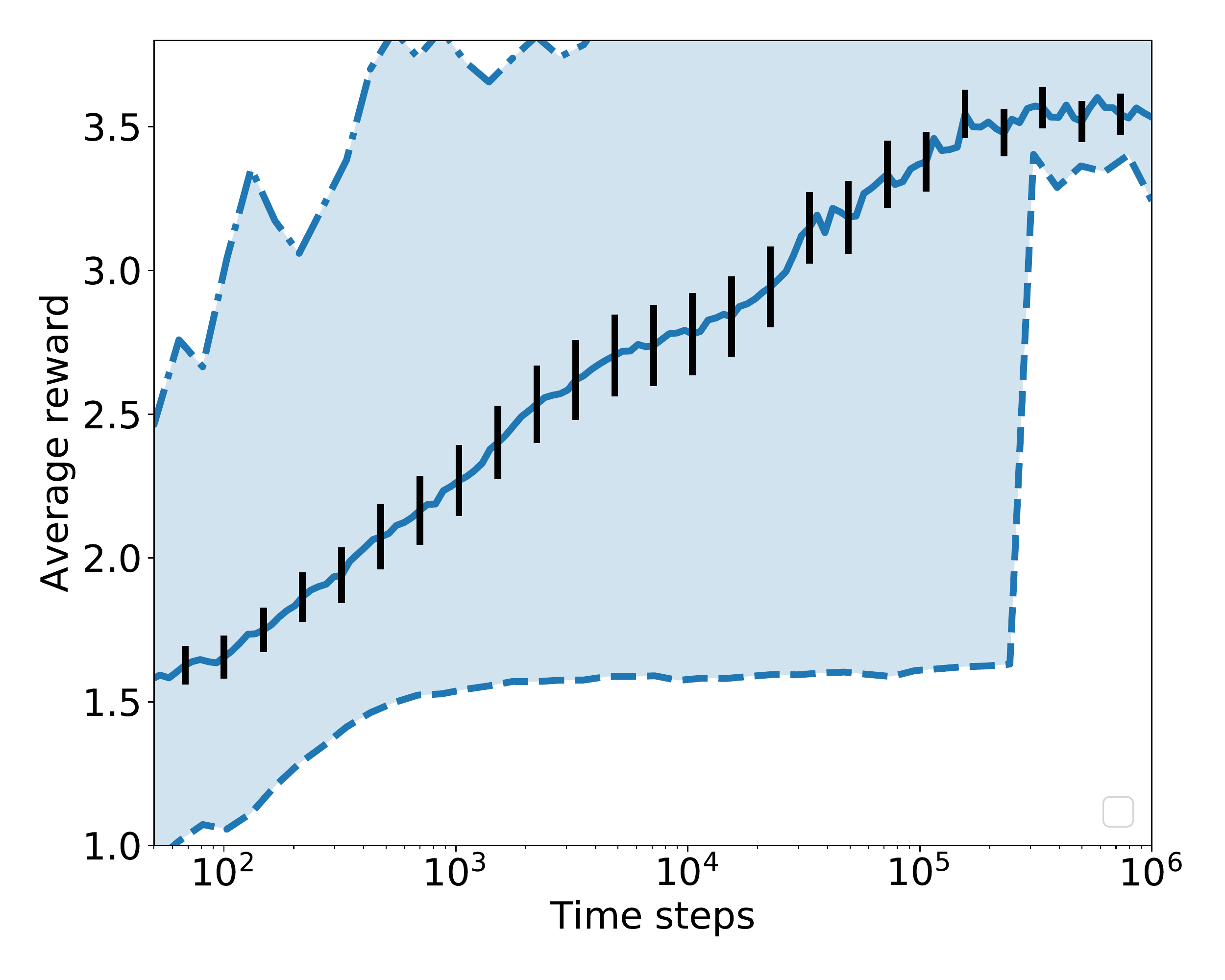}
		\caption{BBI}
		\label{fig:chainLGP_ci}
	\end{subfigure}
	\begin{subfigure}[t]{0.32\textwidth}
		\includegraphics[width=\linewidth]{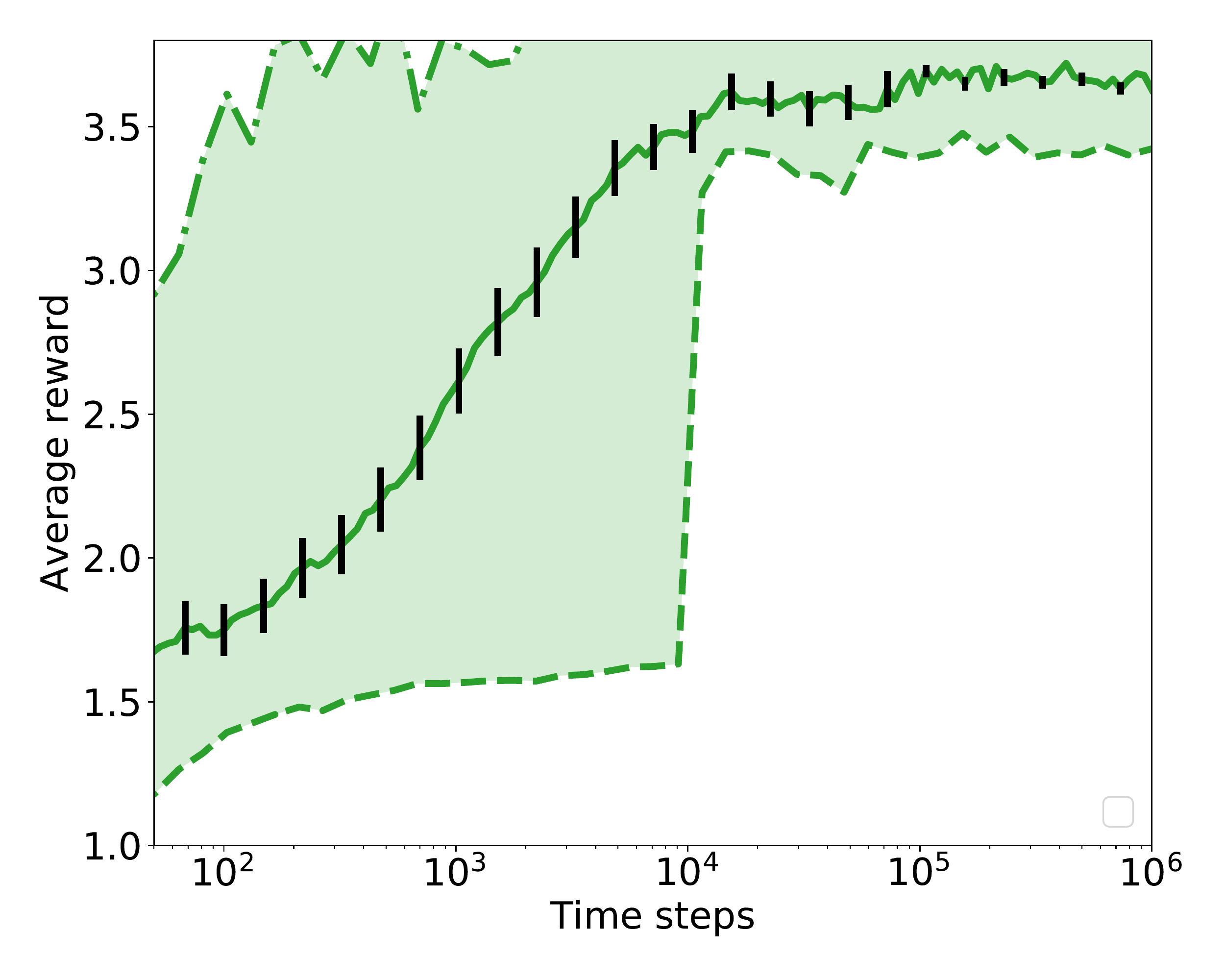}
		\caption{PSRL}
		\label{fig:chainPSRL_ci}
	\end{subfigure}
	\begin{subfigure}[t]{0.32\textwidth}
		\includegraphics[width=\linewidth]{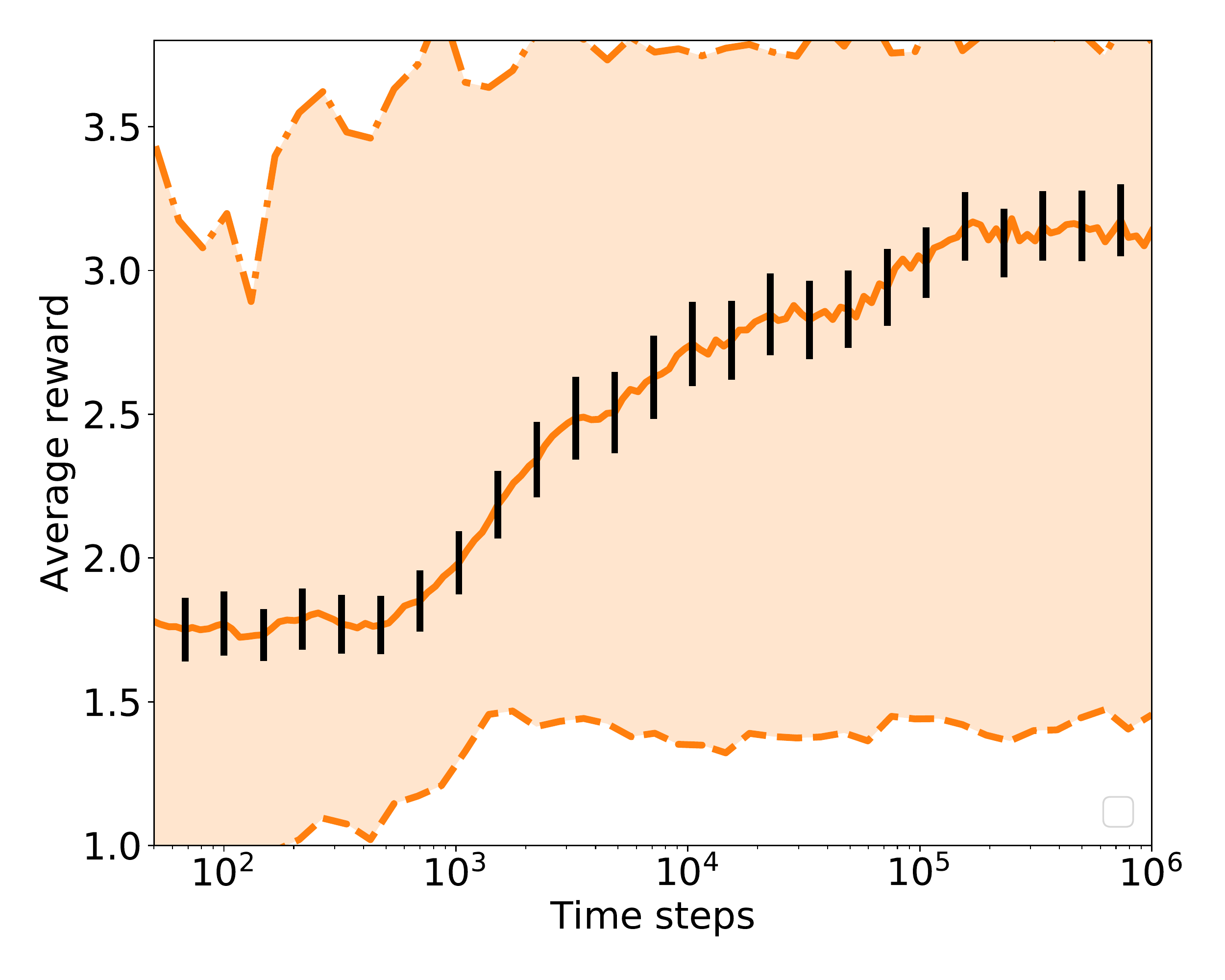}
		\caption{VDQN}
		\label{fig:chainVDQN_ci}
	\end{subfigure}
	\\
	\begin{subfigure}[t]{0.32\textwidth}
		\includegraphics[width=\linewidth]{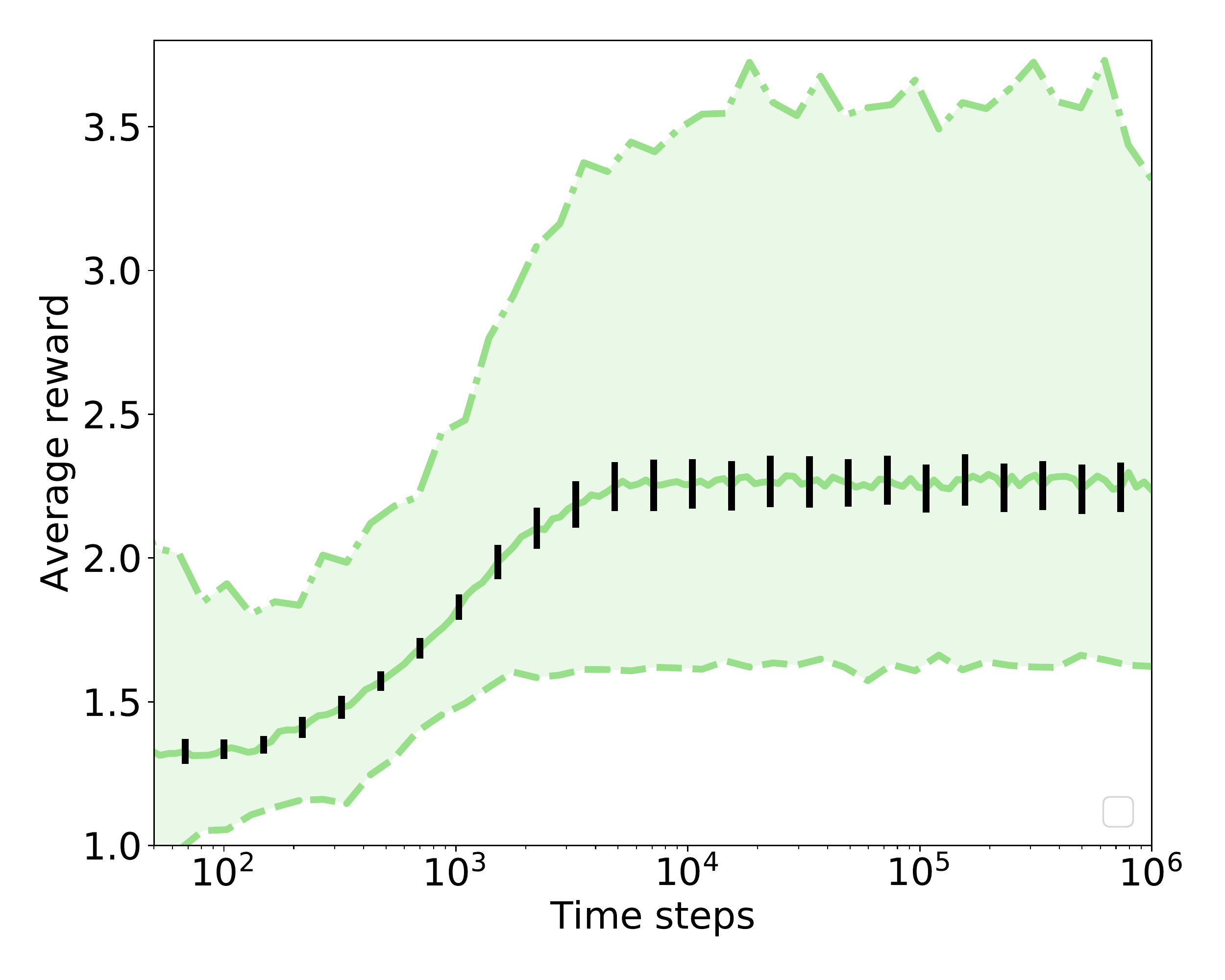}
		\caption{BQL}
		\label{fig:chainBQL_ci}
	\end{subfigure}
	\begin{subfigure}[t]{0.32\textwidth}
		\includegraphics[width=\linewidth]{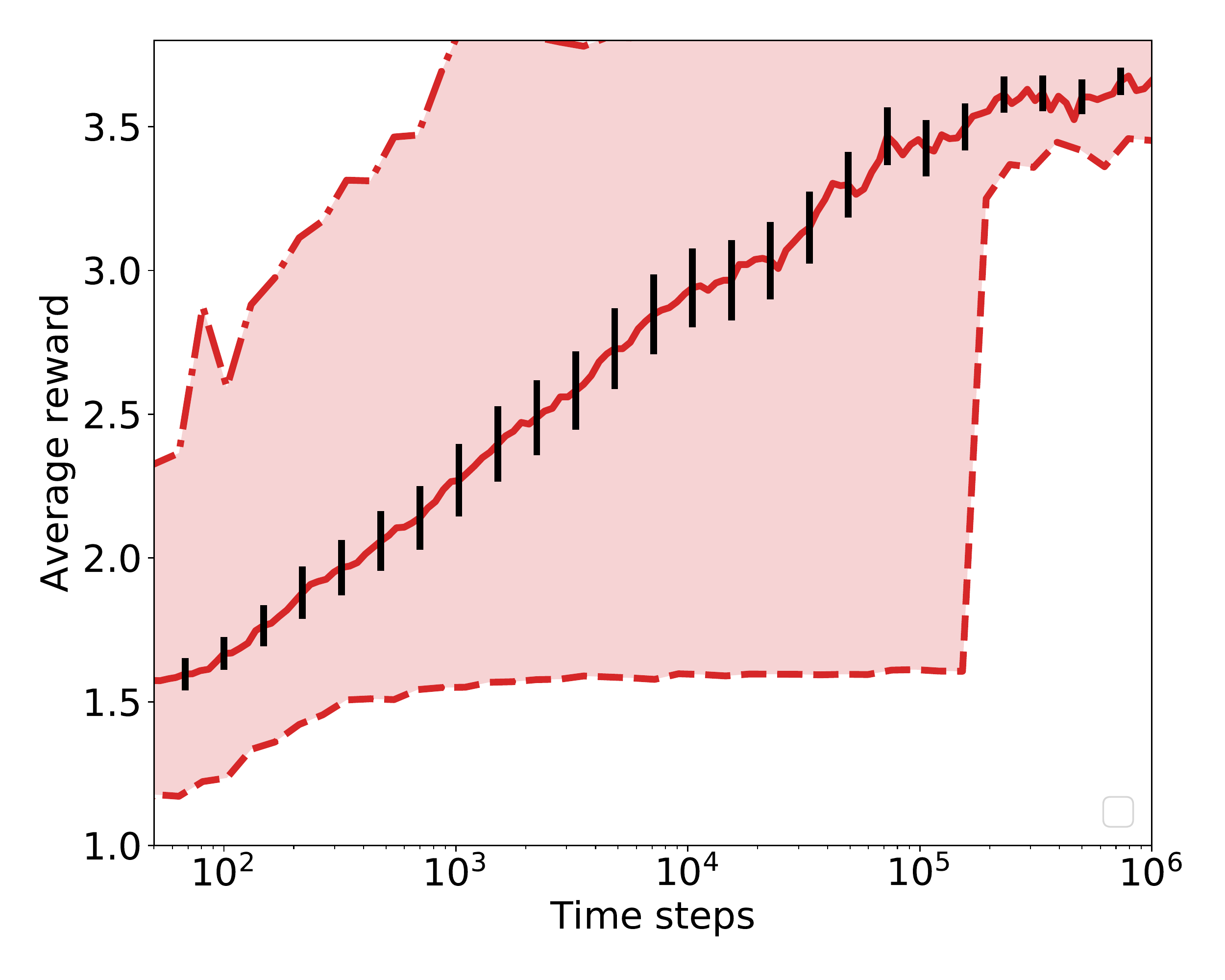}
		\caption{MMBI}
		\label{fig:chainMMBI_ci}
	\end{subfigure}
	\begin{subfigure}[t]{0.32\textwidth}
		\includegraphics[width=\linewidth]{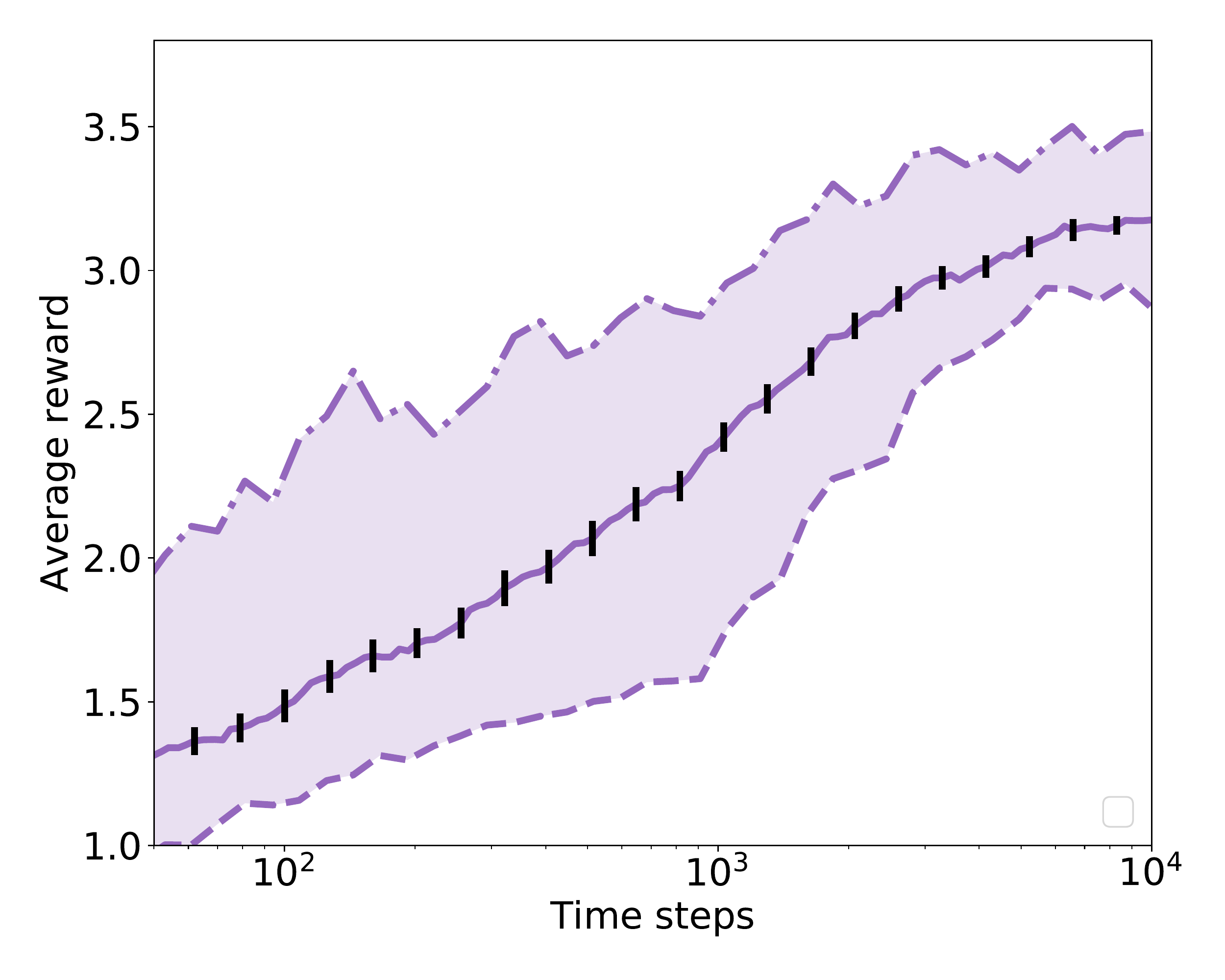}
		\caption{BSS}
		\label{fig:chainBSS_ci}
	\end{subfigure}
	\caption{Evolution of average reward for NChain environment with 50 runs of length $10^6$ for each algorithm. For computational reasons BSS is only run for $10^4$ steps. The runs are exponentially smoothened with a half-life $1000$. The mean as well as the 5th and 95th percentile performance is shown for each algorithm and the standard error is illustrated with black lines.}\label{fig:ci_chain}
\end{figure*}

\begin{figure*}[t!]
	\begin{subfigure}[t]{0.32\textwidth}
		\includegraphics[width=\linewidth]{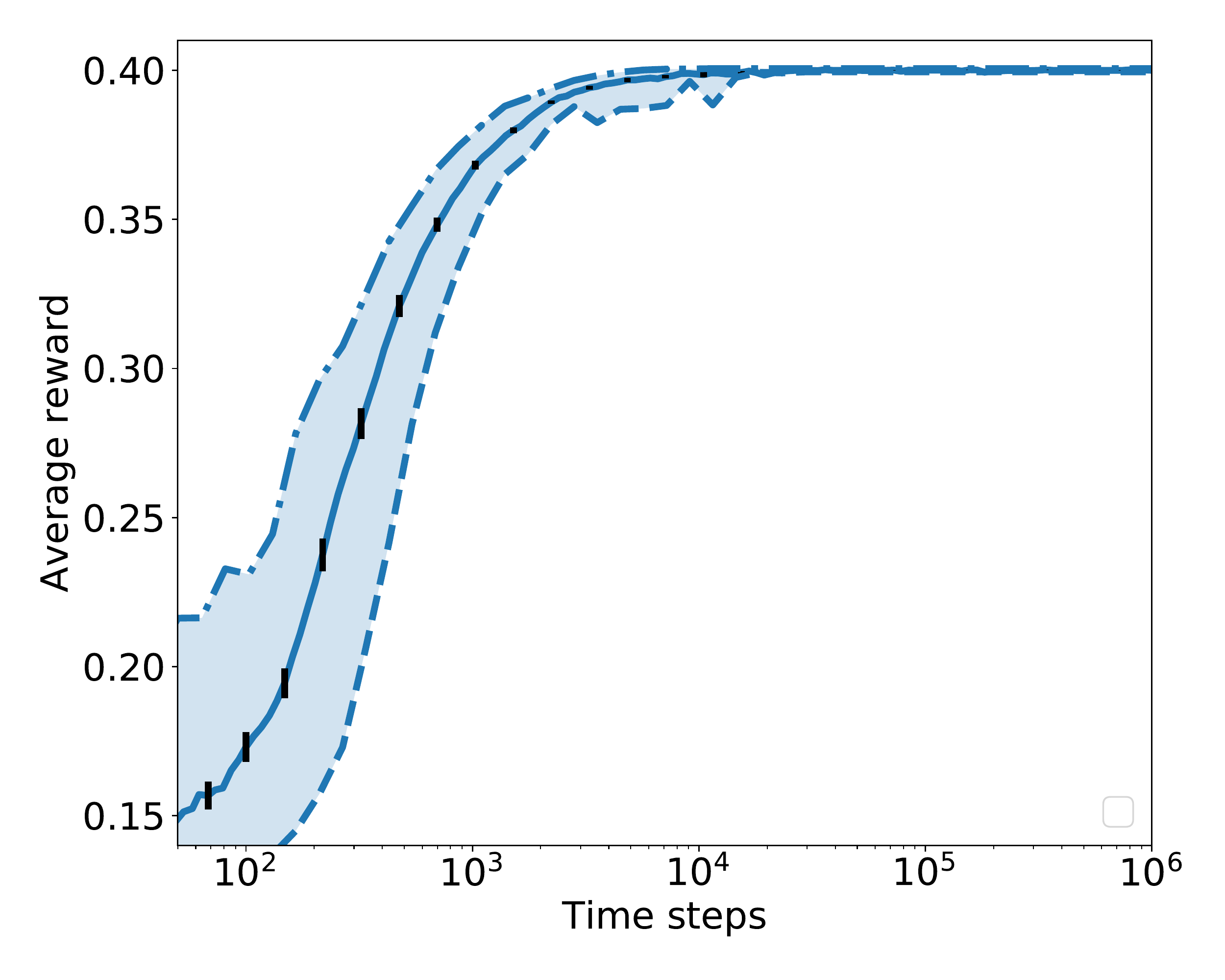}
		\caption{BBI}
		\label{fig:loopLGP_ci}
	\end{subfigure}
	\begin{subfigure}[t]{0.32\textwidth}
		\includegraphics[width=\linewidth]{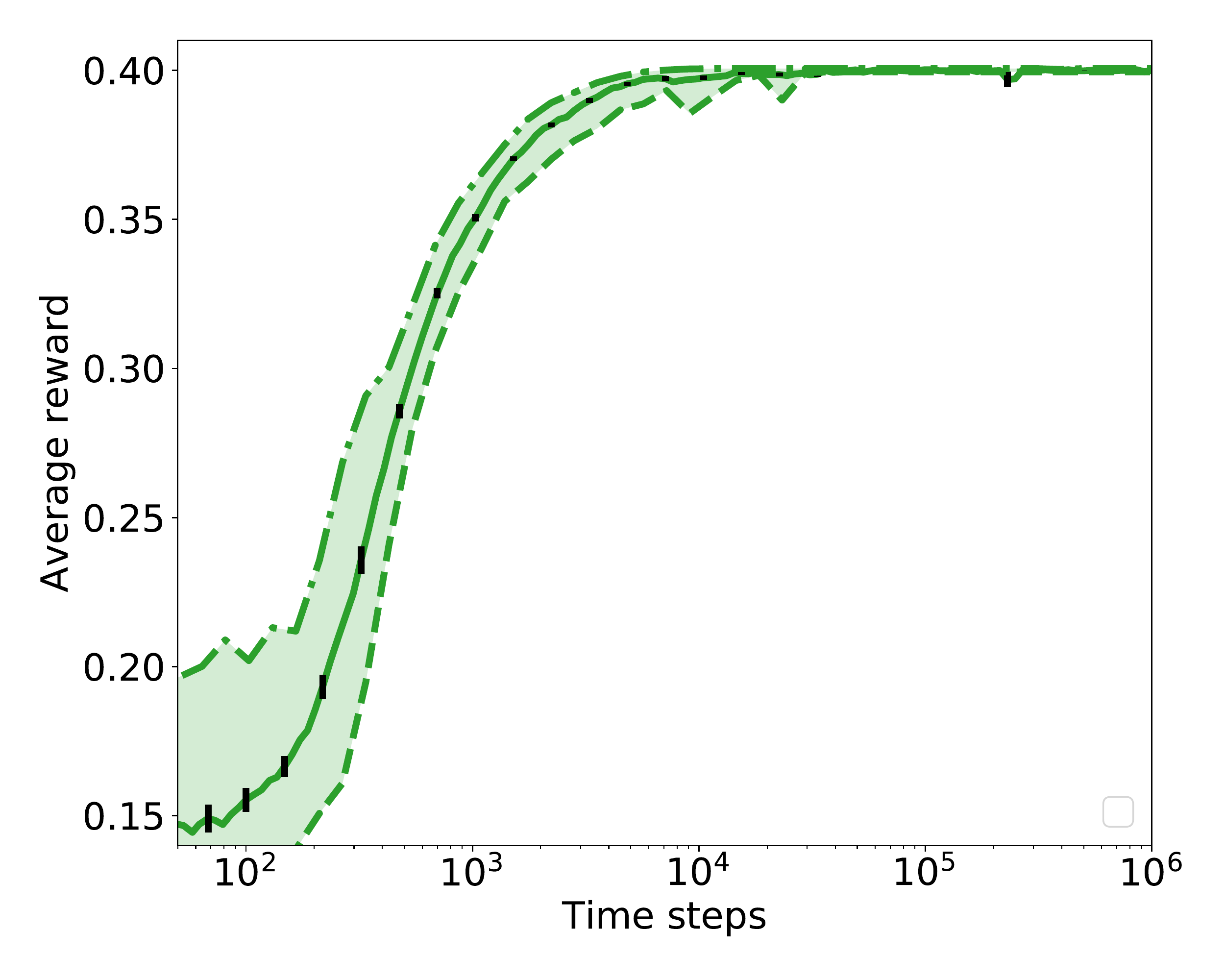}
		\caption{PSRL}
		\label{fig:loopPSRL_ci}
	\end{subfigure}
	\begin{subfigure}[t]{0.32\textwidth}
		\includegraphics[width=\linewidth]{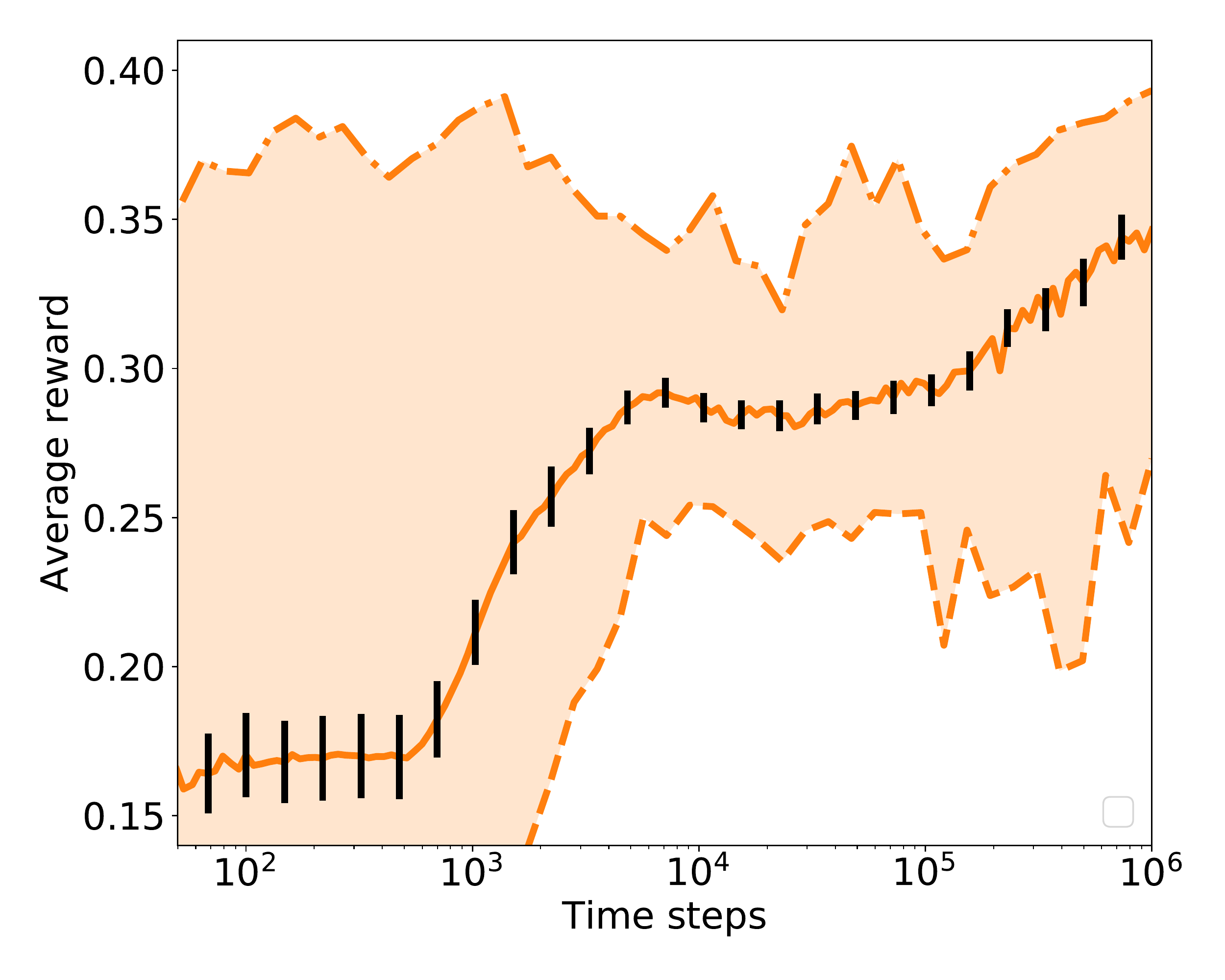}
		\caption{VDQN}
		\label{fig:loopVDQN_ci}
	\end{subfigure}
	\\
	\begin{subfigure}[t]{0.32\textwidth}
		\includegraphics[width=\linewidth]{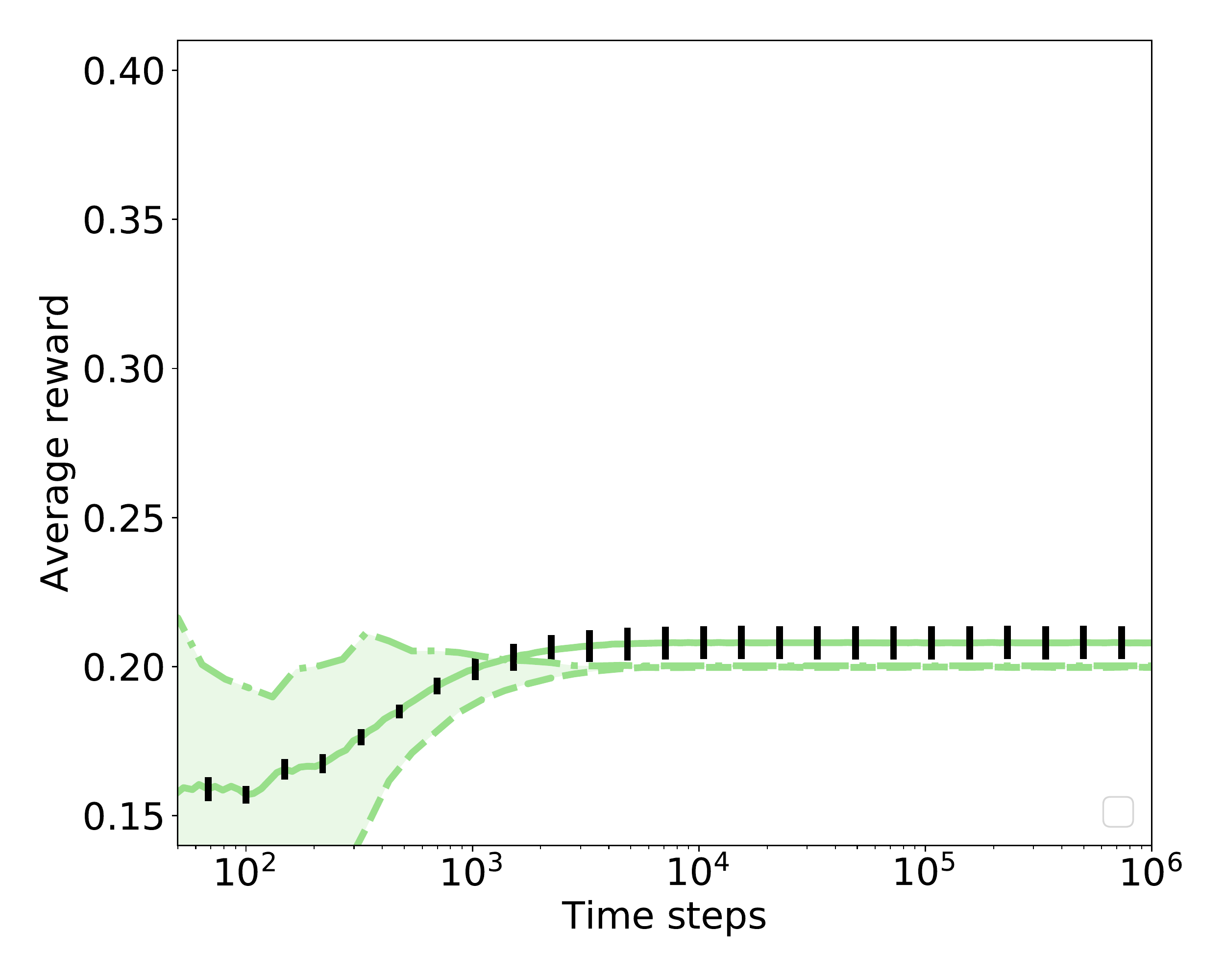}
		\caption{BQL}
		\label{fig:loopBQL_ci}
	\end{subfigure}
	\begin{subfigure}[t]{0.32\textwidth}
		\includegraphics[width=\linewidth]{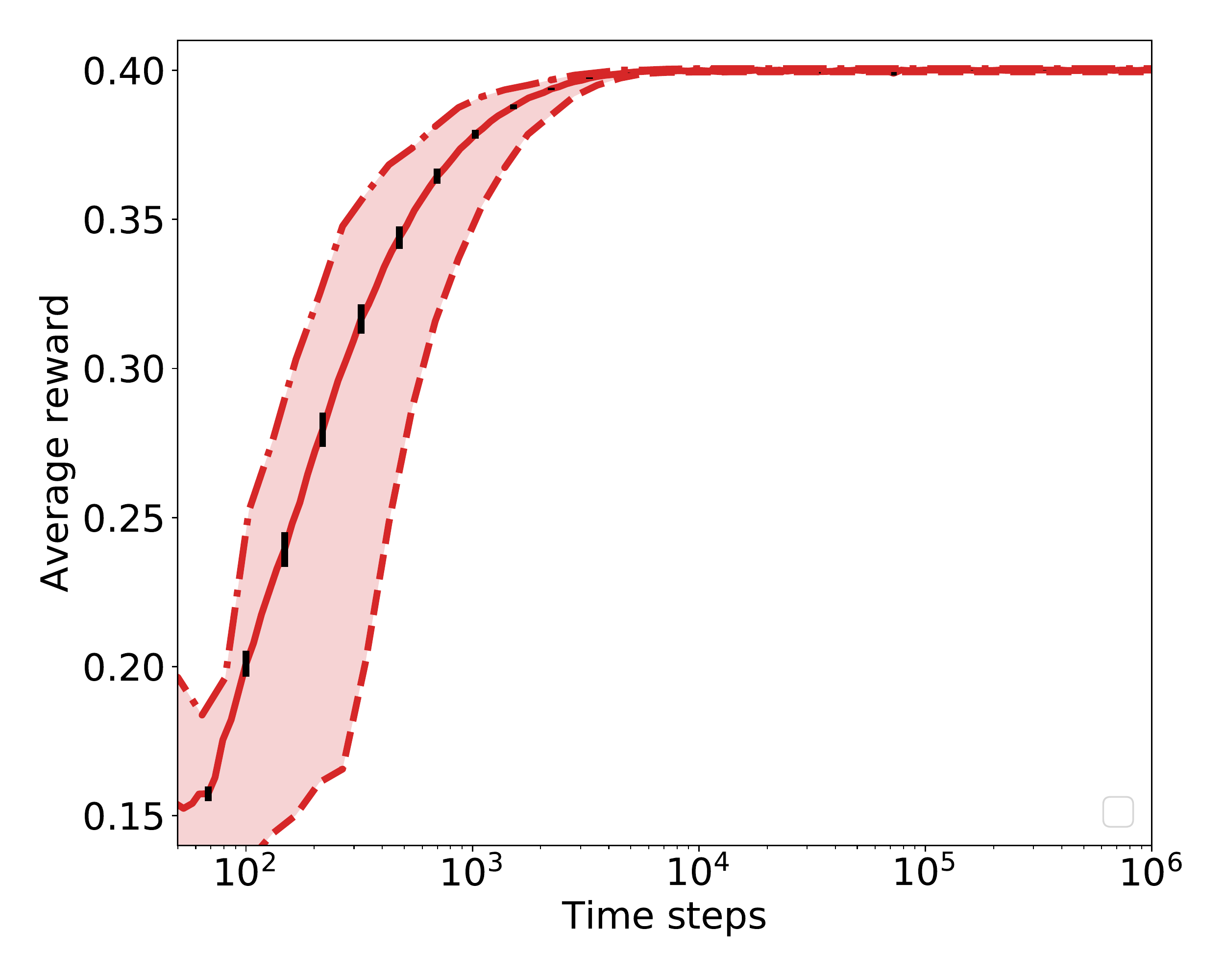}
		\caption{MMBI}
		\label{fig:loopMMBI_ci}
	\end{subfigure}
	\begin{subfigure}[t]{0.32\textwidth}
		\includegraphics[width=\linewidth]{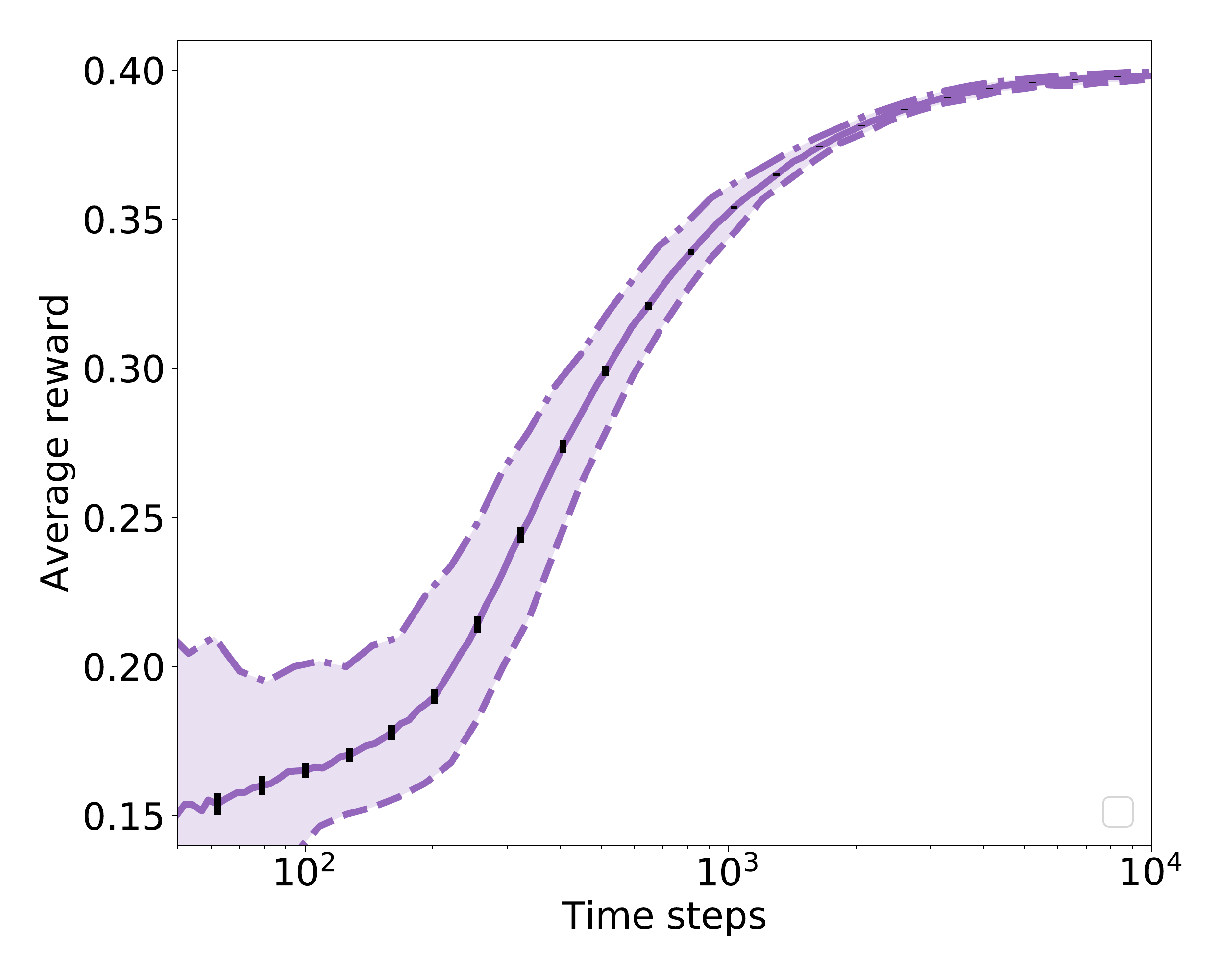}
		\caption{BSS}
		\label{fig:loopBSS_ci}
	\end{subfigure}
	\caption{Evolution of average reward for DoubleLoop environment with 50 runs of length $10^6$ for each algorithm. For computational reasons BSS is only run for $10^4$ steps. The runs are exponentially smoothened with a half-life $1000$. The mean as well as the 5th and 95th percentile performance is shown for each algorithm and the standard error is illustrated with black lines.}\label{fig:ci_loop}
\end{figure*}

\begin{figure*}[t!]
	\begin{subfigure}[t]{0.49\textwidth}
		\includegraphics[width=\linewidth]{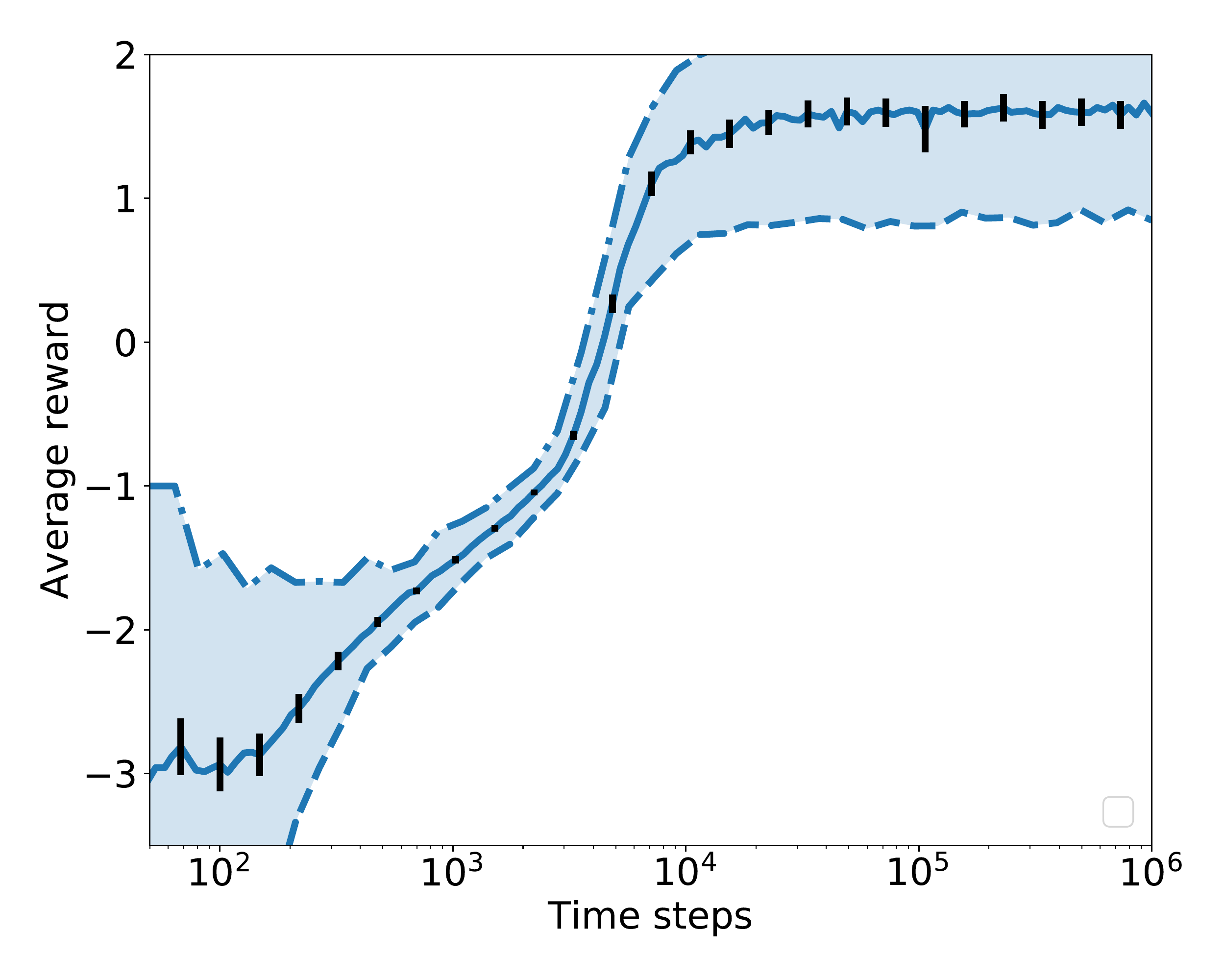}
		\caption{BBI}
		\label{fig:lava_lake_smallLGP_ci}
	\end{subfigure}
	\begin{subfigure}[t]{0.49\textwidth}
		\includegraphics[width=\linewidth]{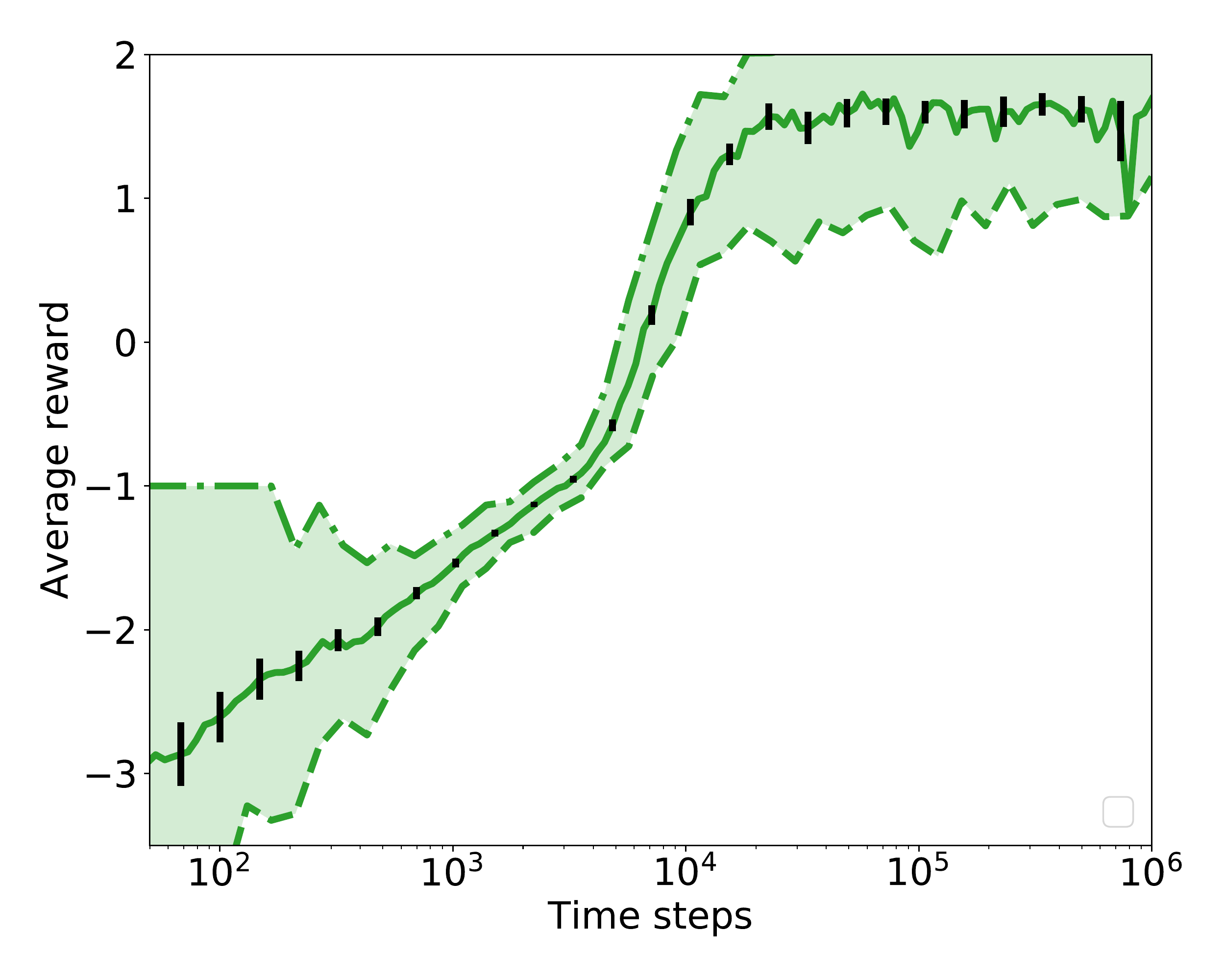}
		\caption{PSRL}
		\label{fig:lava_lake_smallPSRL_ci}
	\end{subfigure}
	\\
	\begin{subfigure}[t]{0.32\textwidth}
		\includegraphics[width=\linewidth]{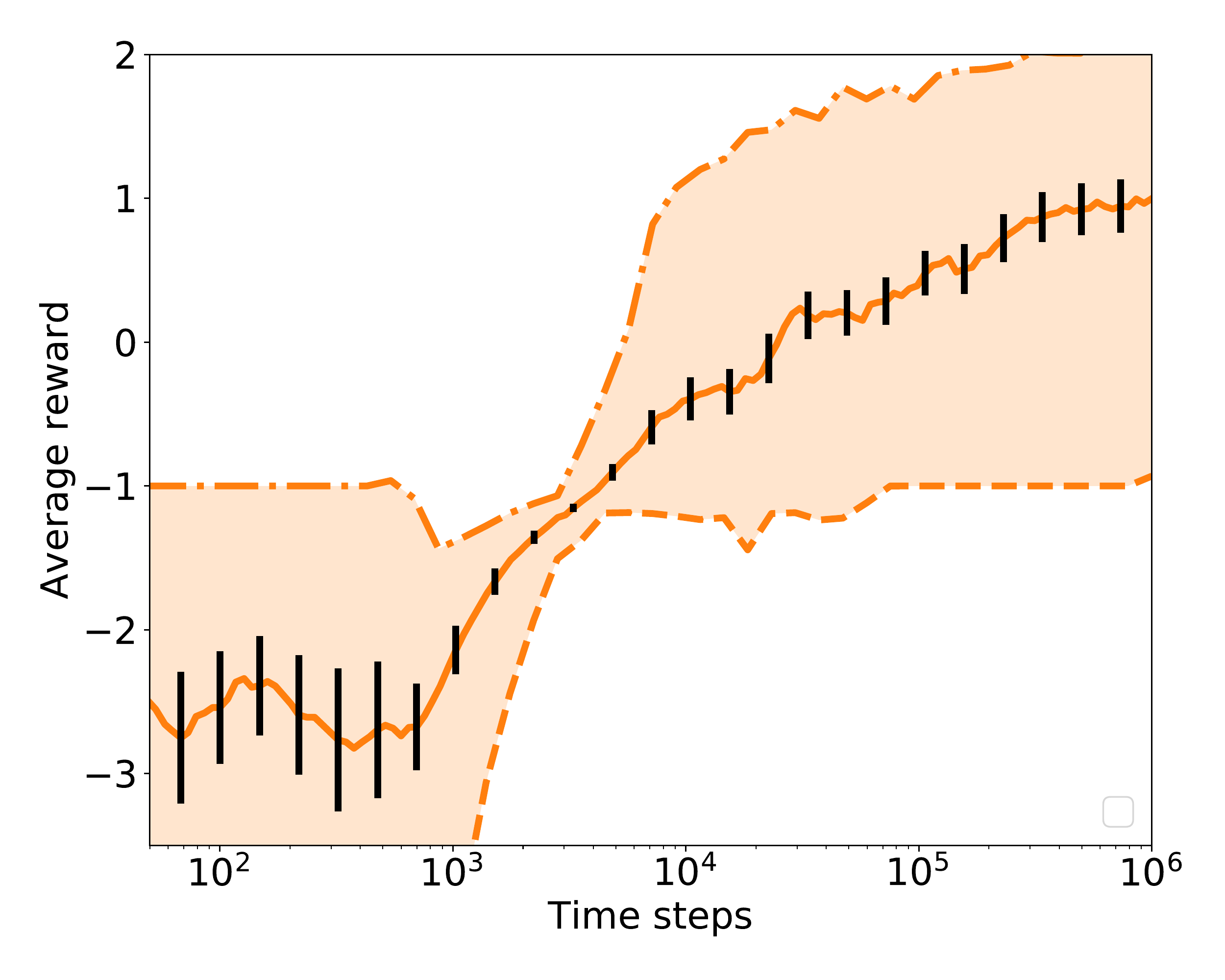}
		\caption{VDQN}
		\label{fig:lava_lake_smallVDQN_ci}
	\end{subfigure}
	\begin{subfigure}[t]{0.32\textwidth}
		\includegraphics[width=\linewidth]{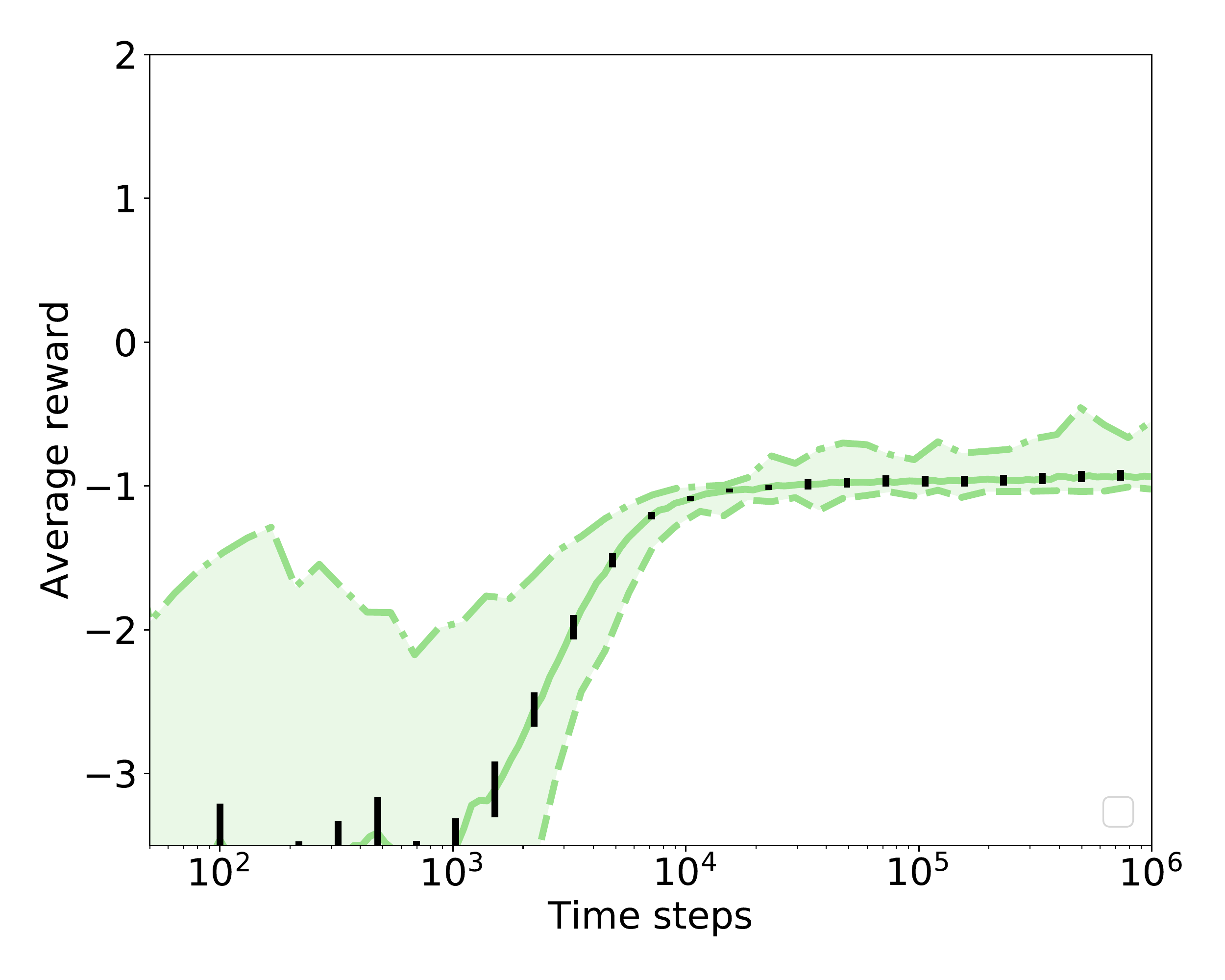}
		\caption{BQL}
		\label{fig:lava_lake_smallBQL_ci}
	\end{subfigure}
	\begin{subfigure}[t]{0.32\textwidth}
		\includegraphics[width=\linewidth]{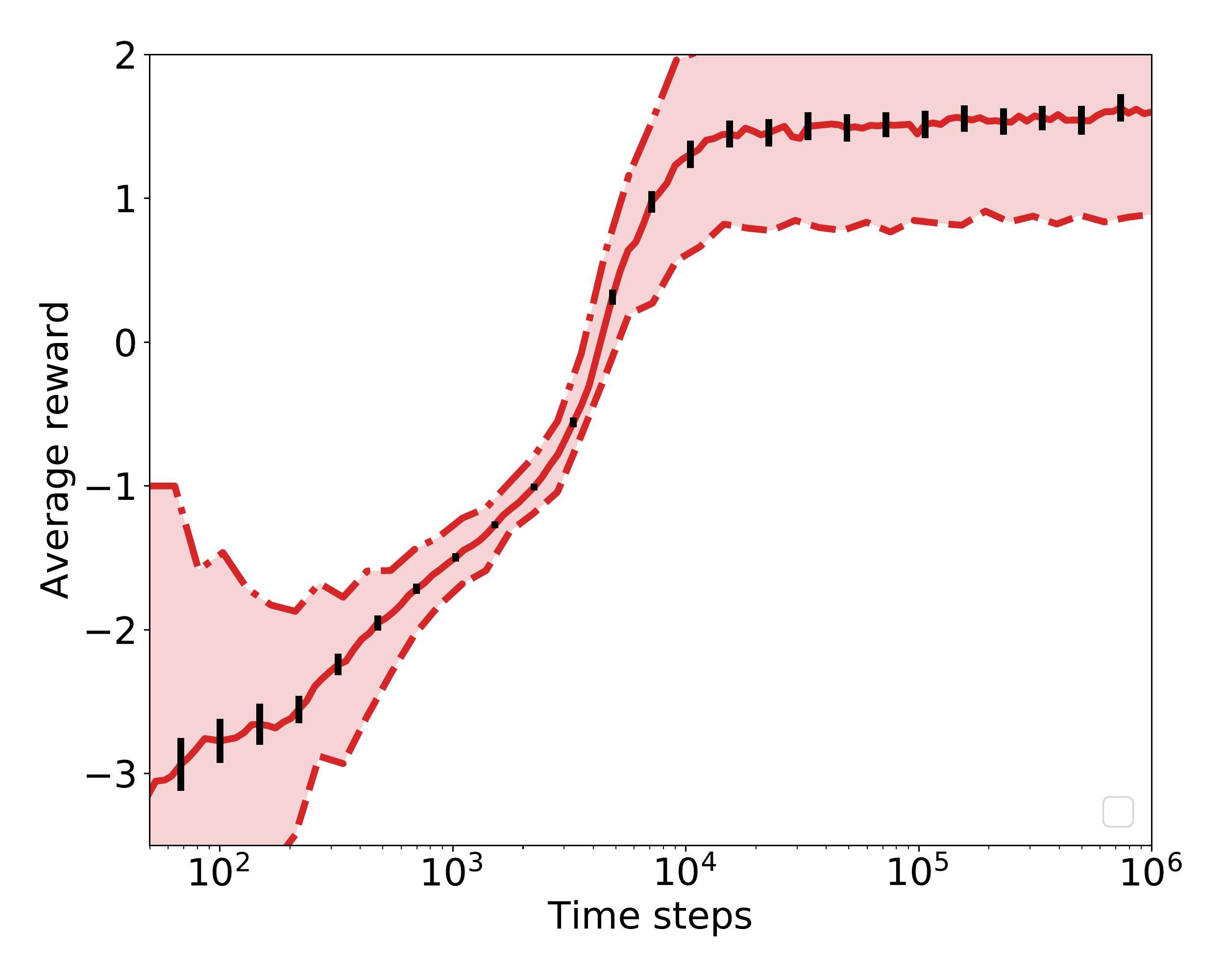}
		\caption{MMBI}
		\label{fig:lava_lake_smallMMBI_ci}
	\end{subfigure}
	\caption{Evolution of average reward for LavaLake $5 \times 7$ environment with 20 runs of length $10^6$ for each algorithm. The runs are exponentially smoothened with a half-life $1000$. The mean as well as the 5th and 95th percentile performance is shown for each algorithm and the standard error is illustrated with black lines.}\label{fig:ci_lava_lake_small}
\end{figure*}

\begin{figure*}[t!]
	\begin{subfigure}[t]{0.49\textwidth}
		\includegraphics[width=\linewidth]{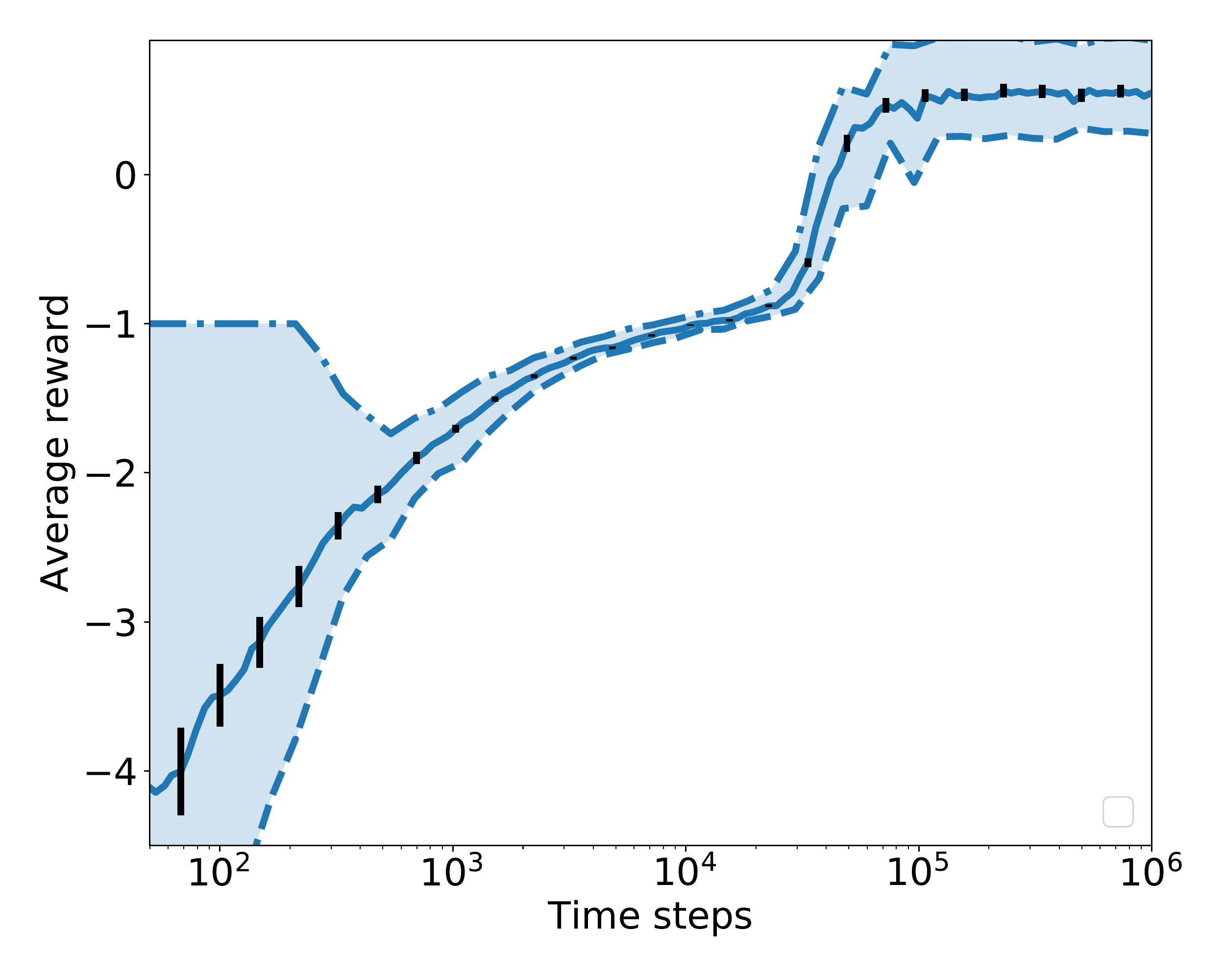}
		\caption{BBI}
		\label{fig:lava_lakeLGP_ci}
	\end{subfigure}
	\begin{subfigure}[t]{0.49\textwidth}
		\includegraphics[width=\linewidth]{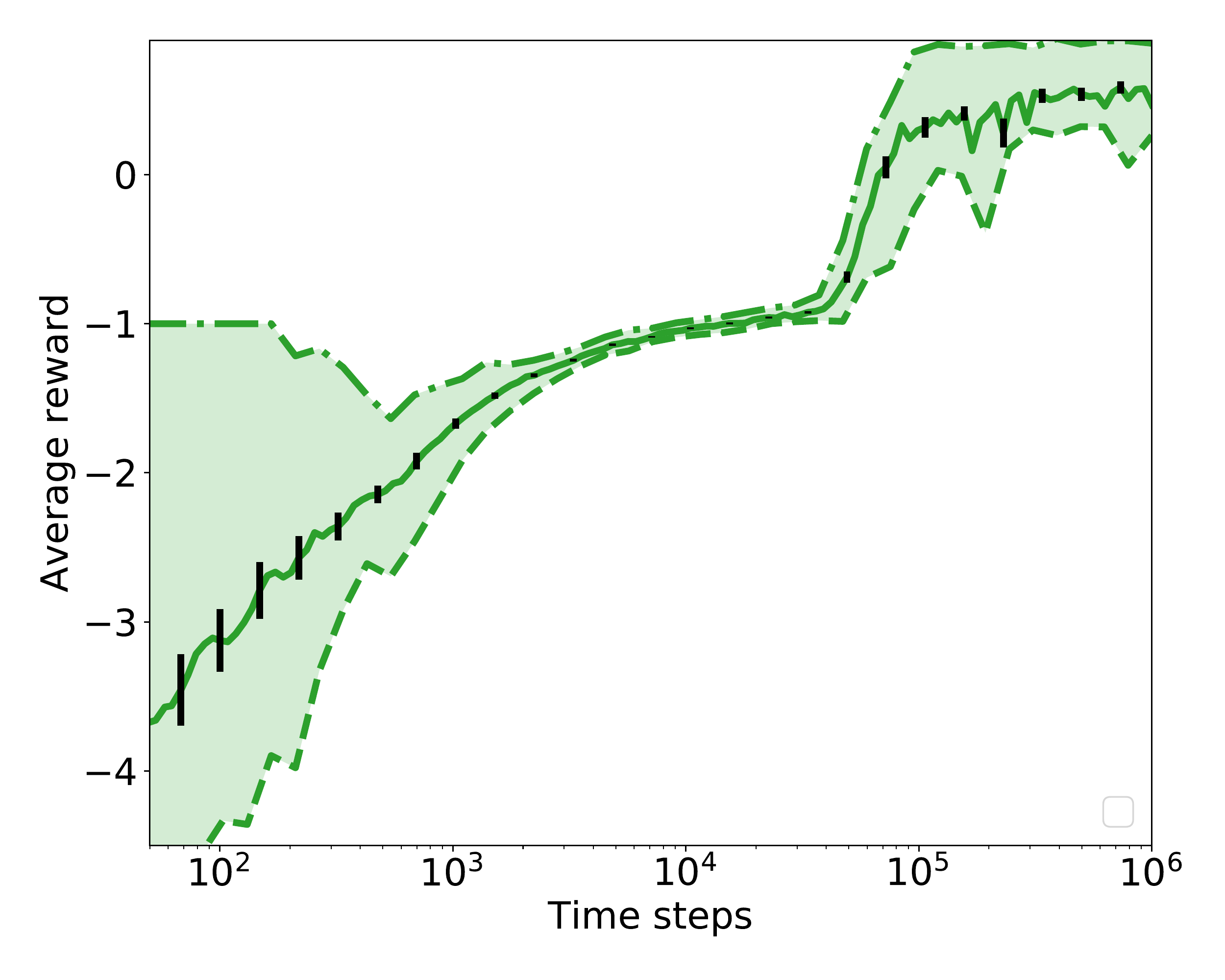}
		\caption{PSRL}
		\label{fig:lava_lakePSRL_ci}
	\end{subfigure}
	\\
	\begin{subfigure}[t]{0.32\textwidth}
		\includegraphics[width=\linewidth]{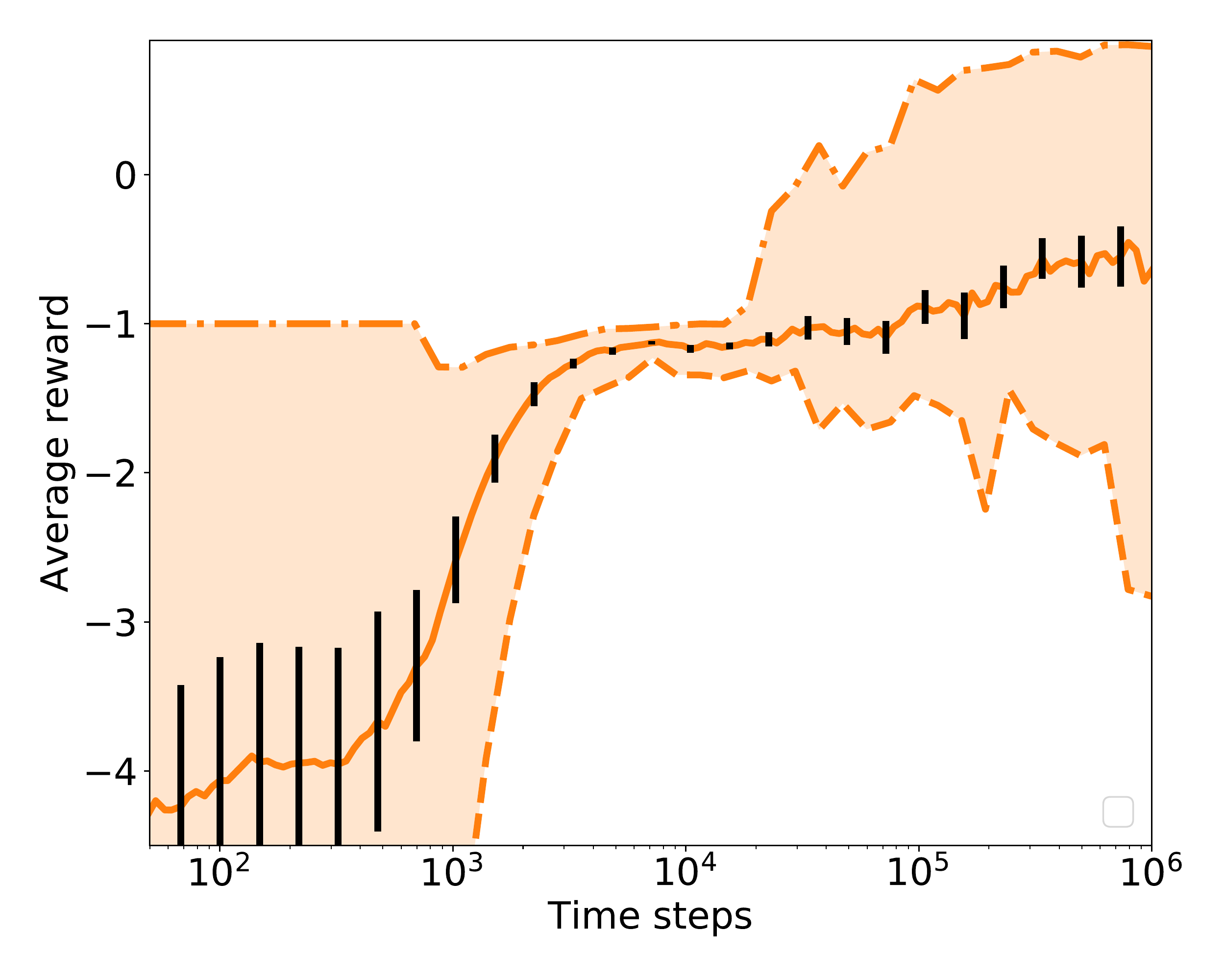}
		\caption{VDQN}
		\label{fig:lava_lakeVDQN_ci}
	\end{subfigure}
	\begin{subfigure}[t]{0.32\textwidth}
		\includegraphics[width=\linewidth]{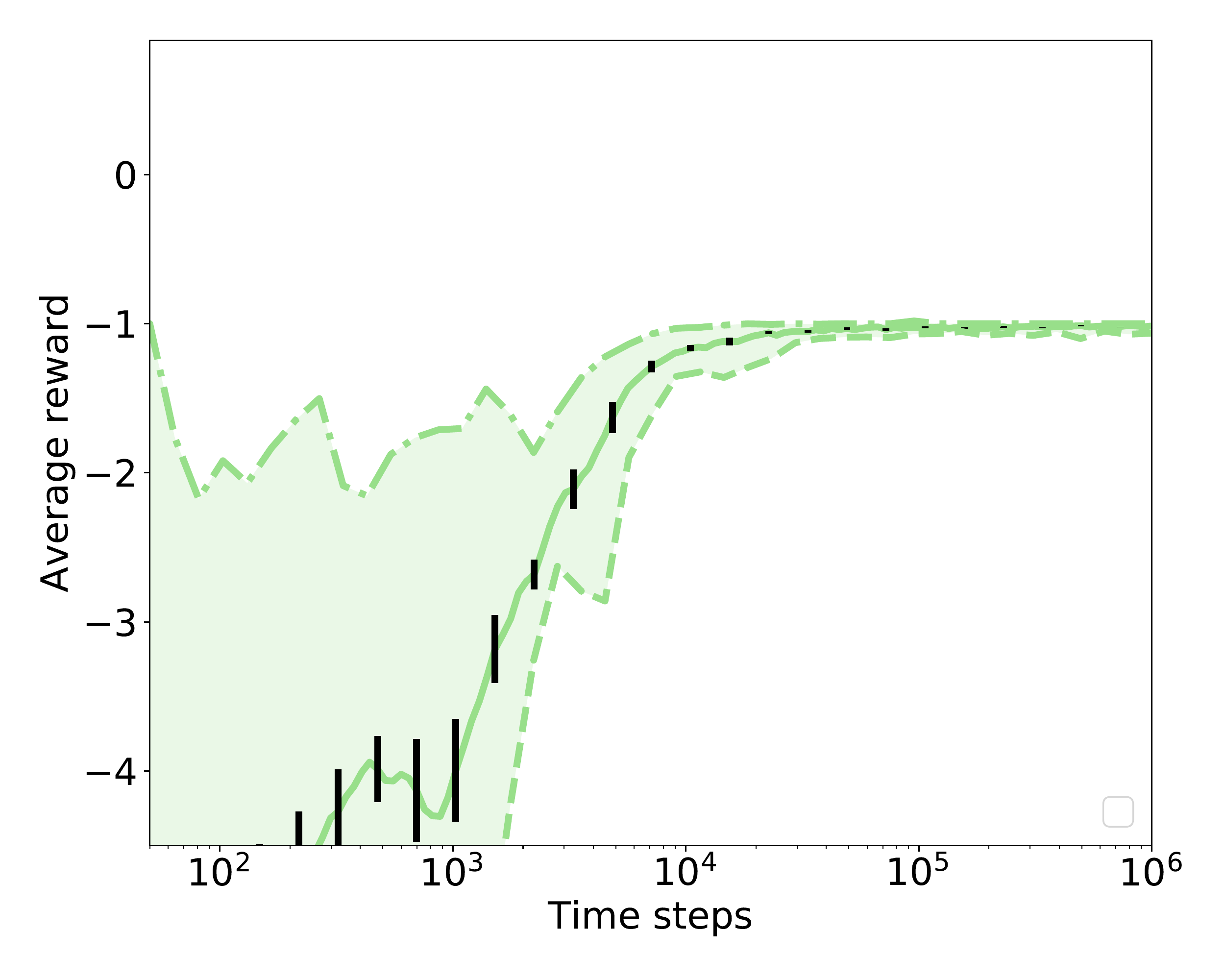}
		\caption{BQL}
		\label{fig:lava_lakeBQL_ci}
	\end{subfigure}
	\begin{subfigure}[t]{0.32\textwidth}
		\includegraphics[width=\linewidth]{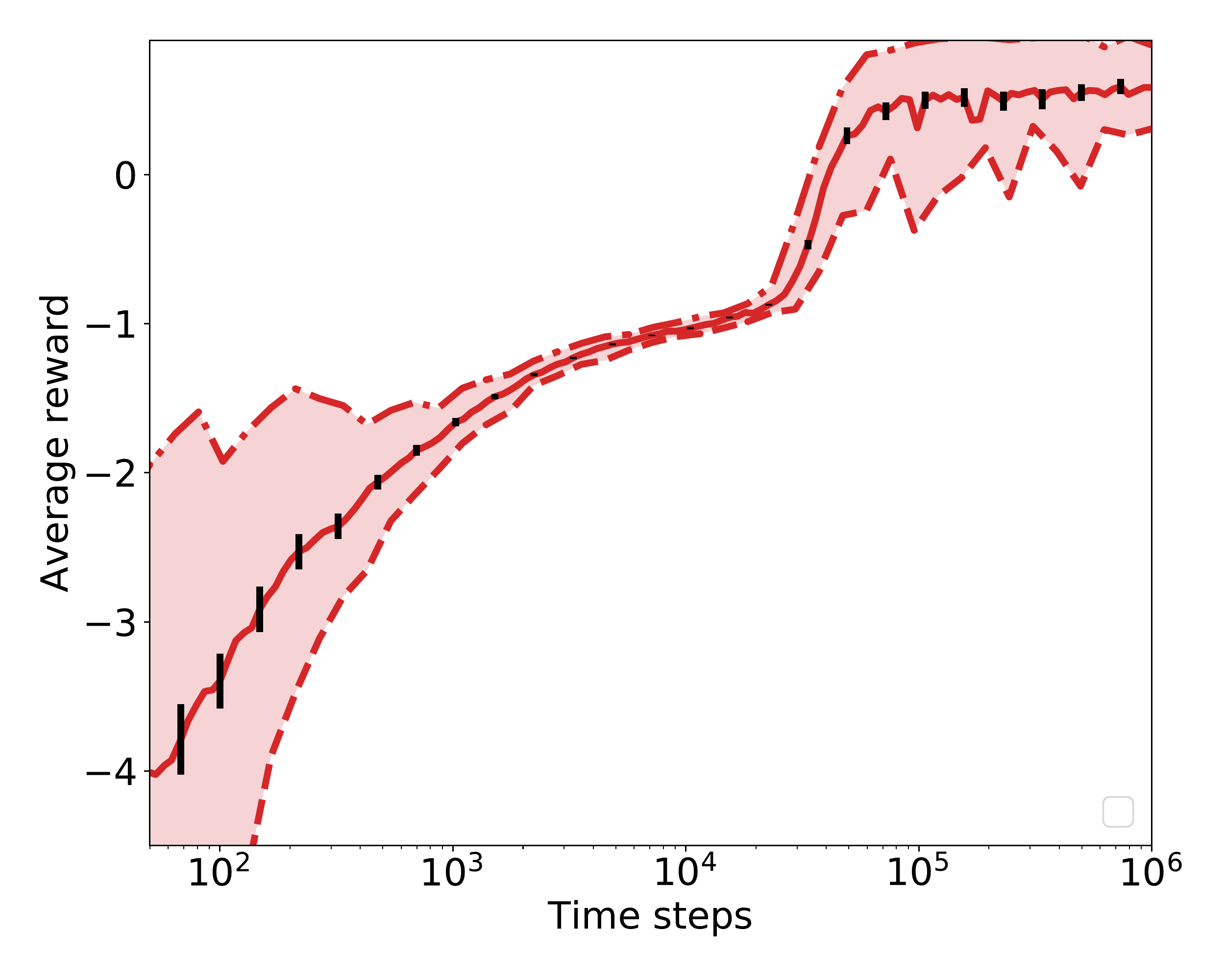}
		\caption{MMBI}
		\label{fig:lava_lakeMMBI_ci}
	\end{subfigure}
	\caption{Evolution of average reward for LavaLake $10\times 10$ environment with 30 runs of length $10^6$ for each algorithm. The runs are exponentially smoothened with a half-life $1000$. The mean as well as the 5th and 95th percentile performance is shown for each algorithm and the standard error is illustrated with black lines.}\label{fig:ci_lava_lake}
\end{figure*}

\begin{figure*}[t!]
	\begin{subfigure}[t]{0.49\textwidth}
		\includegraphics[width=\linewidth]{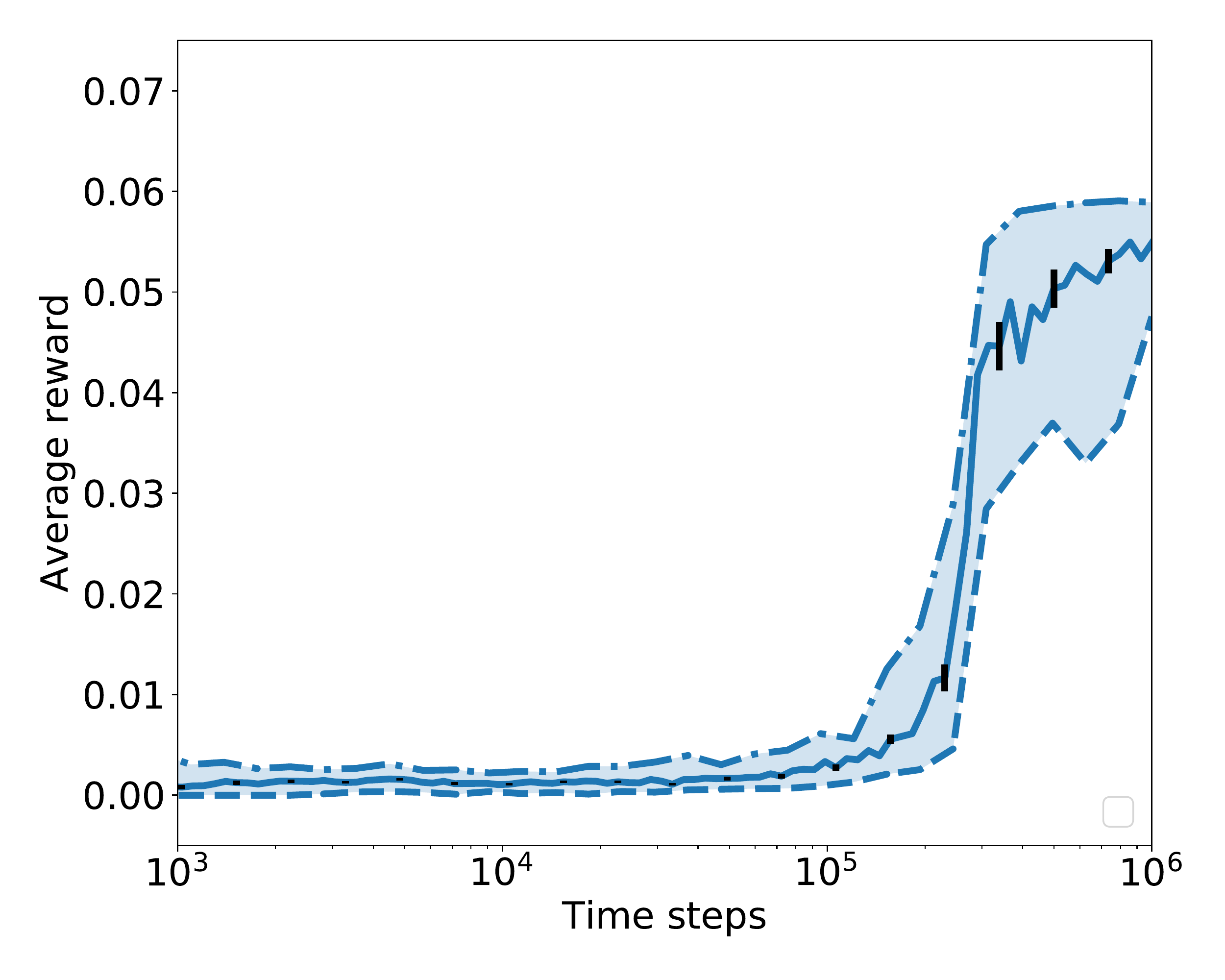}
		\caption{BBI}
		\label{fig:mazeLGP_ci}
	\end{subfigure}
	\begin{subfigure}[t]{0.49\textwidth}
		\includegraphics[width=\linewidth]{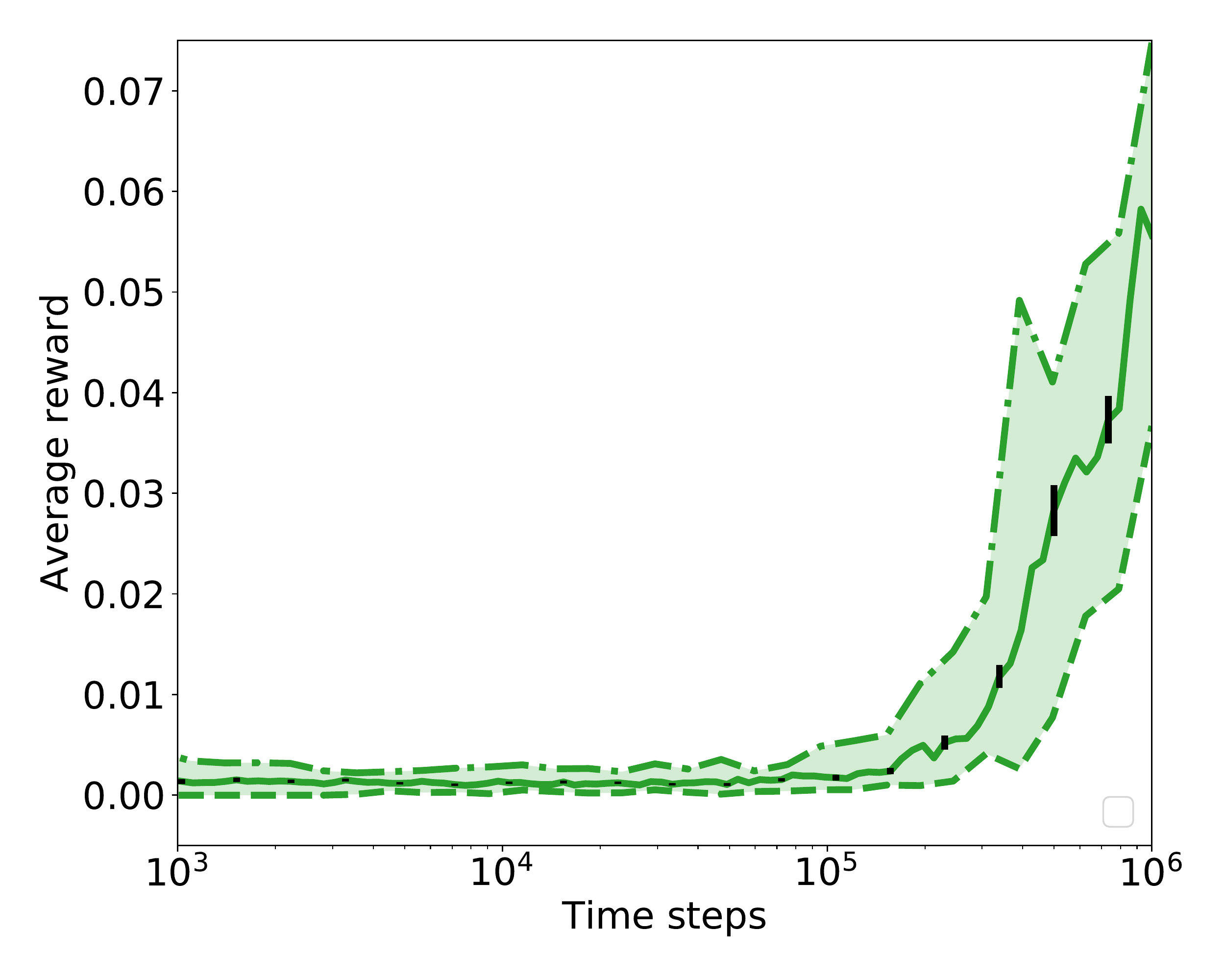}
		\caption{PSRL}
		\label{fig:mazePSRL_ci}
	\end{subfigure}
	\\
	\begin{subfigure}[t]{0.49\textwidth}
		\includegraphics[width=\linewidth]{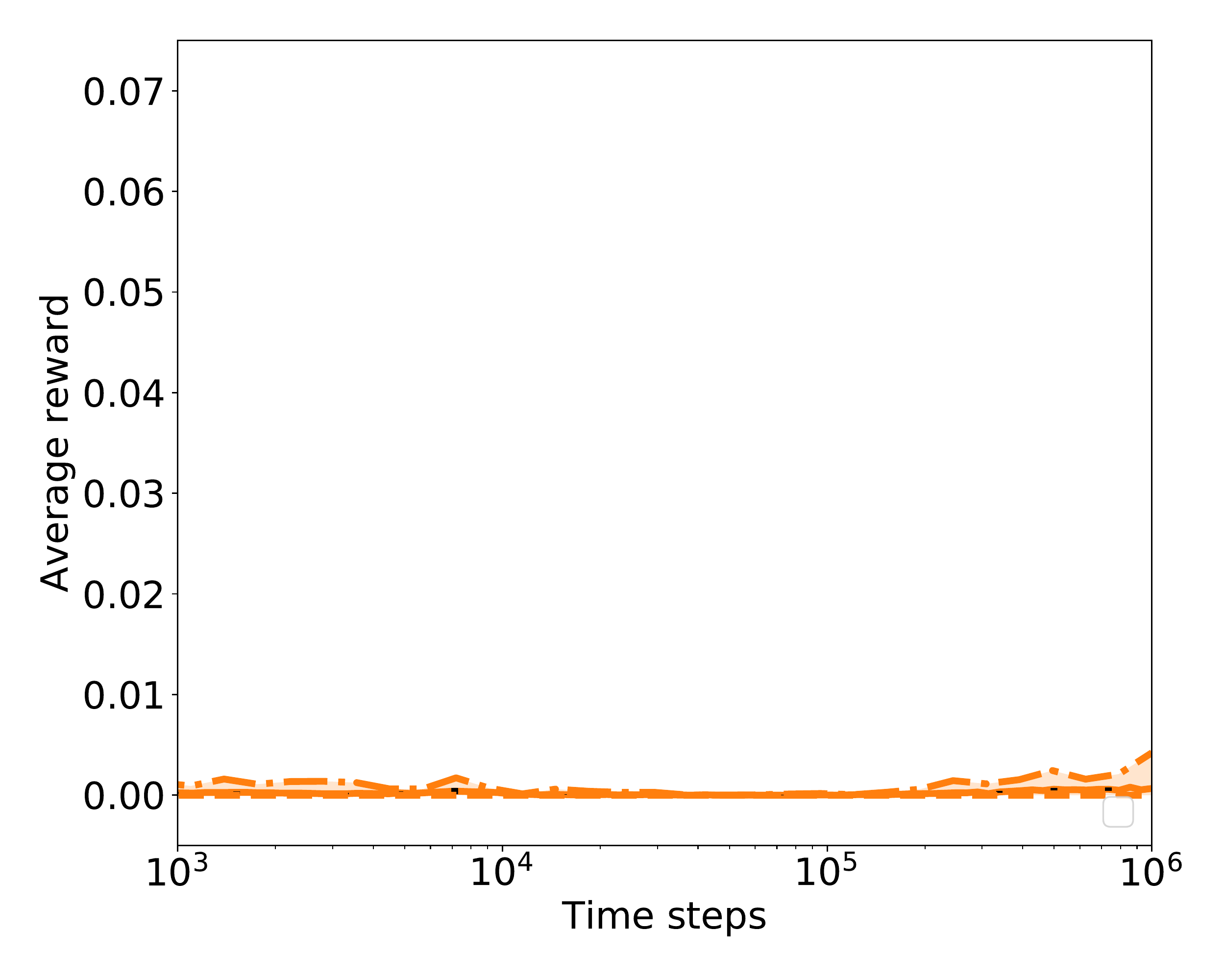}
		\caption{VDQN}
		\label{fig:mazeVDQN_ci}
	\end{subfigure}
	\begin{subfigure}[t]{0.49\textwidth}
		\includegraphics[width=\linewidth]{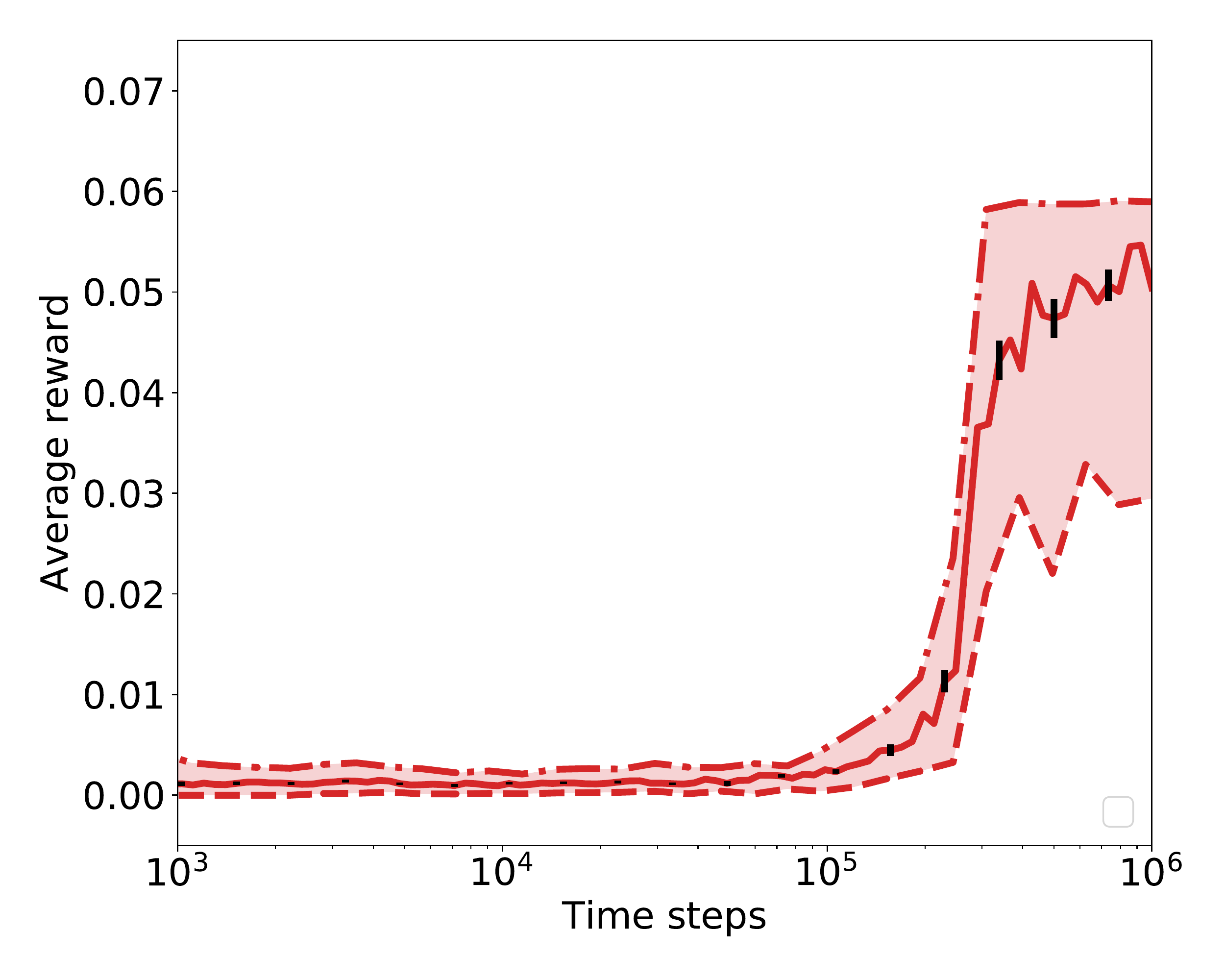}
		\caption{MMBI}
		\label{fig:mazeMMBI_ci}
	\end{subfigure}
	\caption{Evolution of average reward for Maze environment with 30 runs of length $10^6$ for each algorithm. The runs are exponentially smoothened with a half-life $1000$. The mean as well as the 5th and 95th percentile performance is shown for each algorithm and the standard error is illustrated with black lines.}\label{fig:ci_maze}
\end{figure*}

In \Cref{fig:ci_chain,fig:ci_loop,fig:ci_lava_lake_small,fig:ci_lava_lake,fig:ci_maze}, we illustrate the variability in performance of different algorithms for each environment. The black lines illustrate the standard error and the 5th and 95th percentile performance is highlighted. The results indicate that BSS is the most stable algorithm, followed by BBI, MMBI and PSRL, which nevertheless have better mean performance. VDQN is quite unstable, however.

\end{document}